\documentclass{article}

\usepackage{microtype}
\usepackage{graphicx}
\usepackage{algorithmic}
\usepackage{booktabs} % for professional tables
\usepackage{amsmath}
\usepackage{amsthm}
\usepackage{amssymb} % triangleq
\usepackage{color, colortbl}
\usepackage{xcolor}
\usepackage{hyperref}
\usepackage{cleveref}

\usepackage{enumitem}
\usepackage{bm}
\usepackage{bbm}
\usepackage{graphicx}
\usepackage{thmtools}
\usepackage{thm-restate}
\usepackage{mathtools}
\usepackage[skip=0pt]{caption}
\usepackage[skip=0pt]{subcaption}
\usepackage{booktabs} % toprule
\usepackage{fontawesome}
\usepackage{enumitem}
\usepackage{accents}

\usepackage{threeparttable, array, float} % for adding note at the end of a table

\definecolor{red}{HTML}{E51400}  %red
\definecolor{blue}{HTML}{0050EF} %cobalt
\definecolor{green}{HTML}{008A00} %emerald
\definecolor{purple}{HTML}{AA00FF} %violet
\definecolor{dark-red}{rgb}{0.4, 0.15, 0.15}
\definecolor{dark-blue}{rgb}{0.15, 0.15, 0.4}
\definecolor{medium-red}{rgb}{0.5, 0, 0}
\definecolor{medium-blue}{rgb}{0, 0, 0.5}
\definecolor{light-red}{rgb}{0.7, 0, 0}
\definecolor{light-blue}{rgb}{0, 0, 0.7}

\hypersetup{
    colorlinks=true,
    linkcolor=blue,
    filecolor=magenta,      
    urlcolor=blue,
}

\newtheorem{theorem}{\bf Theorem}
\newtheorem{lemma}{\bf Lemma}
\newtheorem{fact}{\bf Fact}
\newtheorem{condition}{\bf Condition}

\newtheorem{proposition}{\bf Proposition}

\newtheorem{definition}{\bf Definition}

\theoremstyle{definition}
\newtheorem{remark}{\bf Remark}

\definecolor{red}{HTML}{E51400} %red
\definecolor{blue}{HTML}{0050EF} %cobalt
\definecolor{green}{HTML}{008A00} %emerald
\definecolor{purple}{HTML}{AA00FF} %violet
\definecolor{orange}{HTML}{FF7F00}
\definecolor{gray}{HTML}{848482}
\definecolor{Gray}{gray}{0.85}
\definecolor{LightGray}{gray}{0.96}

\newcommand{\ccmab}{{C$^2$MAB}}

\DeclareMathOperator*{\argmin}{argmin}
\DeclareMathOperator*{\argmax}{argmax}

\newcommand{\ubar}[1]{\underaccent{\bar}{#1}}

\newcommand{\norm}[1]{\left\lVert#1\right\rVert}

\newcommand{\inner}[1]{\langle#1\rangle}
\newcommand{\cS}{\mathcal{S}}
\newcommand{\abs}[1]{\left| #1 \right|}

\newcommand{\R}{\mathbb{R}}

\newcommand{\E}{\mathbb{E}}

\newcommand{\Var}{{\rm Var}}

\newcommand{\I}{\mathbb{I}}

\newcommand{\bA}{\boldsymbol{A}}

\newcommand{\bb}{\boldsymbol{b}}
\newcommand{\bB}{\boldsymbol{B}}

\newcommand{\be}{\boldsymbol{e}}
\newcommand{\bI}{\boldsymbol{I}}

\newcommand{\bM}{\boldsymbol{M}}
\newcommand{\bG}{\boldsymbol{G}}

\newcommand{\bv}{\boldsymbol{v}}

\newcommand{\bX}{\boldsymbol{X}}

\newcommand{\bx}{\boldsymbol{x}}
\newcommand{\by}{\boldsymbol{y}}
\newcommand{\bZ}{\boldsymbol{Z}}

\newcommand{\bmu}{\boldsymbol{\mu}}

\newcommand{\btheta}{\boldsymbol{\theta}}
\newcommand{\boldeta}{\boldsymbol{\eta}}
\newcommand{\boldzeta}{\boldsymbol{\zeta}}
\newcommand{\bphi}{\boldsymbol{\phi}}

\newcommand{\cA}{\mathcal{A}}

\newcommand{\cD}{\mathcal{D}}
\newcommand{\cE}{\mathcal{E}}
\newcommand{\cF}{\mathcal{F}}

\newcommand{\cH}{\mathcal{H}}

\newcommand{\cM}{\mathcal{M}}
\newcommand{\cN}{\mathcal{N}}

\newcommand{\cW}{{\mathcal{W}}}

\newcommand{\trace}{\mathrm{trace}}

\newcommand{\tDelta}{\tilde{\Delta}}

\makeatletter
\newcommand*{\rom}[1]{\expandafter\@slowromancap\romannumeral #1@}
\makeatother

\newcommand{\ts}[1]{}

\newcommand{\compilefullversion}{true}%SHOW full version
\ifthenelse{\equal{\compilefullversion}{false}}{%
	\newcommand{\OnlyInFull}[1]{}
	\newcommand{\OnlyInShort}[1]{#1}
}{%
	\newcommand{\OnlyInFull}[1]{#1}%
	\newcommand{\OnlyInShort}[1]{}%
}%

\newcommand{\compilehidecomments}{false}%HIDE comments
\ifthenelse{ \equal{\compilehidecomments}{true} }{%
	\newcommand{\wei}[1]{}
	\newcommand{\xutong}[1]{}
	\newcommand{\jinhang}[1]{}
	\newcommand{\siwei}[1]{}
        \newcommand{\carlee}[1]{}

}{
\newcommand{\wei}[1]{{\color{blue}{[Wei: #1]}}}
\newcommand{\xutong}[1]{{\color{green} [Xutong: #1]}}
\newcommand{\jinhang}[1]{{\color{orange} [\text{Jinhang:} #1]}}
\newcommand{\siwei}[1]{{\color{red} [\text{Siwei:} #1]}}

}
\allowdisplaybreaks

\newenvironment{talign}
 {\let\displaystyle\textstyle\align}
 {\endalign}
\newenvironment{talign*}
 {\let\displaystyle\textstyle\csname align*\endcsname}
 {\endalign}


\renewcommand{\algorithmiccomment}[1]{\hfill$\triangleright$\textit{\textcolor{blue}{#1}}}
\usepackage[accepted]{icml2023}

\icmltitlerunning{Contextual Combinatorial Bandits with Probabilistically Triggered Arms}

\begin{document}

\twocolumn[
% \icmltitle{Contextual Combinatorial Bandits Revisited}
\icmltitle{Contextual Combinatorial Bandits with Probabilistically Triggered Arms}
% \icmltitle{Improved CMAB-T Algorithms with Gini-smooth Reward Functions}





\begin{icmlauthorlist}
\icmlauthor{Xutong Liu}{cuhk}
\icmlauthor{Jinhang Zuo}{umass,caltech}
\icmlauthor{Siwei Wang}{msra}
\icmlauthor{John C.S. Lui}{cuhk}
\icmlauthor{Mohammad Hajiesmaili}{umass}
\icmlauthor{Adam Wierman}{caltech}
\icmlauthor{Wei Chen}{msra}
\end{icmlauthorlist}

\icmlaffiliation{cuhk}{The Chinese University of Hong Kong, Hong Kong SAR, China}
\icmlaffiliation{umass}{University of Massachusetts Amherst, MA, United States}
\icmlaffiliation{caltech}{California Institute of Technology, CA, United States}
\icmlaffiliation{msra}{Microsoft Research, Beijing, China}
\icmlcorrespondingauthor{Xutong Liu}{liuxt@cse.cuhk.edu.hk}
\icmlcorrespondingauthor{Wei Chen}{weic@microsoft.com}
\icmlcorrespondingauthor{John C.S. Lui}{cslui@cse.cuhk.edu.hk}
\icmlkeywords{Machine Learning, ICML}

\vskip 0.3in
]



\printAffiliationsAndNotice{}  % leave blank if no need to mention equal contribution

\begin{abstract}

We study contextual combinatorial bandits with probabilistically triggered arms (C$^2$MAB-T) under a variety of smoothness conditions that capture a wide range of applications, such as contextual cascading bandits and contextual influence maximization bandits. Under the triggering probability modulated (TPM) condition, we devise the C$^2$-UCB-T algorithm and propose a novel analysis that achieves an $\tilde{O}(d\sqrt{KT})$ regret bound, removing a potentially exponentially large factor $O(1/p_{\min})$, where $d$ is the dimension of contexts, $p_{\min}$ is the minimum positive probability that any arm can be triggered, and batch-size $K$ is the maximum number of arms that can be triggered per round. Under the variance modulated (VM) or triggering probability and variance modulated (TPVM) conditions, we propose a new variance-adaptive algorithm  VAC$^2$-UCB and derive a regret bound $\tilde{O}(d\sqrt{T})$, which is independent of the batch-size $K$.  As a valuable by-product, our analysis technique and variance-adaptive algorithm can be applied to the CMAB-T and C$^2$MAB setting, improving existing results there as well. We also include experiments that demonstrate the improved performance of our algorithms compared with benchmark algorithms on synthetic and real-world datasets.

\end{abstract}

%%\vspace{-20pt}
\section{Introduction}

\begin{table*}[t]
	\caption{Summary of the main results for \ccmab-T, and additional results for CMAB-T and \ccmab.}	\label{tab:ccmab_res}	
	\centering
	\resizebox{1.96\columnwidth}{!}{
	\centering
	\begin{threeparttable}
	\begin{tabular}{|ccccc|}
 \hline
 % \multicolumn{5}{}{}
 % & \textbf{Summary }\\
 % \hline\hline
		\textbf{\ccmab-T}&\textbf{Algorithm}&\textbf{Condition}& \textbf{Coefficient} & \textbf{Regret Bound}\\
  \hline
    & C$^3$-UCB   \cite{li2016contextual}$^{*}$  & 1-norm & $B_1$   & $O({B_1d\sqrt{KT} \cdot\log T}/{p_{\min}})$\\
        %   \rowcolor{Gray}
        % & C$^2$-UCB-T  (\Cref{alg: C2UCB}) & 1-norm TPM& $B_1$  & $O(B_1d\sqrt{T}\cdot K^{1/2}\cdot\log (KT))$\\
                  \rowcolor{Gray}
       \textbf{(Main Result \ref{thm:c2mab_reg})} & C$^2$-UCB-T  (\Cref{alg: C2UCB}) & 1-norm TPM& $B_1$  & $O(B_1d\sqrt{KT}\cdot\log T)$\\
                     \rowcolor{Gray}
\textbf{(Main Result \ref{thm:ccmab_reg_VM})}& VAC$^2$-UCB (\Cref{alg:va_cucb})  & VM & $B_v$$^{\dagger}$& $O(B_v d\sqrt{T} \cdot \log T/\sqrt{p_{\min}})$\\
 \rowcolor{Gray}
	\textbf{(Main Result \ref{thm:ccmab_reg_TPVM_lambda})}& 	VAC$^2$-UCB 
 (\Cref{alg:va_cucb})  & TPVM  & $B_v$, $\lambda\ge1$$^{\ddagger}$ & $O(B_v d\sqrt{T} \cdot \log T)$\\
          % \rowcolor{Gray}
	% & 	VAC$^2$-UCB-K 
 % (Appendix XXX)  & TPVM, known variance bound   & $B_v$, $\lambda\ge1$ & $O(B_v d\sqrt{T} \cdot \log T)$\\
    \hline\hline
    \textbf{CMAB-T}&\textbf{Algorithm}&\textbf{Condition}& \textbf{Coefficient} & \textbf{Regret Bound}\\
  \hline
 % \textbf{Summary }&\textbf{of }&\textbf{additional }& \textbf{results} & \textbf{}\\
 % \hline\hline	
	 % &	CUCB~\cite{chen2016combinatorial}&  1-norm & $B_1$  & $O(B_1\sqrt{mKT}\cdot (\log T)^{1/2} /p_{\min})$ \\
  %  &	CUCB~\cite{wang2017improving}&  1-norm TPM & $B_1$  & $O(B_1\sqrt{mKT}\cdot (\log T)^{1/2})$ \\
  %  &BCUCB-T \cite{liu2022batch} & TPVM& $B_v$, $\lambda\ge2$ & $O(B_v \sqrt{m(\log K)T}\cdot(\log T)^{1/2})$\\
		&BCUCB-T \cite{liu2022batch} & TPVM& $B_v$, $\lambda\ge 1$  & $O(B_v \sqrt{m(\log K)T}\cdot\log T)$\\
  \rowcolor{Gray}
  % &BCUCB-T (Our new results)  & TPVM& $B_v$, $\lambda\ge 1$   & $O(B_v \sqrt{mT}\cdot(\log K )^{1/2}\cdot(\log T)^{1/2}\cdot p_{\max}^{(\lambda-1)/2})$$^{\ddagger}$ \\
   \textbf{(Additional Result 1)} &BCUCB-T (Our new analysis)  & TPVM& $B_v$, $\lambda\ge 1$   & $O(B_v \sqrt{m(\log K )T}\cdot(\log T)^{1/2})^{**}$ \\
		\hline\hline
    \textbf{\ccmab} &\textbf{Algorithm}&\textbf{Condition}& \textbf{Coefficient} & \textbf{Regret Bound}\\
    \hline
   &	C$^2$UCB~\cite{qin2014contextual}&  2-norm & $B_2$$^{\S}$  & $O(B_2d\sqrt{T}\log T)$ \\
  &	C$^2$UCB~\cite{takemura2021near}&  1-norm & $B_1$  & $O(B_1d\sqrt{KT}\log T)$ \\
       \rowcolor{Gray}
   \textbf{(Additional Result 2)}&	VAC$^2$-UCB~(\Cref{alg:va_cucb})& VM & $B_v$  & $O(B_vd\sqrt{T}\log T)$ \\
	\hline
	\end{tabular}
	  \begin{tablenotes}[para, online,flushleft]
	\footnotesize%\smallskip
 \item[]\hspace*{-\fontdimen2\font}$^{*}$ This work is specified for contextual combinatorial cascading bandits, without formally defining the  arm triggering process.
 \item[]\hspace*{-\fontdimen2\font}$^\dagger$ Generally,  
 coefficient $B_v=O(B_1\sqrt{K})$ and the existing regret bound is improved when $B_v=o(B_1\sqrt{K})$
 \item[]\hspace*{-\fontdimen2\font}$^{\ddagger}$ $\lambda$ is a coefficient in TPVM condition: when $\lambda$ is larger, the condition is stronger with smaller regret but can include less applications.
 \item[]\hspace*{-\fontdimen2\font}$^{**}$ We also show improved distribution-dependent regret bound in \Cref{apdx_sec:NC_C2MAB};
 \item[]\hspace*{-\fontdimen2\font}$^{\S}$ Almost all applications satisfy $B_2=\Theta(B_1\sqrt{K})$.
 
	% \item[]\hspace*{-\fontdimen2\font}$^*$ Requires exact offline oracle instead of $(\alpha, \beta)$-approximate oracle;
	% 	\item[]\hspace*{-\fontdimen2\font}$^{\dagger}$ This work gives sufficient smoothness condition with factor $\gamma_g$ and translates to $B_v=3\sqrt{2}\gamma_g$ in our setting;
	% 	\item[]\hspace*{-\fontdimen2\font}$^{*}$ Using our new and simplified  analysis;
	% \item[]\hspace*{-\fontdimen2\font}$^\dagger$ Generally,  
 % coefficient $B_v=O(B_1\sqrt{K})$ and the existing regret bound is improved when $B_v=o(B_1\sqrt{K})$; Coefficient $\lambda$ trades off the regret and satisfiability: when $\lambda$ is larger, the regret is better but TPVM condition can include less applications. 
 % 	\item[]\hspace*{-\fontdimen2\font}$^{\S}$ This work is specified for contextual combinatorial cascading bandits, without formally defining the  arm triggering process.
	% 	\item[]\hspace*{-\fontdimen2\font}$^{\ddagger}$ We also show improved distribution-dependent regret bound in Appendix XXX;
	\end{tablenotes}
			\end{threeparttable}
	}
\end{table*}

\ts{CMAB short intro}

The stochastic multi-armed bandit (MAB) problem is a classical sequential decision-making problem that has been widely studied~\cite{robbins1952some,auer2002finite,bubeck2012regret}. As an extension of MAB, combinatorial multi-armed bandits (CMAB) have drawn attention due to fruitful applications in online advertising, network optimization, and healthcare systems \cite{gai2012combinatorial, kveton2015combinatorial, chen2013combinatorial, chen2016combinatorial, wang2017improving, merlis2019batch}.
CMAB is a sequential decision-making game between a learning agent and an environment. In each round, the agent chooses a combinatorial action that triggers a set of base arms (i.e., a super-arm) to be pulled simultaneously, and the outcomes of these pulled base arms are observed as feedback (typically known as semi-bandit feedback). The goal of the agent is to minimize the expected \textit{regret}, which is the difference in expectation for the overall rewards between always playing the best action (i.e., the action with the highest expected reward) and playing according to the agent's own policy. 
% For CMAB, the agent not only needs to handle the exploration-exploitation tradeoff: whether the agent should explore arms in search for a better action, or should the agent stick to the best action observed so far to gain rewards; but also n eeds to deal with the exponential explosion of all possible actions. 

\ts{Intro of \ccmab~and drawbacks of \ccmab}

Motivated by large-scale applications with a huge number of items (base arms), there exists a prominent line of work that advances the CMAB model: the combinatorial contextual bandits (or \ccmab~for short) ~\cite{qin2014contextual,li2016contextual,takemura2021near}. Specifically, \ccmab~incorporates contextual information and adds the simple yet effective linear structure assumption to allow scalability, which provides regret bounds that are independent of the number of base arms $m$. Despite \ccmab's success in leveraging contextual information for better scalability, existing works fail to formulate the general arm triggering process, which is essential to model a wider range of applications, e.g., cascading bandits (CB) and influence maximization (IM), and more importantly, they do not provide satisfying results for settings with probabilistically triggered arms. For example, \citet{qin2014contextual,takemura2021near} only consider the deterministic semi-bandit feedback for \ccmab. \citet{li2016contextual,wen2017online} implicitly consider the arm triggering process for specific CB or IM applications but only gives sub-optimal results with unsatisfying factors (e.g., $1/p_{\min}, K$ that could be as large as the number of base arms), owing to loose analysis, weak conditions, or inefficient algorithms that explore the unknown parameters too conservatively. 

To handle the above issues, we enhance the \ccmab~framework by considering an arm triggering process.
Specifically, we propose the general framework of contextual combinatorial bandits with probabilistically triggered arms (or \ccmab-T for short). At the base arm level, \ccmab-T uses a time-varying feature map $\bphi_t$ to model the contextual information at each round $t$, and the mean outcome of each arm $i \in [m]$ is a linear product of the feature vector $\bphi_t(i) \in \R^d$ and an unknown vector $\theta^* \in \R^d$ (where $d\ll m$ to handle large-scale applications). At the (combinatorial) action level, inspired by the non-contextual CMAB with probabilistically triggered arms (or CMAB-T) works~\cite{chen2016general,wang2017improving,liu2022batch}, 
% \wei{\cite{liu2022batch} should be referring to the NeurIPS'2022 paper now.}
we formally define an arm-triggering process to cover more general feedback models such as semi-bandit, cascading, and probabilistic feedback.
We also inherit smoothness conditions for the non-linear reward function to cover different application scenarios, such as CB, IM, and online probabilistic maximum coverage (PMC) problems~\cite{chen2016combinatorial, wang2017improving}. With this formulation, \ccmab-T retains \ccmab's scalability while also enjoying CMAB-T's rich reward functions and general feedback models.

\textbf{Contributions.} 
%
% \textbf{General Framework.} 
% To improve the model part, we propose the general framework of contextual combinatorial bandits with probabilistically triggered arms (or \ccmab-T for short). At the base arm level, \ccmab-T uses a time-varying feature map $\bphi_t$ to model the contextual information at each round $t$, and the mean of the outcomes of each arm $i \in [m]$ is a linear product of the feature vector $\bphi_t(i) \in \R^d$ and the unknown vector $\theta^* \in \R^d$ (where $d\ll m$ to handle large-scale applications). At the (combinatorial) action level, inspired by CMAB-T work, we formally define the arm-triggering process to cover general feedback models including semi-bandit, cascading, and probabilistic feedback.
% We also inherit proper smoothness conditions for the non-linear reward function to cover different application scenarios, such as cascading bandits (CB), online influence maximization (IM), and online probabilistic maximum coverage (PMC) problems. With this formulation, \ccmab-T enjoys \ccmab's scalability for large-scale applications, and the CMAB-T's rich reward functions and probabilistic feedback models.
% It is noted that there are prior works studying special cases that fit into our framework (e.g., contextual CB~\cite{li2016contextual} and contextual OIM~\cite{wen2017online}), but we are the first to study the unified framework and more importantly, we give new algorithms/analysis for improved results.
Our main results are shown in \Cref{tab:ccmab_res}. 

First, we study \ccmab-T under the triggering probability modulated (TPM) smoothness condition, a condition introduced by~\citet{wang2017improving} to remove a factor of $1/p_{\min}$ in the pioneer CMAB-T work~\cite{chen2016combinatorial}. This result follows a similar vein by devising C$^2$-UCB-T algorithm and proving a $\tilde{O}(d\sqrt{KT})$ regret, which removes a $1/p_{\min}$ factor for prior contextual CB applications~\cite{li2016contextual} (Main Result 1 in \Cref{tab:ccmab_res}). The key technical challenge is that the triggering group (TG) analysis~\cite{wang2017improving} for CMAB-T cannot handle the triggering probability determined by time-varying contexts. To tackle this issue, we devise a new technique, called the triggering probability equivalence (TPE), which links the triggering probabilities with the random triggering event under expectation. 
In this way, we no longer need to bound the regret caused by possibly triggered arms, but only need to bound the regret caused by actually triggered arms. 
As a result, we can then directly apply the simple non-triggering \ccmab~analysis to obtain the regret bound for \ccmab-T. 
In addition, our TPE can reproduce the results for CMAB-T in a similar way.

Second, we study the \ccmab-T under the variance modulated (VM) smoothness condition~\cite{liu2022batch}, in light of the recent variance-adaptive algorithms to remove the batch size dependence $O(\sqrt{K})$ for CMAB-T~\cite{merlis2019batch,liu2022batch,vial2022minimax}. We propose a new variance-adaptive algorithm VAC$^2$-UCB and prove a batch-size independent regret $\tilde{O}(d\sqrt{T/p_{\min}})$ under VM condition (Main Result 2 in \Cref{tab:ccmab_res}). The main technical difficulty is to deal with the unknown variance. Inspired by \citet{lattimore2015linear}, we use the UCB/LCB value to construct an optimistic variance and on top of that, we prove a new concentration bound to incorporate the triggered arms and optimistic variance to get the desirable results. 

Third, we investigate the stronger triggering probability and variance modulated (TPVM) condition~\cite{liu2022batch} in order to remove the additional $1/\sqrt{p_{\min}}$ factor. 
The key challenge is that we cannot directly use TPE to link the \textit{true} triggering probability with the random trigger event as before, since the TPVM condition only yields a mismatched triggering probability associated with the optimistic variance used in the algorithm. Our solution is to bound this additional mismatch by lower-order terms based on mild conditions on the triggering probability, which achieves the $\tilde{O}(d\sqrt{T})$ regret bounds (Main Result 3 in \Cref{tab:ccmab_res}). 
% The key challenge is that in order to use the variance information, the TPVM condition only produces a mismatched triggering probability rather than the true probability, so we cannot trivially apply the TPE trick as before. Our solution is to either consider mild conditions on the triggering probability to bound the amount of this mismatch and prove that this match, which eventually achieves the $\tilde{O}(d\sqrt{T})$ regret bounds (Main Result 3 in \Cref{tab:ccmab_res}). 

As a valuable by-product, our TPE analysis and VAC$^2$-UCB algorithm can be applied to non-contextual CMAB-T and \ccmab, improving the existing results by a factor $\sqrt{\log T}$ (Additional Result 1 in \Cref{tab:ccmab_res}) and $\sqrt{K}$ (Additional Result 2 in \Cref{tab:ccmab_res}), respectively.
Our empirical results on both synthetic and real data demonstrate that the VAC$^2$-UCB algorithm outperforms the state-of-art variance-agnostic and variance-aware bandit algorithms in the linear cascading bandit application that satisfies the TPVM condition.

% In addition to the general \ccmab-T setting, we find that the TPE trick and the variance-adaptive algorithm can help to improve the degenerate CMAB-T and \ccmab~settings, respectively. For CMAB-T, our TPE trick reproduces the results under 1-norm TPM condition and improves the results by a factor $\sqrt{\log T}$, compared with the more involved TG type of analysis. For \ccmab, our variance-adaptive algorithm achieves the first batch-size independent regrets for applications satisfying VM conditions.

% \adam{Do you want to add a short paragraph about numeric results here?}

\textbf{Related Work.}
% The multi-armed bandit (MAB) problem is first studied by \citet{robbins1952some} and then extended by many studies (cf. \cite{bubeck2012regret, slivkins2019introduction, lattimore2020bandit}).
% \jinhang{can be removed}
The stochastic CMAB model has received significant attention.  The literature was initiated by \cite{gai2012combinatorial}.  Since, its regret has been improved by \citet{kveton2015tight}, \citet{combes2015combinatorial}, \citet{chen2016combinatorial}. There exist two prominent lines of work in the literature related to our study: contextual CMAB and CMAB with probabilistically triggered arms (\ccmab~and CMAB-T). 
%At a high level, our work can be viewed as the first framework that unifies the \ccmab~and CMAB-T, which enjoys the best of both lines' properties and results. There are many studies considering specific applications rather than a general \ccmab-T framework, including contextual CB~\cite{zong2016cascading,li2016contextual,vial2022minimax}, contextual IM~\cite{wen2017online}, etc. These applications can fit into our framework by proving that they satisfy the TPM, VM, or TPVM conditions, and achieve improved results regarding $K, p_{\min}$ factors. We defer the detailed technical comparison to \Cref{sec:c2mab}, \Cref{sec:va_c2mab}, and \Cref{sec:app}. \citet{zuo2022online} studies the online competitive IM and also uses \ccmab-T to denote their contextual setting. Their context is at the action level, and only affects the reward function (or regret) but not the base arms' estimation, which is different and much simpler than our setting. 

For \ccmab~works, \citet{qin2014contextual} is the first study, which proposes C$^2$UCB algorithm that considers reward functions under 2-norm $B_2$ smoothness condition. Then \citet{takemura2021near} replaces the 2-norm smoothness condition with a new 1-norm $B_1$ smoothness condition, and proves a $O(B_1d\sqrt{KT}\log T)$ regret bound. 
In this work, we extend the \ccmab~with triggering arms to cover more application scenarios (e.g., contextual CB and contextual IM). Moreover, we further consider the stronger VM condition and propose a new variance-adaptive algorithm that can achieve a $\sqrt{K}$ factor improvement in the regret upper bound for applications like PMC.
%
%at least give matching regret bounds, and further achieve a $\sqrt{K}$ factor improvement under VM condition with $B_v=o(B_1\sqrt{K})$ (for applications like PMC).

For CMAB-T works, \citet{chen2016combinatorial} is the first work that considers the arm triggering process to cover CB and IM applications. The authors propose the CUCB algorithm, and give an $O(B_1\sqrt{mKT\log T}/p_{\min})$ regret bound under 1-norm $B_1$ smoothness condition. Then \citet{wang2017improving} proposes the stronger 1-norm triggering probability modulated (TPM) $B_1$ smoothness condition, and use the triggering group (TG) analysis to remove a $1/p_{\min}$ factor in the previous regret. Recently, \citet{liu2022batch} incorporates the variance information, and proposes the variance-adaptive algorithm BCUCB-T, which also uses the TG analysis and further reduces the regret's dependency on batch-size from $O(K)$ to $O(\log K)$ under the new variance and triggering probability modulated (TPVM) condition. The smoothness conditions considered in this work are mostly inspired by the above works, but directly following their algorithm and TG analysis fail to obtain any meaningful result for our \ccmab-T setting. 
% paralyzes \adam{not sure what you mean with "paralyzes" here} the direct use of CMAB-T's TG analysis. 
Conversely, our new TPE analysis can be applied to CMAB-T, reproducing CMAB-T's result under the 1-norm TPM condition, and improving a factor of $(\sqrt{\log T})$ under the TPVM condition. 

There are also many studies considering specific applications under the \ccmab-T framework (by unifying \ccmab~and CMAB-T), including contextual CB~\cite{li2016contextual,vial2022minimax}, contextual IM~\cite{wen2017online}, etc. One can see that these applications fit into our framework by verifying that they satisfy the TPM, VM, or TPVM conditions; thus achieving improved results regarding $K, p_{\min}$ factors.  We defer a detailed theoretical and empirical comparison to \Cref{sec:c2mab,sec:va_c2mab,sec:app}. \citet{zuo2022online} study the online competitive IM and also uses \ccmab-T to denote their contextual setting. However, their meaning of ``contexts" is the action of the competitor, which acts at the action level and only affects the reward function (or regret) but not the base arms' estimation.
This is very different from our setting, where contexts act at the base arm level and hence one cannot directly apply their results.

\section{Problem Setting}\label{sec: problem setting}
\ts{Overall nations and the \ccmab-T tuple.}

We study contextual combinatorial bandits with probabilistically triggered arms (\ccmab-T). We use $[n]$ to represent set $\{1,...,n\}$. We use boldface lowercase letters and boldface CAPITALIZED letters for column vectors and matrices, respectively. $\norm{\bx}_p$ denotes the $\ell_p$ norm of vector $\bx$. For any symmetric positive semi-definite (PSD) matrix $\bM$ (i.e., $\bx^{\top} \bM \bx \ge 0, \forall \bx$), $\norm{\bx}_{\bM}=\sqrt{\bx^{\top} \bM \bx}$ denotes the matrix norm of $\bx$ regarding matrix $\bM$.

We specify a \ccmab-T problem instance using a tuple $([m], \cS, \Phi, \Theta, D_{\text{trig}}, R)$, where $[m]=\{1,2,...,m\}$ is the set of base arms (or arms); 
$\cS$ is the set of eligible actions where $S \in \cS$ is an action;\footnote{$\cS$ is a general action space. When $\cS$ is a collection of subsets of $[m]$, we often 
refer to $S\in \cS$ as a super arm.} 
$\Phi$ is the set of possible feature maps where any feature map $\bphi \in \Phi$ is a function $[m]\rightarrow \R^d$ that maps an arm to a $d$-dimensional feature vector (and w.l.o.g. we normalize $\norm{\bphi(i)}_2 \le 1$); $\Theta\subseteq R^d$ is the parameter space; 
$ D_{\text{trig}}$ is the probabilistic triggering function to characterize the arm triggering process (and feedback), and $R$ is the reward function.

\ts{The learning protocol.}

In \ccmab-T, a learning game is played between a learning agent (or player) and the unknown environment in a sequential manner.
Before the game starts, the environment chooses a parameter $\btheta^* \in \Theta$ unknown to the agent (and w.l.o.g. we also assume $\norm{\theta^*}_2 \le 1$). At the beginning of round $t$, the environment reveals feature vectors $(\bphi_t(1), ..., \bphi_t(m))$ for each arm, where $\bphi_t \in \Phi$ is the feature map known to the agent.  Given $\bphi_t$, the agent selects an action $S_t \in \cS$, and the environment draws Bernoulli outcomes $\bX_t=(X_{t,1},...X_{t,m})\in \{0,1\}^m$ for base arms\footnote{ We assume $X_{t,i}$ are Bernoulli for the ease of exposition, yet our setting can handle any distribution with bounded $X_{t,i}\in[0,1]$.}, with mean $\E[X_{t,i} | \cH_t]=\inner{\btheta^*, \bphi_{t}(i)}$ for each base arm $i$. 
Here $\cH_t$ denotes the history before the agent chooses $S_t$ and will be specified shortly after. Note that the outcome $\bX_t$ is assumed to be conditional independent across arms given history $\cH_t$, similar to previous works~\cite{qin2014contextual,li2016contextual,vial2022minimax}. For convenience, we use $\bmu_t\triangleq(\inner{\btheta^*, \bphi_{t}(i)})_{i \in [m]}$ to denote the mean vector and $\cM\triangleq\{\inner{\btheta, \bphi(i)}_{i \in [m]}: \bphi \in \Phi, \btheta \in \Theta\}$ to denote all possible mean vectors generated by $\Phi$ and $\Theta$. 
% \siwei{Should we say that $\inner{\btheta^*, \bphi_{t}(i)} \in [0,1]$?}.

After the action $S_t$ is played on the outcome $\bX_t$, base arms in a random set $\tau_t \sim D_{\text{trig}}(S_t, \bX_t)$ are triggered, 
meaning that the outcomes of arms in $\tau_t$, i.e. $(X_{t,i})_{i\in \tau_t}$ are revealed as the feedback to the agent, and are involved in determining the reward of action $S_t$. 
At the end of round $t$, the agent will receive a non-negative reward $R(S_t, \bX_t, \tau_t)$, determined by $S_t,
\bX_t$ and $\tau_t$, and similar to~\cite{wang2017improving}, the expected reward is assumed to be $r(S_t;\bmu_t)\triangleq \E[R(S_t,\bX_t,\tau_t)]$, a function of 
the unknown mean vector $\bmu_t$, where the expectation is taken over the randomness of $\bX_t$ and $\tau_t \sim D_{\text{trig}}(S_t,\bX_t)$.
To allow the algorithm to estimate the underlying parameter $\btheta^*$ directly from samples, we assume the outcome does not depend on whether the arm $i$ is triggered, i.e., $\E_{\bX \sim D, \tau \sim D_{\text{trig}}(S,\bX)}[X_i | i\in \tau]=\E_{\bX\sim D}[X_i]$.
To this end, we can give the formal definition of the history $\cH_t=(\bphi_s, S_s, \tau_s, (X_{s,i})_{i\in \tau_s})_{s<t} \bigcup \bphi_t$, which contains all information before round $t$, as well as the contextual information $\bphi_t$ at round $t$.

\ts{The goal and the evaluation metrics (approximate regret).}

The goal of CMAB-T is to accumulate as much reward as possible over $T$ rounds by learning the underlying parameter $\theta^*$.
The performance of an online learning algorithm $A$ is measured by its \textit{regret}, defined as the difference of the expected cumulative reward between always playing the best action $S^*_t \triangleq \argmax_{S \in \cS}r(S;\bmu_t)$ in each round $t$ and playing actions chosen by algorithm $A$.
%Let $S^*=\argmax_{S \in \cS}r(S;\bmu)$ be the optimal action. 
For many reward functions, it is NP-hard to compute the exact $S^*_t$ even when $\bmu_t$ is known, so similar to~\cite{wang2017improving}, we assume that the algorithm $A$ has access to an offline $(\alpha, \beta)$-approximation oracle, which for mean vector $\bmu$ outputs an action $S$ such that $\Pr\left[r(S;\bmu)\ge \alpha \cdot r(S^*;\bmu)\right] \ge \beta$. 
% \jinhang{$\beta$}
The $T$-round $(\alpha, \beta)$-approximate regret is defined as
% \resizebox{1.0\columnwidth}{!}{
% \begin{minipage}{\columnwidth}
\begin{equation}\textstyle
    \text{Reg}(T)=\E\left[\sum_{t=1}^T \left(\alpha\beta  \cdot r(S^*_t;\bmu_t)-r(S_t;\bmu_t)\right)\right],
\end{equation}
% \end{minipage}}
where the expectation is taken over the randomness of outcomes $\bX_1, ..., \bX_T$, the triggered sets $\tau_1, ..., \tau_T$, as well as the randomness of algorithm $A$ itself.

% \wei{I feel that the remarks do not need to be put in italic style. Only theorems and lemmas are put into italic to emphasize them. But either is fine.}

\begin{remark}[\textbf{Difference with CMAB-T}]\label{rmk:cmab-T_diff}
\ccmab-T strictly generalizes CMAB-T by allowing a probably time-varying feature map $\bphi_t$. Specifically, let $\btheta^*=(\mu_1, ..., \mu_m)$ and fix $\bphi_t(i)=\be_i$ where $\be_i \in \R^{m}$ is the one-hot vector with $1$ at the $i$-th entry and $0$ elsewhere, one can easily reproduce the CMAB-T setting in~\cite{wang2017improving}.
\end{remark}

\begin{remark}[\textbf{Difference with C$^2$MAB}]
\ccmab-T enhances the modeling power of prior \ccmab~\cite{qin2014contextual,takemura2021near} by capturing the probabilistic nature of the feedback (v.s. the deterministic semi-bandit feedback). This enables a wider range of applications such as combinatorial CB, multi-layered network exploration, and online IM~\cite{wang2017improving,liu2022batch}.
\end{remark}

\subsection{Key Quantities and Conditions}
\ts{Introduce two key quantities.}

In the \ccmab-T model, there are several quantities and assumptions that are crucial to the subsequent study.
We define \textit{{triggering probability}} $p_i^{\bmu, D_{\text{trig}}, S}$ as the probability that base arm $i$ is 
	triggered when the action is $S$, the mean vector is $\bmu$, and the probabilistic triggering function is $D_{\text{trig}}$.
Since $D_{\text{trig}}$ is always fixed in a given application context, we ignore it in the notation for simplicity, and use $p_{i}^{\bmu,S}$ henceforth.
Triggering probabilities  $p_i^{\bmu,S}$'s are crucial for the triggering probability modulated bounded smoothness conditions to be defined below.
We define $\tilde{S}$ to be the set of arms that can be triggered by $S$,  i.e.,$\{i \in [m]: p_i^{\bmu,S} > 0, \text{ for any }\bmu \in \cM\}$, the  \textit{{batch size}} $K$ as the maximum number of arms that can be triggered, i.e., $K=\max_{S \in \cS}|\tilde{S}|$, and $p_{\min}=\min_{i \in [m], \bmu \in \cM, S \in \cS, p_i^{\bmu,S}>0}{p_i^{\bmu,S}}$.

% Our main contribution of this paper is to remove or reduce the regret dependency on batch size $K$, where $K$ could be quite large, e.g., $K$ can be hundreds of thousands in a large social network.

\ts{Explain necessary conditions: monotonicity, TPM, TPVM smoothness.}

Owing to the nonlinearity and the combinatorial structure of the reward, it is essential to give some conditions for the reward function in order to achieve any meaningful regret bounds~\cite{chen2013combinatorial, chen2016combinatorial, wang2017improving,merlis2019batch,liu2022batch}. 
For \ccmab-T, we consider the following conditions.

% For the reward function $r(S;\bmu): \cS \times [0,1]^m \rightarrow \R$, we are interested in functions that are continuous for $\bmu \in [0,1]^m$ and differentiable for $\bmu \in (0,1)^m$, for any $S \in \cS$.
% For CMAB-T framework, \citet{wang2017improving} introduce two standard conditions for the reward function to guarantee the regret bounds, which we also use in this work, as we shall introduce in the following.

\begin{condition}[\textbf{Monotonicity}]\label{cond:mono}
We say that a \ccmab-T problem instance satisfies monotonicity condition, if for any action $S \in \cS$, any mean vectors $\bmu,\bmu' \in [0,1]^m$ such that $\mu_i \le \mu'_i $ for all $i \in [m]$, we have $ r(S;\bmu) \le r(S;\bmu') $. 
\end{condition}

\begin{condition}[\textbf{1-norm TPM Bounded Smoothness}, \cite{wang2017improving}]\label{cond:TPM}
% We say that a CMAB-T instance satisfies TPM $B_1$-bounded smoothness condition, if for any action $S \in \cS$, any distribution $D,D'$ with mean vector $\bmu \in (0,1)^m$, we have $\frac{\partial r(S;\bmu)}{\partial \mu_i } \frac{1}{p_{i}^{D,S}}\le B_1 $ for $i \in \tilde{S}$ and $\frac{\partial r(S;\bmu)}{\partial \mu_i }=0$ for $i \in [m] \backslash \tilde{S}$. 
We say that a \ccmab-T problem instance satisfies the triggering probability modulated (TPM) $B_1$-bounded smoothness condition, if for any action $S \in \cS$, any mean vectors $\bmu, \bmu' \in [0,1]^m$, we have $|r(S;\bmu')-r(S;\bmu)|\le B_1\sum_{i \in [m]}p_{i}^{\bmu,S}|\mu_i-\mu'_i|$.
%\siwei{seems no formal definition of this $p_{i}^{D,S}$?} 
\end{condition}

\begin{condition}[\textbf{VM Bounded Smoothness}, \cite{liu2022batch}]\label{cond:VM}
We say that a \ccmab-T problem instance satisfies the variance modulated (VM) $(B_v, B_1)$-bounded smoothness condition, if for any action $S \in \cS$, mean vector $\bmu, \bmu' \in (0,1)^m$, for any $\boldzeta, \boldeta \in [-1,1]^m$ s.t. $\bmu'=\bmu+\boldzeta+\boldeta$, we have
% $|r(S;\bmu')-r(S;\bmu)|\le B_v\norm{\left(\frac{(p_i^{\bmu,S})^{\lambda/2}}{\sqrt{(1-\mu_i)\mu_i}}\zeta_i\right)_{i\in[m]}}_{2} + B_1\norm{ (p_i^{\bmu,S}\eta_i)_{i \in [m]}}_1.$
$|r(S;\bmu')-r(S;\bmu)|\le B_v\sqrt{\sum_{i\in \tilde{S}}\frac{\zeta_i^2 }{(1-\mu_i)\mu_i}} + B_1 \sum_{i\in\tilde{S}}\abs{\eta_i}$. 
\end{condition}

\begin{condition}[\textbf{TPVM Bounded Smoothness}, \cite{liu2022batch}]\label{cond:TPVMm}
We say that a \ccmab-T problem instance satisfies the triggering probability and variance modulated (TPVM) $(B_v, B_1,\lambda)$-bounded smoothness condition, if for any action $S \in \cS$, mean vector $\bmu, \bmu' \in (0,1)^m$, for any $\boldzeta, \boldeta \in [-1,1]^m$ s.t. $\bmu'=\bmu+\boldzeta+\boldeta$, we have
% $|r(S;\bmu')-r(S;\bmu)|\le B_v\norm{\left(\frac{(p_i^{\bmu,S})^{\lambda/2}}{\sqrt{(1-\mu_i)\mu_i}}\zeta_i\right)_{i\in[m]}}_{2} + B_1\norm{ (p_i^{\bmu,S}\eta_i)_{i \in [m]}}_1.$
$|r(S;\bmu')-r(S;\bmu)|\le B_v\sqrt{\sum_{i\in [m]}(p_i^{\bmu,S})^{\lambda}\frac{\zeta_i^2 }{(1-\mu_i)\mu_i}} + B_1 \sum_{i\in[m]}p_i^{\bmu,S}\abs{\eta_i}$. 
\end{condition}

% \wei{It is unclear from the context whether these conditions are new or defined before. Perhaps adding reference to each condition.}

% \siwei{Should we put Condition 4 before Condition 3, as the order of your main results?}

\ts{Intuitions and remarks of the conditions}

Condition~\ref{cond:mono} indicates the reward is monotonically increasing when the parameter $\bmu$ increases. Condition~\ref{cond:TPM}, \ref{cond:VM} and \ref{cond:TPVMm} all bound the reward smoothness/sensitivity.%, i.e., the amount of the reward change caused by the parameter change from $\bmu$ to $\bmu'$.

% As mentioned in~\cite{liu2022batch}, Condition~\ref{cond:TPVMm} is stronger than Condition~\ref{cond:TPM} and Condition~\ref{cond:VM}, as the former degenerates to the other two conditions by setting $\boldzeta=\boldsymbol{0}$ and $p_i^{\bmu,S}=1 \text{ for } i \in \tilde{S}$, respectively. Conversely, by applying the Cauchy-Schwartz inequality, one can verify that if a reward function is TPM $B_1$-bounded smooth, then it is 
% TPVM $(B_1\sqrt{K}/2, B_1, \lambda)$-bounded smooth for any $\lambda\le 2$ (or similarly VM $(B_1\sqrt{K}/2, B_1)$-bounded smooth). \siwei{Why not mention VM condition here?}

%\begin{remark}
For Condition~\ref{cond:TPM}, the key feature is that the parameter change in each base arm $i$ is modulated by the triggering probability $p_i^{\bmu,S}$. Intuitively, for base arm $i$ that is unlikely to be triggered/observed (small $p_i^{\bmu, S}$), Condition~\ref{cond:TPM} ensures that a large change in $\mu_i$ (due to insufficient observation) only causes a small change (multiplied by $p_i^{\bmu,S}$) in reward, which helps to save a $p_{\min}$ factor for non-contextual CMAB-T.
%\end{remark}

%\begin{remark}
For Condition~\ref{cond:VM}, intuitively if we ignore the denominator $(1-\mu_i)\mu_i$ of the leading $B_v$ term, the reward change would be $O(B_v \sqrt{K}\Delta )$ when the amount of parameter change $|\mu'_i-\mu_i|=\Delta$ for each arm $i$. This introduces a $O(\sqrt{K})$ factor reduction in the reward change and translates to a $O(K)$ improvement in the regret, compared with $O(B_1 K\Delta)$ reward change when applying the non-triggering version of Condition~\ref{cond:TPM} (i.e., $p_i^{\bmu,S}=1$ if $i \in \tilde{S}$ and $p_i^{\bmu,S}=0$ otherwise). However, for real applications, $B_1=\Theta(B_1 \sqrt{K})$ which cancels the $O(\sqrt{K})$ improvement.
To reduce $B_v$ coefficient, the leading $B_v$ term is modulated by the inverse of the variance $V_i=(1-\mu_i)\mu_i$, and thus allow applications to achieve a $B_v$ coefficient independent of $K$ (or at least $B_v=o(B_1\sqrt{K}$)), leading to significant savings in the regret bound for applications like PMC~\cite{liu2022batch}. The relation between Condition \ref{cond:TPM} and \ref{cond:VM} is generally not comparable, but compared with Condition~\ref{cond:TPM}'s non-triggering counterpart (i.e., 1-norm condition), Condition~\ref{cond:VM} is stronger. 
%\end{remark}

%\begin{remark}
%\label{rmk:TPVM}
Finally, for Condition~\ref{cond:TPVMm}, it combines both the triggering-probability modulation from Condition~\ref{cond:TPM} and the variance modulation from Condition~\ref{cond:VM}. The exponent $\lambda$ of $p_i^{\bmu,S}$ gives additional flexibility to trade-off between the strength of the condition and the regret, i.e., with a larger $\lambda$, one can obtain a smaller regret bound, while
with a smaller $\lambda$, the condition is easier to satisfy to include more applications. In general, Condition~\ref{cond:TPVMm} is stronger than Condition~\ref{cond:TPM} and Condition~\ref{cond:VM}, as the former degenerates to the other two conditions by setting $\boldzeta=\boldsymbol{0}$ and the fact that $p_i^{\bmu,S}\le1 \text{ for } i \in \tilde{S}$ and $p_i^{\bmu,S}=0$ otherwise, respectively. Conversely, by applying the Cauchy-Schwartz inequality, one can verify that if a reward function is TPM $B_1$-bounded smooth, then it is TPVM $(B_1\sqrt{K}/2, B_1, \lambda)$-bounded smooth for any $\lambda\le 2$ or similarly VM $(B_1\sqrt{K}/2, B_1)$-bounded smooth, respectively.
% More importantly, the $B_v$ term is further modulated by the inverse of the variance $V_i=(1-\mu_i)\mu_i$, and thus reduces their $B_v$ coefficient from $B_1\sqrt{K}/2$ to a coefficient independent of $K$, leading to significant savings in the regret bound of CMAB-T~\cite{liu2022batch}. 
% Another subtle yet critical point is that Condition~\ref{cond:TPVMm} strengthens the prior directional TPVM condition proposed by \citet{liu2022batch} (assuming $\boldeta, \boldzeta \in [0,1]^m$), which are necessary to guarantee desirable regret bounds for \ccmab-T as in \Cref{sec:va_c2mab}.
%\end{remark}

% \begin{remark}
%     For Condition~\ref{cond:VM}, it is a weaker version of Condition~\ref{cond:TPVMm}, by letting $p_i^{\bmu,S}=1$ for $i \in \tilde{S}$ and $0$ otherwise. 
%     % Condition~\ref{cond:VM} is stronger than Condition~\ref{cond:TPM}'s non-triggering counterpart (i.e., 1-norm condition).
% \end{remark}

In light of the above conditions that significantly advance the non-contextual CMAB-T, the goal of subsequent sections is to design algorithms and conduct analysis to derive the (improved) results for the contextual setting. And later in \Cref{sec:app}, we 
demonstrate how these conditions are applied to applications, such as CB and online IM, to achieve both theoretical and empirical improvements. Due to the space limit, the detailed proofs are included in the Appendix.
% \siwei{Maybe we can state some applications of these conditions here, and say "see details in Section 5".}
\section{Algorithm and Regret Analysis for \ccmab-T under the TPM Condition}\label{sec:c2mab}
\begin{algorithm}[t]
	\caption{C$^2$-UCB-T: \textbf{C}ontextual \textbf{C}ombinatorial \textbf{U}pper \textbf{C}onfidence \textbf{B}ound Algorithm for \ccmab-\textbf{T}}\label{alg: C2UCB}
			\resizebox{.96\columnwidth}{!}{
\begin{minipage}{\columnwidth}
	\begin{algorithmic}[1]
	    \STATE {\textbf{Input:}} Base arms $[m]$, dimension $d$, regularizer $\gamma$, failure probability $\delta=1/T$, offline oracle ORACLE. 
	   % \OUTPUT Budget allocation $\bk$.
	   \STATE \textbf{Initialize:} Gram matrix $\bG_{1}=\gamma\boldsymbol{I}$, vector $\bb_{1}=\boldsymbol{0}$. 
	   \FOR{$t=1, ...,T$ }
	   \STATE $\hat{\btheta}_{t}=\bG_{t}^{-1}\bb_{t}$. \label{line:c2mab_estimate}%\COMMENT{Compute the least-square estimator.}
	   %\STATE Construct the confidence set, one for decision, $C^{\theta}_t=\{\bnu: \norm{\bnu-\hat{\btheta}_t}_{\bG_{t}}^2 \le \beta\}$.
	   \FOR{$i \in [m]$}
	    \STATE $\bar{\mu}_{t,i} = \inner{\bphi_t(i), \hat{\btheta}_t} + \rho(\delta)\norm{\bphi_{t}(i)}_{\bG_{t}^{-1}}$.\label{line:c2_ucb}
	    %\COMMENT{UCB value for base arm $i$.}
	    % \STATE $\bar{\mu}_{t,i} =\min\{\bar{\mu}_{t,i},1\} $
	    \ENDFOR
	   %\STATE $G_t=\gamma \boldsymbol{I} + \sum_{\tau < t}\bphi(i)\bphi(i)^{\top}$.
	   %$\rho_{t,i}=\sqrt{\frac{6\hat{V}_{t-1, i} \log t}{T_{t-1, i}}} + \frac{9\log t}{T_{t-1, i}}$.
	   %\STATE Set UCB value $\bar{\mu}_{t,i}=\min \{\hat{\mu}_{t-1, i}+\rho_{t, i},1\}$.
	   %\STATE $S_t=\text{ORACLE}(C_t^{\theta}, \phi_t)$.  \qquad\qquad\qquad\qquad\qquad//$S_t=\arg\max_{S \in \cS, \bnu \in C_t^{\theta}} r(S;\bnu, \phi_t)$.
	   \STATE $S_t=\text{ORACLE}(\bar{\mu}_{t,1}, ..., \bar{\mu}_{t,m})$. %\COMMENT{Computational oracle using UCBs.}
	   \STATE Play $S_t$ and observe triggering arm set $\tau_t$ and observation set $(X_{t,i})_{i\in \tau_{t}}$.
	   % \FOR{$ i \in \tau_t$ } 
	   %\STATE $v_{t,i}=\max_{\bnu \in C^{v}_{t}}\inner{\bphi_t(i), \bnu}(1-\inner{\bphi_t(i), \bnu})$.
	   \STATE $\bG_{t+1}=\bG_{t}+\sum_{i \in \tau_t} \bphi_{t}(i)\bphi_{t}(i)^{\top}$.\label{line:c2_G}
	   \STATE $\bb_{t+1}=\bb_{t}+\sum_{i \in \tau_t} \bphi_{t}(i) X_{t,i}$.\label{line:c2_b} %\COMMENT{Update statistics using the triggered arms' observation.}
	   %\STATE Play $S_t$, which triggers arms $\tau_t \subseteq [m]$ with outcome $X_{t,i}$'s, for $i \in \tau_t$.\label{line:trigger_obs}.
	   % \STATE For every $i \in \tau_t$, update $T_{t,i}= T_{t-1, i}+1$, $\hat{\mu}_{t,i}= \hat{\mu}_{t-1,i}+(X_{t,i}-\hat{\mu}_{t-1,i})/T_{t,i}$,
	   %% $\hat{V}_{t, i}=\frac{T_{t-1, i}(\hat{V}_{t-1,i}+\hat{\mu}_{t-1}^2)+X_{t,i}^2}{T_{t, i}} -\hat{\mu}_{t,i}^2$. 
	   %$\hat{V}_{t, i}= \frac{T_{t-1, i}}{T_{t, i}}\left(\hat{V}_{t-1,i} + \frac{1}{T_{t, i}} \left(\hat{\mu}_{t-1,i} - X_{t,i}\right)^2\right) $.
	   % \label{line:trigger_upd}
	   %\ENDFOR
	    	   \ENDFOR
		\end{algorithmic}
           		\end{minipage}}
\end{algorithm}
Our proposed algorithm C$^2$-UCB-T (\Cref{alg: C2UCB}) is a generalization of the C$^3$-UCB algorithm originally designed for contextual combinatorial cascading bandits~\cite{li2016contextual}. Our main contribution is to show an improved regret bound by a factor of $1/p_{\min}$ under the 1-norm TPM condition.

Recall that we define the data about the history as $\cH_t=(\bphi_s, S_s, \tau_s, (X_{s,i})_{i\in \tau_s})_{s<t} \bigcup \bphi_t$.
Different from the CUCB algorithm~\cite{wang2017improving} that directly estimates the mean $\bmu_{t,i}$ for each arm, \Cref{alg: C2UCB} estimates the underlying parameter $\btheta^*$ via a ridge regression problem over the history data $\cH_t$.
More specifically, we estimate $\btheta^*$ by solving the following $\ell_2$-regularized least-square problem with regularization parameter $\gamma >0$:
\begin{equation}
 \resizebox{.91\linewidth}{!}{$
            \displaystyle
    \hat{\btheta}_t = \argmin_{\btheta \in \Theta} \sum_{s<t}\sum_{i \in \tau_{s}}(\inner{\btheta, \bphi_{s}(i)}-X_{s,i})^2 + \gamma \norm{\btheta}_2^2.$}
\end{equation}
The closed form solution is precisely the $\hat{\theta}_t$ calculated in line~\ref{line:c2mab_estimate}, where the Gram matrix $\bG_t=\sum_{s< t}\sum_{i \in \tau_{s}}\bphi_{s}(i) \bphi_{s}(i)^{\top}$ and the b-vector $\bb_t=\sum_{s< t}\sum_{i \in \tau_{s}}\bphi_{s}(i)X_{s,i}$ are computed in line~\ref{line:c2_G} and \ref{line:c2_b}. We claim that $\hat{\theta}_t$ is a good estimator of $\theta^*$ by bounding their difference via the following lemma, which is also used in \cite{qin2014contextual,li2016contextual}.

% \adam{Weird to call this "fact"..maybe Proposition?}
\begin{proposition}[Theorem 2, \cite{abbasi2011improved}]\label{lem:c2mab_good_est}
% Let $\rho(\delta)=\sqrt{\log\left(\frac{\det(\bG_{t})}{\gamma^d \cdot \delta^2}\right)}+\sqrt{\gamma}$, then with probability at least $1-\delta$, for all round $t\ge 0$, the estimate $\hat{\btheta}_t$ satisfies,
Let $\rho(\delta)=\sqrt{\log\left(\frac{(\gamma+KT/d)^d}{\gamma^d \cdot \delta^2}\right)}+\sqrt{\gamma}$, then with probability at least $1-\delta$, for all $t\in [T]$, $\norm{\hat{\btheta}_t-\btheta^*}_{\bG_{t}} \le \rho(\delta).$
% \begin{align}\label{eq:c2mab_est}
%     \norm{\hat{\btheta}_t-\btheta^*}_{\bG_{t}} \le \rho(\delta).
% \end{align}
\end{proposition}
% \siwei{The notation $\beta$ is used in the $(\alpha,\beta)$-oracle. Can we change a notation?}
% The proof of \cref{lem:c2mab_good_est} is based on the self-normalized theory and by considering $\norm{\btheta^*}_2\le 1$, $\norm{\bphi_t(i)}_2\le 1$, and $\det(\bG_{t+1})\le (\gamma+KT/d)^d$ in Theorem 2 of \citet{abbasi2011improved}. 
% \siwei{I'd like to write it as a Fact, since it is the same as some results in prior works. Also, we do not need to say anything about its proof, except that it is very important.}

Building on this, we construct an optimistic estimation of each arm's mean $\bar{\mu}_{t,i}$ in line~\ref{line:c2_ucb}, where $\rho(\delta)$ is in \Cref{lem:c2mab_good_est}, $\inner{\bphi_t(i), \hat{\btheta}_t}$ and $\rho(\delta)\norm{\bphi_{t}(i)}_{\bG_{t}^{-1}}$ are the empirical mean and confidence interval towards the direction $\bphi_t(i)$, respectively. As a convention, we clip $\bar{\mu}_{t,i}$ into $[0,1]$ if $\bar{\mu}_{t,i}> 1$ or $\bar{\mu}_{t,i}< 0$.
%that guides the exploration: the larger the confidence interval, the more bonus the agent gets to explore arm $i$, which reduces the uncertainty towards the direction $\bphi_t(i)$. 
% Similar to~\Cref{rmk:cmab-T_diff}, when $\bphi_t(i)=\be_i$, the confidence interval reproduces that of CUCB~\cite{wang2017improving}, where $\bG_t$ becomes a diagonal matrix $\text{diag}(T_{t,1},..., T_{t,m})$, and the confidence interval is proportional to $\sqrt{\log t/T_{t,i}}$, where $T_{t,i}$ is the number of times arm $i$ is triggered before round $t$. \siwei{Really? You have the $\gamma$ here, and hence it may not be exactly the same. Event if we let $\gamma = 0$, it also seems that the confidence bound is not the same as that one in CUCB, e.g., another factor of $\sqrt{d}$. Maybe we can remove this part...}

Thanks to \Cref{lem:c2mab_good_est}, we have the following lemma for the desired amount of the base arm level optimism,
\begin{restatable}{lemma}{lemccmabOptimism}
\label{lem:c2mab_optimism}
    With probability at least $1-\delta$, we have $\mu_{t,i}\le \bar{\mu}_{t,i}\le \mu_{t,i} + 2\rho(\delta)\norm{\bphi_t(i)}_{\bG_{t}^{-1}}$ for all $i \in [m], t \in [T]$. 
\end{restatable}
\begin{proof}
    See \Cref{apdx_sec:lemmas_thm1}.
\end{proof}
% \siwei{What do you mean by "when Lemma 1 holds"... Lemma 1 always holds, you can only say w.p. at least $1-\delta$, or when some event holds...}
% \begin{lemma}\label{lem:c2mab_optimism}
%     When \Cref{eq:c2mab_est} holds for $t-1$, we have $\mu_{t,i}\le \bar{\mu}_{t,i}\le \mu_{t,i} + 2\rho(\delta)\norm{\bphi_t(i)}_{\bG_{t}}$ for all $i \in [m]$. 
% \end{lemma}

After computing the UCB values $\bar{\bmu}_t$, the agent selects action $S_t$ via the offline oracle with $\bar{\bmu}_t$ as input.
Then base arms in $\tau_t$ are triggered, and the agent receives observation set $(X_{t,i})_{i\in \tau_{t}}$ as feedback to improve future decisions.
%
% Compared with~\cite{li2016contextual}, we generalize the cascading feedback to the general arm triggering process, whose outcomes are then used for future decision rounds. \siwei{Do not understand the last sentence...}

% \subsection{Analysis and Discussion}
\begin{theorem}\label{thm:c2mab_reg}
     For a \ccmab-T instance that satisfies monotonicity (Condition~\ref{cond:mono}) and TPM smoothness (Condition~\ref{cond:TPM}) with coefficient $B_1$, C$^2$-UCB-T (\Cref{alg: C2UCB}) with an $(\alpha,\beta)$-approximation oracle achieves an $(\alpha,\beta)$-approximate regret bounded by $ O\left(B_1(\sqrt{d\log (KT/\gamma)}+\sqrt{\gamma})\sqrt{KTd\log(KT/\gamma)}\right).$
\end{theorem}
% \siwei{Why there is no $\delta$ in this theorem?}

% \begin{remark}
\textbf{Discussion.} Looking at \Cref{thm:c2mab_reg}, we achieve an $O(B_1d\sqrt{KT}\log T)$ regret bound when $d \ll K \le m \ll T$, which is independent of the number of arms $m$ and the minimum triggering probability $p_{\min}$. Consider the combinatorial cascading bandits~\cite{li2016contextual} satisfying $B_1=1$ (see \Cref{sec:app}), our result improves the~\citet{li2016contextual} by a factor of $1/p_{\min}$. Consider the linear reward function~\cite{takemura2021near} without triggering arms (i.e., $p_i^{\bmu,S}=1$ for $i \in S$,  and $0$ otherwise), one can easily verify $B_1=1$ and our regret matches the lower bound ${\Omega}(d\sqrt{KT})$~\citet{takemura2021near} up to logarithmic factors.
% \end{remark}

% 1. regret bound independent of $m$ 2. compare with degenerate CMAB-T
% 3. lower bound 

\textbf{Analysis.}
%\begin{proof}[Proof Sketch]
Here we explain how to prove a regret bound that removes the $1/p_{\min}$ factor under the 1-norm TPM condition. The main challenge is that the mean vector $\bmu_t$ and the triggering probability $p_i^{\bmu_t,S}$ are dependent on \textit{time-varying} contexts $\bphi_t(i)$, so it is impossible to derive any meaningful concentration inequality or regret bound based on $T_{t,i}$, which is the number of times that arm $i$ is triggered, and has been used by the \textit{triggering group (TG)} technique~\cite{wang2017improving} to remove $1/p_{\min}$. To deal with this problem, we bypass the quantity $T_{t,i}$ and use the \textit{triggering-probability equivalence (TPE)} technique that equalizes $p_i^{\bmu_t, S}$ with $\E_t[\I\{i \in \tau_t\}]$, which in turn replaces the expected regret produced by all possible arms with the expected regret produced by $i \in \tau_t$ to avoid $p_{\min}$. To sketch our proof idea, we assume the oracle is deterministic with $\beta=1$ (the randomness of the oracle and $\beta<1$ are handled in \Cref{apdx_sec:TPM}), and let filtration $\cF_{t-1}$ be the history data $\cH_t$ (defined in \Cref{sec: problem setting}). 
% \siwei{Why not just define $\cH_t$ as $\cF_{t-1}$?} 
Denote $\E_t[\cdot]=\E[\cdot\mid \cF_{t-1}]$, the $t$-round regret $\E_t[\alpha\cdot r(S^*_t;\bmu_t)-r(S_t;\bmu_t)]{\le}\E_t[r(S_t;\bar{\bmu}_t)-r(S_t;\bmu_t)]$, based on  Condition~\ref{cond:mono}, \Cref{lem:c2mab_optimism} and definition of $S_t$. Then
% \noindent
% \resizebox{\columnwidth}{!}{
% \begin{minipage}{\columnwidth}
\begin{talign}\label{eq:TPE}
% &\E_t[\alpha\cdot r(S^*_t;\bmu_t)-r(S_t;\bmu_t)]\notag\\
%    &\overset{(a)}{\le}\E_t[\alpha\cdot r(S^*_t;\bar{\bmu}_t)-r(S_t;\bmu_t)]\overset{(b)}{\le} \E_t[r(S_t;\bar{\bmu}_t)-r(S_t;\bmu_t)]\notag\\
    &\E_t[r(S_t;\bar{\bmu}_t)-r(S_t;\bmu_t)]\overset{(a)}{\le}\E_t\left[\sum_{i \in \tilde{S}_t}B_1p_i^{\bmu_t,S_t}(\bar{\mu}_{t,i}-\mu_{t,i})\right]\notag\\
    &\overset{(b)}{=}\E\left[\sum_{i \in \tilde{S}_t}B_1\E_{\tau_t}[\I\{i \in \tau_t\}](\bar{\mu}_{t,i}-\mu_{t,i})\mid \cF_{t-1}\right]\notag\\
    &\overset{(c)}{=}\E_t\left[\sum_{i \in \tilde{S}_t}\I\{i \in \tau_t\}B_1(\bar{\mu}_{t,i}-\mu_{t,i})\right]\notag\\
    &\overset{(d)}{=}\E_t\left[\sum_{i \in \tau_t}B_1(\bar{\mu}_{t,i}-\mu_{t,i})\right],
\end{talign}
% \end{minipage}
% }

where (a) is by Condition~\ref{cond:TPM}, (b) is because $\bar{\mu}_{t,i}, \mu_{t,i}, S_t$ are $\cF_{t-1}$ measurable so that the only randomness is from triggering set $\tau_t$ and we can substitute $p_i^{\bmu_t, S_t}$ with event $\I\{i \in \tau_t\}$ under expectation, $(c)$ is by absorbing the expectation over $\tau_t$ to $\E_t$, and $(d)$ is a change of notation. 
After applying the TPE, we only need to bound the regret produced by $i \in \tau_t$. Hence %without considering the arm-triggering. Hence
% i.e., Reg$(T)\le\E\left[\sum_{t \in [T]}\E_t\left[\sum_{i \in \tau_t}B_1(\bar{\mu}_{t,i}-\mu_{t,i})\right]\right].$ Then its RHS $\le 2B_1\rho(\delta)\E\left[\sum_{t \in [T]}\sum_{i \in \tau_t}\norm{\bphi_t(i)}_{\bG_{t}^{-1}}\right]\le 2B_1\rho(\delta) \E\left[\sqrt{KT\sum_{t \in [T]}\sum_{i \in \tau_t}\norm{\bphi_t(i)}^2_{\bG_{t}^{-1}}}\right]\le {O}(B_1d\sqrt{KT}\log T)$,
% where the first inequality is by~\Cref{lem:c2mab_optimism}, the second inequality by Cauchy Schwarz inequality over both $i$ and $t$, and the last is by the ellipsoidal potential lemma \Cref{apdx_lem:ccmab_potential} in the Appendix.
\begin{talign}\label{eq:c2mab_reg}
&\text{Reg}(T){\le} \E\left[\sum_{t \in [T]}\E_t\left[\sum_{i \in \tau_t}B_1(\bar{\mu}_{t,i}-\mu_{t,i})\right]\right]\notag\\
&\overset{(a)}{\le} \E\left[\sum_{t \in [T]}\E_t\left[\sum_{i \in \tau_t}2B_1 \rho(\delta)\norm{\bphi_t(i)}_{\bG_{t}^{-1}}\right]\right]\notag\\
&\overset{(b)}{\le} 2B_1\rho(\delta) \E\left[\sqrt{KT\sum_{t \in [T]}\sum_{i \in \tau_t}\norm{\bphi_t(i)}^2_{\bG_{t}^{-1}}}\right]\notag\\
&\overset{(c)}{\le}{O}(B_1d\sqrt{KT}\log T).
\end{talign}
where (a) follows from~\Cref{lem:c2mab_optimism}, (b) is by Cauchy Schwarz inequality over both $i$ and $t$, and (c) is by the ellipsoidal potential lemma (\Cref{apdx_lem:ccmab_potential}) in the Appendix.

% \begin{align}\label{eq:c2mab_reg}
% &\text{Reg}(T)\overset{(a)}{=}\E\left[\sum_{t \in [T]}\E_t[\alpha\cdot r(S^*_t;\bmu_t)-r(S_t;\bmu_t)]\right]\notag\\
% &\overset{(b)}{\le} \E\left[\sum_{t \in [T]}\E_t\left[\sum_{i \in \tau_t}B_1(\bar{\mu}_{t,i}-\mu_{t,i})\right]\right]\notag\\
% &\overset{(c)}{\le} \E\left[\sum_{t \in [T]}\E_t\left[\sum_{i \in \tau_t}B_1 \rho(\delta)\norm{\bphi_t(i)}_{\bG_{t}^{-1}}\right]\right]\notag\\
% &\overset{(d)}{=}  B_1\rho(\delta)\E\left[\sum_{t \in [T]}\sum_{i \in \tau_t}\norm{\bphi_t(i)}_{\bG_{t}^{-1}}\right]\notag\\
% &\overset{(e)}{\le}  B_1\rho(\delta) \E\left[\sqrt{KT\sum_{t \in [T]}\sum_{i \in \tau_t}\norm{\bphi_t(i)}^2_{\bG_{t}^{-1}}}\right] \notag\\
% &\overset{(f)}{\le}{O}(B_1d\sqrt{KT}\log T).
% \end{align}
% where (a) follows from the definition and the tower rule, (b) by the TPE~\Cref{eq:TPE}, (c) by~\Cref{lem:c2mab_optimism}, (d) by tower rule, (e) by Cauchy Schwarz inequality, and (f) by the ellipsoidal potential lemma \Cref{apdx_lem:ccmab_potential_new} in the Appendix.
\begin{remark}
In addition to the general \ccmab-T setting, the TPE technique can also replace the more involved TG technique~\cite{wang2017improving} for CMAB-T. Such replacement can save an unnecessary union bound over the group index, which in turn reproduce Theorem 1 of \citet{wang2017improving} under Condition~\ref{cond:TPM}, and improve Theorem 1 of \citet{liu2022batch} under Condition~\ref{cond:TPVMm} by a factor of $O(\sqrt{\log T})$, see \Cref{apdx_sec:NC_C2MAB} for details.
\end{remark}

\section{Variance-Adaptive Algorithm and Analysis for \ccmab-T under VM/TPVM Condition}\label{sec:va_c2mab}

\begin{algorithm}[t]
	\caption{VAC$^2$-UCB: \textbf{V}ariance-\textbf{A}daptive \textbf{C}ontextual \textbf{C}ombinatorial \textbf{U}pper \textbf{C}onfidence \textbf{B}ound Algorithm}\label{alg:va_cucb}
			\resizebox{.96\columnwidth}{!}{
\begin{minipage}{\columnwidth}
	\begin{algorithmic}[1]
	    \STATE {\textbf{Input:}} Base arms $[m]$, dimension $d$, regularizer $\gamma$, failure probability $\delta=1/T$, offline oracle ORACLE.
	   % \OUTPUT Budget allocation $\bk$.
	   \STATE \textbf{Initialize:} Gram matrix $\bG_{1}=\gamma\boldsymbol{I}$, regressand $\bb_{1}=\boldsymbol{0}$. 
	   \FOR{$t=1, ...,T$ }
	   \STATE $\hat{\btheta}_{t}=\bG_{t}^{-1}\bb_{t}$. %\COMMENT{Compute empirical estimator.}
	   %\STATE Construct two confidence sets, one for decision and one for the variance, $C^{\theta}_t=\{\bnu: \norm{\bnu-\hat{\btheta}_t}_{\bG_{t}}^2 \le \beta\}$, $C^{v}_t=\{\bnu: \norm{\bnu-\hat{\btheta}_t}_{\bG_{t}}^2 \le 4\beta\}$.
	   \FOR{$i \in [m]$}
	   \STATE $\bar{\mu}_{t,i} = \inner{\bphi_t(i), \hat{\btheta}_t} + 2\rho(\delta)\norm{\bphi_{t}(i)}_{\bG_{t}^{-1}}$\label{line:va_ucb}  %\COMMENT{UCB value for arm $i$.}
	   % \STATE $\bar{\mu}_{t,i}=\min \{\bar{\mu}_{t,i},1\}$.
	   \STATE $\ubar{\mu}_{t,i}=\inner{\bphi_t(i), \hat{\btheta}_t} - 2\rho(\delta)\norm{\bphi_{t}(i)}_{\bG_{t}^{-1}}$\label{line:va_lcb} %\COMMENT{LCB value for arm $i$.}
	   % \STATE $\ubar{\mu}_{t,i}=\max\{\ubar{\mu}_{t,i},0\}$.
	   %\STATE Compute the optimistic variance by solving $\max_{\bnu \in C^{v}_t}{\inner{\bphi_t(i), \bnu}(1-\inner{\bphi_t(i), \bnu})}$ 
    \STATE Set the optimistic variance $\bar{V}_{t,i}$ as \Cref{eq:var_upper}. \label{line:var_opt} %\COMMENT{Optimistic variance for true $V_{t,i}$.}
% \STATE Let $ \tilde{\bnu}_{t}(i), \tilde{\mu}_{t,i}$ are parameters that achieve $v_{t,i}$.
	   \ENDFOR
	   \STATE $S_t=\text{ORACLE}(\bar{\mu}_{t,1}, ..., \bar{\mu}_{t,m})$. %\COMMENT{Computational oracle using UCBs.}
	   \STATE Play $S_t$ and observe triggering arm set $\tau_t$ and observation set $(X_{t,i})_{i\in \tau_{t}}$.
	   % \FOR{$ i \in \tau_t$ }\COMMENT{Update statistics.}\label{line:update_G}
	   \STATE $\bG_{t+1}=\bG_{t}+\sum_{i \in \tau_t} \bar{V}_{t,i}^{-1}\bphi_{t}(i)\bphi_{t}(i)^{\top}$.\label{line:update_G}
	   \STATE $\bb_{t+1}=\bb_{t}+\sum_{i \in \tau_{t}} \bar{V}_{t,i}^{-1}\bphi_{t}(i) X_{t,i}$.\label{line:update_b}%\COMMENT{Update statistics.}
	   \ENDFOR
	    	  % \ENDFOR
		\end{algorithmic}
           		\end{minipage}}
\end{algorithm}

In this section, we propose a new variance-adaptive algorithm VAC$^2$-UCB (\Cref{alg:va_cucb}) to further remove the $O(\sqrt{K})$ factor and achieve $\tilde{O}(B_v d\sqrt{T})$ regret bound for applications satisfying stronger VM/TPVM conditions.

% \siwei{Mention algorithm name here?}

Different from \Cref{alg: C2UCB}, VAC$^2$-UCB leverages the second-order statistics (i.e., variance) to speed up the learning process. 
To get some intuition, we first assume the variance $V_{s,i}=\text{Var}[X_{s,i}]$ for each base arm $i$ at round $s$ is known in advance. 
In this case, VAC$^2$-UCB adopts the \textit{weighted ridge-regression} to learn the parameter $\btheta^*$:
\begin{equation}\label{eq:var_opt}
 \resizebox{.91\linewidth}{!}{$
            \displaystyle
    \hat{\btheta}_t = \argmin_{\btheta \in \Theta} \sum_{s<t}\sum_{i \in \tau_s}(\inner{\btheta, \bphi_s(i)}-X_{s,i})^2/V_{s,i} + \gamma \norm{\btheta}_2^2,
    $}
\end{equation}
where the first term is ``weighted" by the true variance $V_{s,i}$. The closed-form solution of the above estimator is $\hat{\btheta}_{t}=\bG_{t}^{-1}\bb_{t}$ where the Gram matrix $\bG_t=\sum_{s< t}\sum_{i \in \tau_{s}}V_{s,i}^{-1}\bphi_{s}(i) \bphi_{s}(i)^{\top}$ and the b-vector $\bb_t=\sum_{s< t}\sum_{i \in \tau_{s}}V_{s,i}^{-1}\bphi_{s}(i)X_{s,i}$, which enjoys the similar form (but uses different weights $\bar{V}_{s,i}$) of  line~\ref{line:update_G} and line~\ref{line:update_b}.

The intuition of using the inverse of $V_{s,i}$ to re-weight the observation is that: the smaller the variance, the more accurate the observation $(\bphi_t(i), X_{t,i})$ is, and thus more important for the agent to learn unknown $\btheta^*$.
%
%Hence 
%
%The intuition of using the variance in this way is from the exploitation perspective: the smaller the variance, the more accurate the observation $(\bphi_t(i), X_{t,i})$ is, and thus more important for the agent to learn unknown $\btheta^*$. So we use the inverse of $V_{s,i}$ to re-weight the observation.
In fact, the above estimator $\hat{\btheta}_t$ is \textit{closely related} to the best linear unbiased estimator (BLUE)~\cite{henderson1975best}. Concretely, in the literature of linear regression, Equation (5) is the lowest variance estimator of $\btheta^*$ among all unbiased linear estimators, when the regularization term $\gamma=0$, $V_{s,i}$ are the true variance proxy of outcomes $(X_{s,i})_{s<t, i \in \tau_s}$ and the context sequence $(\bphi_{s}(i))_{s<t, i \in \tau_s}$ follows the fixed design in Equation (5).

For our \ccmab-T setting, one new challenge arises since the variance $V_{s,i}=\mu_{s,i}(1-\mu_{s,i})$ is not known a priori. Inspired by \cite{lattimore2015linear,zhou2021nearly}, we construct an optimistic estimation $\bar{V}_{s,i}$ to replace the true variance $V_{s,i}$ in \Cref{eq:var_opt}. Indeed, we construct $\bar{V}_{t,i}$ by solving the optimal value for the problem $\max_{\mu\in [\ubar{\mu}_{t,i}, \bar{\mu}_{t,i}]}\mu(1-\mu)$, whose closed form solution immediately follows from the equation below,
\begin{align}\label{eq:var_upper}
    % \STATE If variance upper bound $\bar{\bV}$ is available: $\bar{V}_{t,i}=\bar{V}_{i}$; Otherwise, set the optimistic variance as\begin{equation}\label{eq:var_upper}
 \bar{V}_{t,i}= \begin{cases}
(1-\bar{\mu}_{t,i})\bar{\mu}_{t,i}, &\text{if $\bar{\mu}_{t,i} \le \frac{1}{2}$}\\
(1-\ubar{\mu}_{t,i})\ubar{\mu}_{t,i}, &\text{if $\ubar{\mu}_{t,i} \ge \frac{1}{2}$}\\
\frac{1}{4}, &\text{otherwise}
\end{cases}
\end{align}
% \Cref{eq:var_upper} in line~\ref{line:var_opt}.
where $\bar{\mu}_{t,i}$ and $\ubar{\mu}_{t,i}$ are UCB and LCB values to be introduced later. Notice that with high probability the true $\mu_{t,i}$ lies within LCB and UCB values and as they are getting more accurate, the optimistic variance $V_{t,i}$ is also approaching the true variance $V_{t,i}$.

To guarantee $\hat{\btheta}_t$ is a good estimator, we prove a new lemma (similar to \Cref{lem:c2mab_good_est}) to guarantee the concentration bound of $\btheta_t$ but in face of the unknown variance. Note that the sentinel work \citet{lattimore2015linear} proves a similar concentration bound, the difference is that we have multiple arms triggered in each round instead of a single arm. To address this, we replaced the original concentration bound with the new one below that has an extra $K^4$ factor in $N$, which finally results in $\log K$ factor in the confidence radius $\rho(\delta)$. 

\begin{lemma}\label{lem:va_c2mab_good_est}
Let $\gamma>0, N=(4d^2K^4T^4)^d$ so that $\rho(\delta)=\left(1+\sqrt{\gamma}+4\sqrt{\log\left(\frac{6TN}{\delta}\log\left(\frac{3TN}{\delta}\right)\right)}\right)$. We have for all $t\le T$, with probability at least $1-\delta$, $\norm{\hat{\btheta}_t-\btheta^*}_{\bG_{t}}^2\le \rho(\delta)$.
\begin{proof}
    See \Cref{apdx_sec:TPVM_key_lem}.
\end{proof}
% $\norm{\hat{\btheta}-\btheta_t}_{\bG_{t}}^2\le \beta$
% \begin{align}\label{eq:va_c2mab_optimism}
%     \norm{\hat{\btheta}_t-\btheta^*}_{\bG_{t}}^2\le \beta
% \end{align} 
\end{lemma}
% \siwei{Why not $\rho(\delta)$ here?}

Building on this lemma, we construct $\bar{\mu}_{t,i}$ as an upper bound of $\bmu_{t,i}$ in line~\ref{line:va_ucb}, and $\ubar{\mu}_{t,i}$ as a lower bound of $\bmu_{t,i}$ in line~\ref{line:va_lcb}, based on our variance-adaptive $\hat{\btheta}_t$, $\bG_t$. Note that the doubling of the radius $2\rho(\delta)$ instead of using $\rho(\delta)$ in \Cref{lem:va_c2mab_good_est} is purely for the correctness of our technical analysis. As a convention, we clip $\bar{\mu}_{t,i}, \ubar{\mu}_{t,i}$ into $[0,1]$ if they are above 1 or below 0.
\begin{lemma}\label{lem:va_c2mab_optimism}
    With probability at least $1-\delta$, we have $\mu_{t,i}\le \bar{\mu}_{t,i}\le \mu_{t,i} + 3\rho(\delta)\norm{\bphi_t(i)}_{\bG_{t}^{-1}}$, and $\mu_{t,i}\ge \ubar{\mu}_{t,i}\ge \mu_{t,i} - 3\rho(\delta)\norm{\bphi_t(i)}_{\bG_{t}^{-1}}$ for all $i \in [m]$. 
\end{lemma}
\begin{proof}
This lemma follows from the similar derivation of \Cref{lem:c2mab_optimism}, where we have different definitions of $\bar{\mu}_{t,i}, \ubar{\mu}_{t,i}$ and the concentration now relies on  \Cref{lem:va_c2mab_good_est}.   
\end{proof}
% \siwei{Similar question with that after Lemma 2.}
After the agent plays $S_t$, the base arms in $\tau_t$ are triggered, and the agent receives observation set $(X_{t,i})_{i\in \tau_{t}}$ as feedback. These observations (reweighted by optimistic variance $\bar{V}_{t,i}$) are then used to update $\bG_t$ and $\bb_t$ for future rounds.

% Intuitively, if we know an upper bound of $\sigma_i$,  which we denote as $\bar{\sigma}_i$, and we only select one arm each round (i.e., $K=1$), then $\hat{\btheta}_t$ is the closed form solution of the following \textit{weighted ridge regression} problem:
% \begin{align}
%     \hat{\btheta}_t = \arg\min_{\btheta \in \R^d} \sum_{\tau=1}^{t}\sum_{i \in S_{\tau}}(\inner{\btheta, \bphi_\tau(i)}-X_{\tau,i})/\bar{\sigma}_i^2 + \lambda \norm{\btheta}_2^2.
% \end{align}
% The term ``weighted" corresponds the variance upper bound $\bar{\sigma}$. The above estimator is known as the best linear unbiased estimator (BLUE)~\cite{henderson1975best}. Concretely, $\hat{\btheta}_t$ is the lowest variance estimator of $\btheta$ among all unbiased linear estimators.

% \begin{remark}
% The intuition of using variance is from the exploitation perspective: the smaller the variance, the more accurate the data point is, and thus more important for the agent to learn unknown $\btheta$. So we use the inverse of $\bar{V}_{t,i}$ in line \ref{line:update_G} and \ref{line:update_b} to re-weight the observation.
% \end{remark}

% For our problem, we have three key ingredients: (1) We use $v_{t,i}$ is an estimation of $\bar{\sigma}_i$ at round $t$ instead of assuming a known upper bound $\bar{\sigma}_i$; (2) At each round, we select a combinatorial action which consists of multiple arms, i.e., $K > 1$. (3) We incorporate VM condition (\cref{cond:TPVMm}) to remove the factor of $\sqrt{K}$.

\begin{table*}[t]
	\caption{Summary of the coefficients, regret  bounds and improvements for various applications.}	\label{tab:app}	
         \centering 
		\resizebox{1.96\columnwidth}{!}{
	\centering
	\begin{threeparttable}
	\begin{tabular}{|ccccc|}
		\hline
		\textbf{Application}&\textbf{Condition}& \textbf{$(B_v, B_1, \lambda)$} & \textbf{Regret}& \textbf{Improvement}\\
	\hline
  Online Influence Maximization \cite{wen2017online} & TPM &$(-,|V|,-)$   & $O(d|V|\sqrt{|E|T}\log T)$ & $\tilde{O}(\sqrt{|E|})$\\
	    \hline
		Disjunctive Combinatorial Cascading Bandits~\cite{li2016contextual} &  TPVM & $(1,1,1)$ & $O(d\sqrt{T}\log T)$ & $\tilde{O}(\sqrt{K}/p_{\min})$ \\
			\hline
% 			\rowcolor{LightGray}
		Conjunctive Combinatorial Cascading Bandits~\cite{li2016contextual} & TPVM & $(1,1,1)$ & $O( d\sqrt{T}\log T)$ & $\tilde{O}(\sqrt{K} /r_{\max})$\\
			\hline
   Linear Cascading Bandits~\cite{vial2022minimax}$^*$  & TPVM & $(1,1,1)$ & $O(d\sqrt{T}\log T)$ & $\tilde{O}(\sqrt{K/d})$ \\
		   \hline
				% \rowcolor{LightGray}
% 			Influence Maximization on TPG \cite{wang2017improving} & TPVM & $(\sqrt{2}|V|,|V|,1)$ & $O( \sum_{i \in [m]}\log \frac{|E|}{\Delta_{\min}}\frac{|V|^2\log |E| \log T}{\Delta_i^{\min}})$  & $O(|E|/(\log |E|\log \frac{|E|}{\Delta_{\min}}))$\\
				% \hline

      Multi-layered Network Exploration~\citep{liu2021multi} & TPVM & $(\sqrt{1.25|V|},1,2)$ & $O(d\sqrt{|V|T}\log T)$ & $\tilde{O}(\sqrt{n}/p_{\min})$\\
			\hline
	   % 	\rowcolor{LightGray}
	Probabilistic Maximum Coverage \cite{chen2013combinatorial}$^{**}$& VM & $(3\sqrt{2|V|},1,-)$ & $O(d\sqrt{|V|T}\log T)$ & $\tilde{O}(\sqrt{k})$\\
	    \hline
     	   
	\end{tabular}
	  \begin{tablenotes}[para, online,flushleft]
	\footnotesize%\smallskip
	\item[]\hspace*{-\fontdimen2\font} $|V|, |E|, n, k, L$ denotes the number of target nodes, the number of edges that can be triggered by the set of seed nodes, the number of layers, the number of seed nodes and the length of the longest directed path, respectively;
  \item[]\hspace*{-\fontdimen2\font} $K$ is the length of the ordered list, $r_{\max}=\alpha\cdot\max_{t\in [T], S\in \cS} r(S;\bmu_t)$;
   \item[]\hspace*{-\fontdimen2\font}$^*$ A special case of disjunctive combinatorial cascading bandits. 
 \item[]\hspace*{-\fontdimen2\font}$^{**}$ This row is for \ccmab~application and the rest of rows are for \ccmab-T applications. 
% 	\item[]\hspace*{-\fontdimen2\font}$^\ddagger$ $|V|, |E|, n, |L|$ denotes the number of layers, the number of longest;
% 	\item[]\hspace*{-\fontdimen2\font}$^\ddagger$ $L$ is the number of ;
% 	\item[]\hspace*{-\fontdimen2\font}$^{\S}$ This work gives sufficient smoothness condition with factor $\gamma_g$ and translates to $B_v=3\sqrt{2}\gamma_g$ in our setting.
	\end{tablenotes}
			\end{threeparttable}
	}
\end{table*}

\subsection{Results and Analysis under VM condition}
We first show a regret bound for VAC$^2$-UCB that is independent of batch size $K$ when the VM condition holds.
\begin{theorem}\label{thm:ccmab_reg_VM}
    For a \ccmab-T instance that satisfies monotonicity (Condition~\ref{cond:mono}) and VM smoothness (Condition~\ref{cond:VM}) with coefficient $(B_v,B_1)$, VAC$^2$-UCB (\Cref{alg:va_cucb}) with an $(\alpha,\beta)$-approximation oracle achieves an $(\alpha,\beta)$-approximate regret bounded by $ O\left(\frac{B_v}{\sqrt{p_{\min}}}(\sqrt{d\log (KT/\gamma)}+\sqrt{\gamma})\sqrt{Td\log(KT/\gamma)}\right).$
\end{theorem}
% \begin{remark}
\textbf{Discussion.} Looking at \Cref{thm:ccmab_reg_VM}, we achieve an $O(B_vd\sqrt{T}\log T/\sqrt{p_{\min}})$ regret bound when $d \ll K \le m \ll T$. For combinatorial cascading bandits~\cite{li2016contextual} with $B_v=1$, our regret is independent of $m, K$ and improves~\citet{li2016contextual} by a factor $O(\sqrt{K/p_{\min}})$.
% \end{remark}

% \begin{remark} 
 In addition to the general \ccmab-T setting, one can verify that for non-triggering \ccmab, $p_{\min}=1$, and we obtain the batch-size independent regret bound $O(B_vd\sqrt{T}\log T)$. Recall $B_v=O(B_1\sqrt{K})$ for any \ccmab-T instances, so our regret bound reproduces $O(B_1d\sqrt{KT}\log T)$, and thus matches the similar lower bound~\cite{takemura2021near} for the linear reward functions. For the more interesting non-linear reward function with $B_v=o(B_1\sqrt{K})$, our regret improves non-variance-adaptive algorithm C$^2$UCB, whose regret is $O(B_1d\sqrt{KT}\log T)$~\cite{qin2014contextual,takemura2021near}.
% \end{remark}

\noindent\textbf{Analysis.} At a high level, the improvement of $\sqrt{K}$ comes from the VM condition and the optimistic variance, which together save the use of Cauchy-Schwarz inequality that generates a $O(\sqrt{K})$ factor in the step (b) of \Cref{eq:c2mab_reg}. In order to leverage the variance information, we decompose the regret into term (\rom{1}) and (\rom{2}),
\noindent
\resizebox{.99\columnwidth}{!}{
\begin{minipage}{\columnwidth}
\begin{talign}
&\text{Reg}(T)\le \E\left[\sum_{t=1}^T r(S_t; \bar{\bmu}_t)- r(S_t; \bmu_t)\right]\notag\\
&\le \E[\sum_{t=1}^T \underbrace{\abs{r(S_t; \bar{\bmu}_t)- r(S_t; \tilde{\bmu}_t)}}_{\text{(\rom{1})}}+\underbrace{\abs{r(S_t; \bmu_t)- r(S_t; \tilde{\bmu}_t)}}_{\text{(\rom{2})}}]\label{eq:regret_decompose},
% &\le B_v\E\left[\sum_{t=1}^T\left(\sqrt{\sum_{i \in [m]}\frac{(\bar{\mu}_{t,i}-\tilde{\mu}_{t,i})^2}{\bar{V}_{t,i}}}+\sqrt{\frac{(\mu_{t,i}-\tilde{\mu}_{t,i})^2}{\bar{V}_{t,i}}}\right)\right]\\
% &\le  B_v\E\left[\sqrt{T\sum_{t=1}^T\frac{(\bar{\mu}_{t,i}-\tilde{\mu}_{t,i})^2}{\bar{V}_{t,i}}}+\sqrt{T\sum_{t=1}^T\frac{(\mu_{t,i}-\tilde{\mu}_{t,i})^2}{\bar{V}_{t,i}}}\right]\\
% &\le B_v\sqrt{T}\sqrt{\E\left[\sum_{t=1}^T\frac{(\bar{\mu}_{t,i}-\tilde{\mu}_{t,i})^2}{\bar{V}_{t,i}}\right]}+\sqrt{\E\left[\sum_{t=1}^T\frac{(\mu_{t,i}-\tilde{\mu}_{t,i})^2}{\bar{V}_{t,i}}\right]}\\
\end{talign}
\end{minipage}}
where $\tilde{\bmu}_{t}$ is the vector whose $i$-th entry is the maximizer that achieves optimistic variance $\bar{V}_{t,i}$, i.e., $\tilde{\mu}_{t,i}=\argmax_{\mu\in [\ubar{\mu}_{t,i}, \bar{\mu}_{t,i}]}\mu(1-\mu)$.
Now we show a sketched proof to bound the term  (\rom{1}) and one can bound the term {(\rom{2})} similarly.
\noindent
\resizebox{.99\columnwidth}{!}{
\begin{minipage}{\columnwidth}
\begin{talign}
&\E\left[\sum_{t\in[T]}\text{(\rom{1})}\right] \overset{(a)}{\le} B_v\E\left[\sum_{t=1}^T\sqrt{\sum_{i \in \tilde{S}_t}{(\bar{\mu}_{t,i}-\tilde{\mu}_{t,i})^2}/{\bar{V}_{t,i}}}\right]\notag\\
& \overset{(b)}{\le}  \frac{B_v}{\sqrt{p_{\min}}}\cdot \E\left[\sum_{t=1}^T\sqrt{\sum_{i \in \tilde{S}_t}p_{i}^{\bmu_t,S_t}{(\bar{\mu}_{t,i}-\tilde{\mu}_{t,i})^2}/{\bar{V}_{t,i}}}\right]\notag\\
% & \le  \frac{B_v}{\sqrt{p_{\min}}}\cdot \E\left[\sqrt{T\sum_{t=1}^T\sum_{i \in \tilde{S}_t}p_{i}^{\bmu_t,S_t}{(\bar{\mu}_{t,i}-\tilde{\mu}_{t,i})^2}/{\bar{V}_{t,i}}}\right]\notag\\
& \overset{(c)}{\le}  \frac{B_v}{\sqrt{p_{\min}}}\cdot \sqrt{T\E\left[\sum_{t=1}^T\sum_{i \in \tilde{S}_t}p_{i}^{\bmu_t,S_t}{(\bar{\mu}_{t,i}-\tilde{\mu}_{t,i})^2}/{\bar{V}_{t,i}}\right]}\notag\\
% & \overset{(d)}{=}  \frac{B_v}{\sqrt{p_{\min}}}\cdot \sqrt{T\E\left[\sum_{t=1}^T\sum_{i \in \tau_t}{(\bar{\mu}_{t,i}-\tilde{\mu}_{t,i})^2}/{\bar{V}_{t,i}}\right]}\notag\\
&\overset{(d)}{\le} \frac{B_v}{\sqrt{p_{\min}}}\cdot \sqrt{T\E\left[\sum_{t=1}^T\sum_{i \in \tau_t}{(6\rho(\delta)\norm{\bphi_t(i)}_{\bG_{t}^{-1}})^2}/{\bar{V}_{t,i}}\right]}\notag\\
&\overset{(e)}{\le} O(B_vd\sqrt{T\log T}/\sqrt{p_{\min}}),\label{eq:regret_derive_VM}
\end{talign}
\end{minipage}}
where (a) is by Condition~\ref{cond:VM}, (b) is by the definition of $p_i^{\bmu_t,S_t}\ge p_{\min}$ for $i \in \tilde{S}_t$, (c) is by Cauchy–Schwarz over $t$ and the Jensen's inequality, (d) follows from the TPE and \Cref{lem:va_c2mab_optimism}, (e) follows from \Cref{apdx_lem:ccmab_potential_new}.

\subsection{Results and Analysis under TPVM Condition}
Next, we show that VAC$^2$-UCB can achieve regret bounds that remove the $O(\sqrt{K})$ and $O(1/\sqrt{p_{\min}})$ factor for applications satisfying the stronger TPVM conditions.

We first introduce a mild condition over the triggering probability (which is similar to Condition~\ref{cond:TPM}) to give our regret bounds and analysis.

\begin{condition}[\textbf{1-norm TPM Bounded Smoothness for Triggering Probability}]\label{cond:TPM_P}
% We say that a CMAB-T instance satisfies TPM $B_1$-bounded smoothness condition, if for any action $S \in \cS$, any distribution $D,D'$ with mean vector $\bmu \in (0,1)^m$, we have $\frac{\partial r(S;\bmu)}{\partial \mu_i } \frac{1}{p_{i}^{D,S}}\le B_1 $ for $i \in \tilde{S}$ and $\frac{\partial r(S;\bmu)}{\partial \mu_i }=0$ for $i \in [m] \backslash \tilde{S}$. 
We say that a \ccmab-T problem instance satisfies the triggering probability modulated $B_p$-bounded smoothness condition over the triggering probability, if for any action $S \in \cS$, any mean vectors $\bmu, \bmu' \in [0,1]^m$, and any arm $i\in[m]$, we have $|p_i^{\bmu', S}-p_i^{\bmu, S}|\le B_p\sum_{j \in [m]}p_{j}^{\bmu,S}|\mu_j-\mu'_j|$.
%\siwei{seems no formal definition of this $p_{i}^{D,S}$?} 
\end{condition}

Now we state our main theorem as follows.
\begin{theorem}\label{thm:ccmab_reg_TPVM_lambda}
    For a \ccmab-T instance, when its reward function satisfies monotonicity (Condition~\ref{cond:mono}) and TPVM smoothness (Condition~\ref{cond:TPVMm}) with coefficient $(B_v,B_1, \lambda)$, and its triggering probability $p_i^{\bmu,S}$ satisfies 1-norm TPM smoothness with coefficient $B_p$ (Condition~\ref{cond:TPM_P}),
    if $\lambda \ge 2$, then VAC$^2$-UCB (\Cref{alg:va_cucb}) with an $(\alpha,\beta)$-approximation oracle achieves an $(\alpha,\beta)$-approximate regret bounded by 
    \begin{align}
        O\left(B_v d\sqrt{T}\log T + {B_vB_p\sqrt{K}(d\log T)^2}/{\sqrt{p_{\min}}}\right),
    \end{align}
    and if $\lambda \ge 1$, then VAC$^2$-UCB (\Cref{alg:va_cucb}) achieves an $(\alpha,\beta)$-approximate regret bounded by
    \begin{align}
    O\left(B_v d\sqrt{T}\log T + {B_v\sqrt{B_p}(KT)^{1/4}(d\log T)^{3/2}}/{\sqrt{p_{\min}}}\right).
    \end{align}    
\end{theorem}

\textbf{Discussion.}
The leading term of \Cref{thm:ccmab_reg_TPVM_lambda} is $O(B_vd\sqrt{T}\log T)$ when $d\ll K\le m\ll T$, which removes the $1/\sqrt{p_{\min}}$ factor compared with \Cref{thm:ccmab_reg_VM}. Also, notice that \Cref{thm:ccmab_reg_TPVM_lambda} relies on an additional $B_p$-smoothness condition over the triggering probability. However, we claim that this condition is mild and almost always satisfies with $B_p=B_1$ for applications considered in this paper (see \Cref{apdx_sec:apps}). %The purpose of this condition is to handle the mismatch of the triggering probability when using the TPE, which will be introduced shortly after.
% Finally, when there exists some prior knowledge about the variance bound $\bar{V}_i$, s.t. $\Var[X_{t,i}]\le \bar{V}_i\le C_1\Var[X_{t,i}]$, we can achieve $O(B_vd\sqrt{T}\log T)$ bound for general $\lambda \ge 1$, without assuming the $B_p$ smoothness condition. We refer interested readers to the Appendix for its details.
% \end{remark}

\textbf{Analysis.} We use the regret decomposition of \Cref{eq:regret_decompose} to the same term (\rom{1}) and (\rom{2}), and leverage on TPVM condition (Condition~\ref{cond:TPVMm}) to obtain:
\begin{equation}
 \resizebox{.91\linewidth}{!}{$
            \displaystyle
\E\left[\sum_{t\in[T]}\text{(\rom{1})}\right] \overset{(a)}{\le} B_v\E\left[\sum_{t=1}^T\sqrt{\sum_{i \in \tilde{S}_t}(p_i^{\tilde{\bmu}_t, S_t})^{\lambda}{(\bar{\mu}_{t,i}-\tilde{\mu}_{t,i})^2}/{\bar{V}_{t,i}}}\right]$}.\label{eq:TPVM_mismatch_1}
\end{equation}

However, we cannot use TPE as \Cref{eq:regret_derive_VM} because $p_i^{\tilde{\bmu}_t,S_t}\neq p_i^{\bmu_t,S_t}$ in general. To handle this mismatch, we use the fact that triggering probability usually satisfies a smoothness condition in Condition~\ref{cond:TPM_P}, and prove that the mismatch only affect the lower-order term as follows: 

By Condition~\ref{cond:TPM_P}, $(p_i^{\tilde{\bmu}_t,S_t})^\lambda$ is upper bounded by $(p_i^{\bmu_t,S_t}+\min\{1,\sum_{j \in \tilde{S}_t}B_p p_j^{\bmu, S_t}\abs{{\mu}_{t,j}-\tilde{\mu}_{t,j}}\})^2$ when $\lambda \ge 2$, and the regret is bounded by the terms as shown below:
\noindent
\resizebox{.97\columnwidth}{!}{
\begin{minipage}{\columnwidth}
\begin{talign*}
&\text{Eq. }(\ref{eq:TPVM_mismatch_1}) \le \underbrace{\E\left[\sum_{t=1}^T B_v \sqrt{\sum_{i \in \tilde{S}_t}3p_i^{\bmu_t,S_t}{(\bar{\mu}_{t,i}-\tilde{\mu}_{t,i})^2}/{\bar{V}_{t,i}}} \right]}_{\text{leading term}}\notag\\
&+\underbrace{\frac{B_vB_p\sqrt{K}}{\sqrt{p_{\min}}}\E\left[ \sum_{t=1}^T \sum_{i \in \tilde{S}_t}{p_i^{\bmu_t,S_t}(\bar{\mu}_{t,i}-\ubar{\mu}_{t,i})^2}/{\bar{V}_{t,i}}\right]}_{\text{lower-order term}},
\end{talign*}
\end{minipage}}
where the leading term is of order $O(B_v \sqrt{T}\log T)$ by using the same derivation of step (c)-(e) in \Cref{eq:regret_derive_VM}, and the lower order term is bounded by $O(B_vB_p\sqrt{K/p_{\min}}\log T)$ by TPE and the weighted ellipsoidal potential lemma (\Cref{apdx_lem:ccmab_potential_new}). For $\lambda \ge 1$, the lower-order term becomes $\frac{B_v\sqrt{B_p}K^{1/4}}{\sqrt{p_{\min}}}\E\left[ \sum_{t=1}^T \left(\sum_{i \in \tilde{S}_t}\frac{p_i^{\bmu_t,S_t}(\bar{\mu}_{t,i}-\ubar{\mu}_{t,i})^2}{\bar{V}_{t,i}}\right)^{3/4}\right]$, which results in a larger lower-order regret term. See \Cref{apdx_sec:proof_TPVM_lambda} for details.

\section{Applications and Experiments}\label{sec:app}
% \jinhang{Comments for this section: 1) be careful about the parameters for different applications; 2) there's no application for Alg 2, maybe put PMC and add some examples there}
% \siwei{Maybe we can move this part to just below Theorem 1. In this section we mainly focus on how the applications follow our condition 3.} \carlee{I agree with this suggestion.} 
We now move to applications and experimental results.  We first show how our theoretical results improve various \ccmab~and \ccmab-T applications under 1-norm TPM, TPVM and VM smoothness conditions with their corresponding $B_1, B_v, \lambda$ coefficients. Then, we provide an empirical comparison in the context of the contextual cascading bandit application.

The instantiation of our theoretical results in the context of a variety of specific \ccmab~ and \ccmab-T applications is shown in \Cref{tab:app}.  The final column of the table details the improvement in regret that our results yield in each case. For detailed settings, proofs, and the discussion of the application results, see \Cref{apdx_sec:apps}.

\begin{figure}[t]
    \centering
    \begin{subfigure}[t]{0.48\linewidth}
    \includegraphics[width=\linewidth]{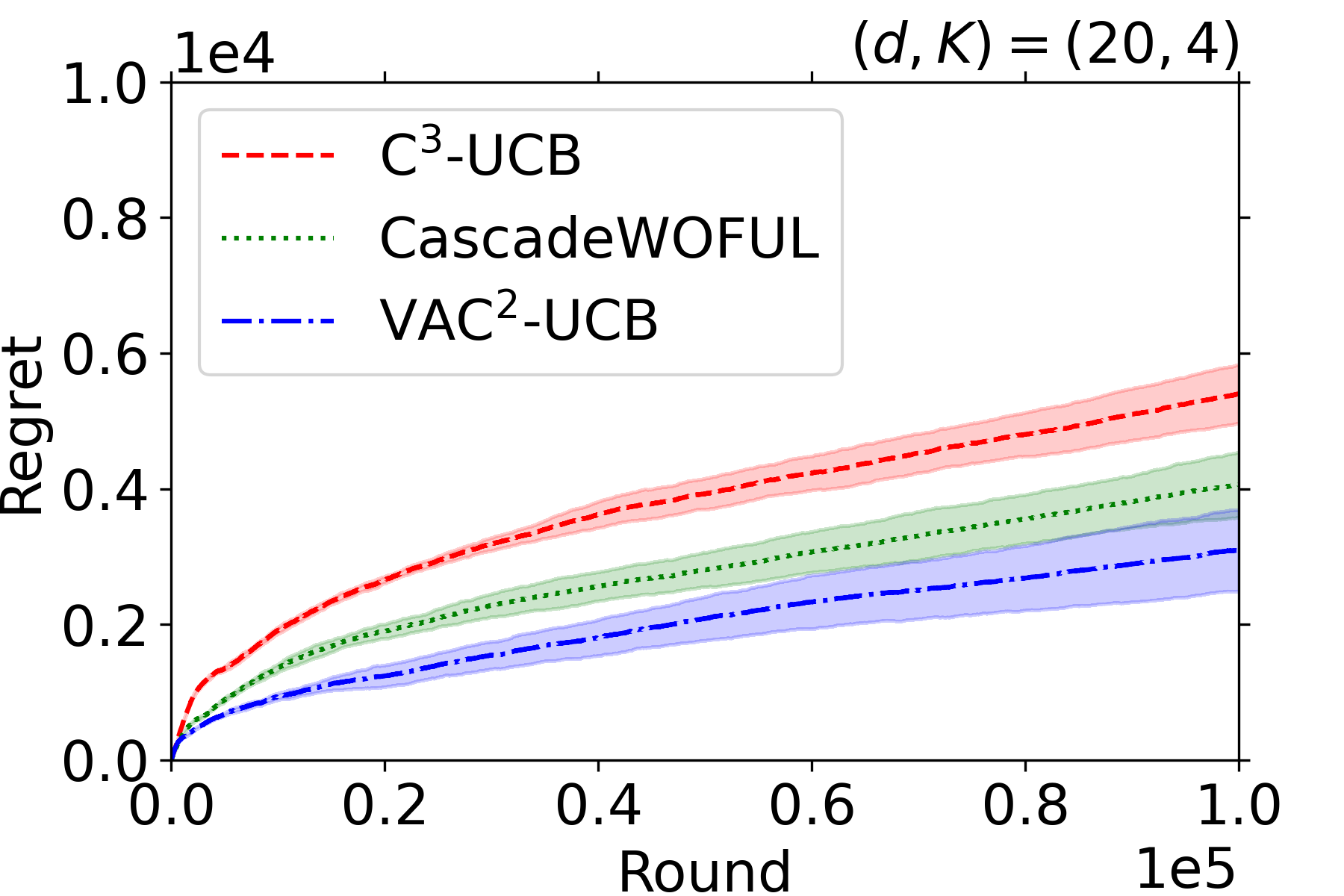}
    \caption{All genres}
    \label{fig:real_all}
    \end{subfigure}
    \begin{subfigure}[t]{0.48\linewidth}
    \includegraphics[width=\linewidth]{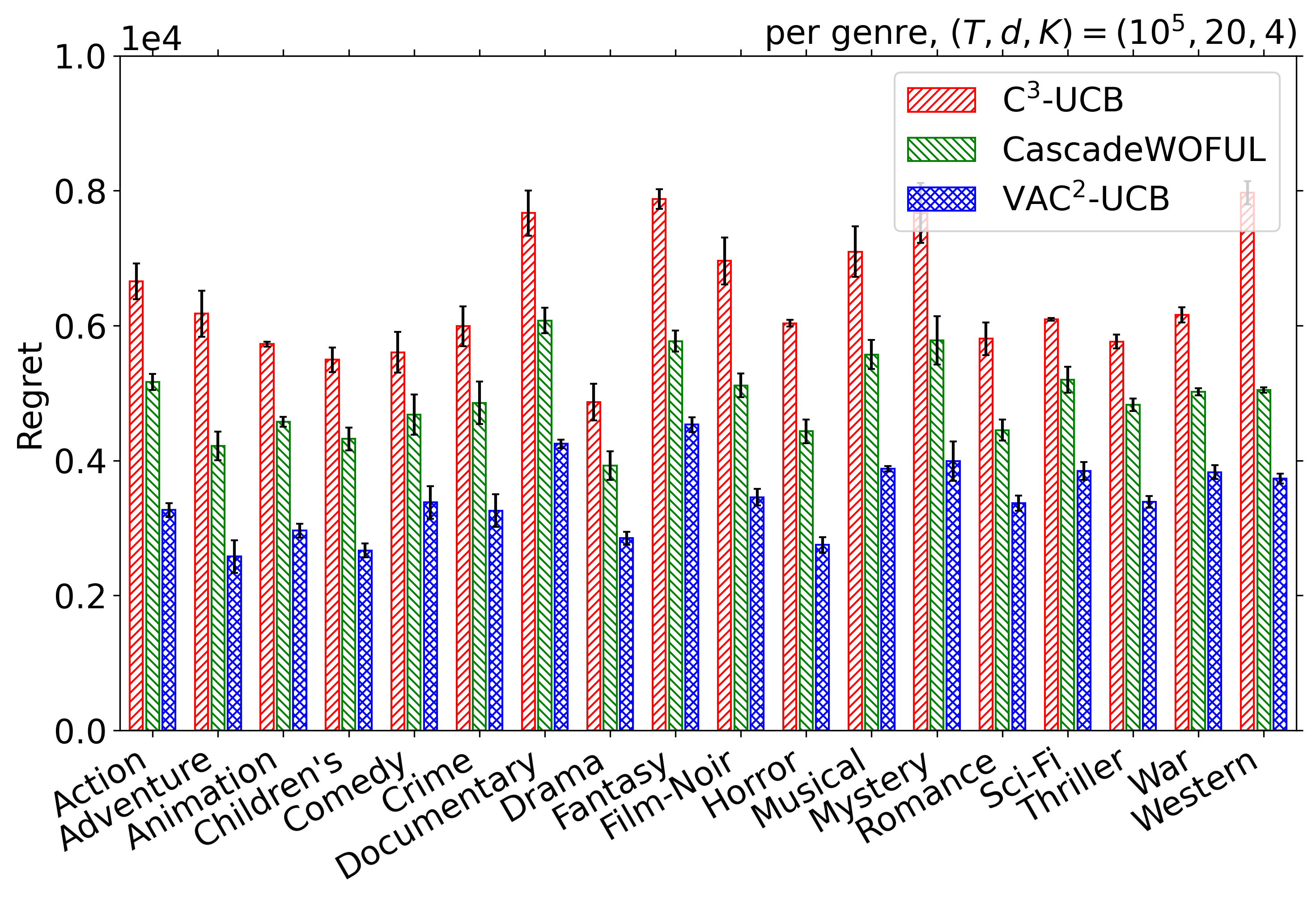}
    \caption{Particular genre}
    \label{fig:real_genre}
    \end{subfigure}
    \caption{Regret results for MovieLens data.}
    \label{fig:reg}
\end{figure}

Our experimental results are summarized in \Cref{fig:reg}, which details experiments on the MovieLens-1M dataset\footnote{ \href{https://grouplens.org/datasets/movielens/1m/}{grouplens.org/datasets/movielens/1m/}}.  Experiments on other data are included in the Appendix. \Cref{fig:reg} illustrates that our VAC$^2$-UCB algorithm outperforms C$^3$-UCB~\cite{li2016contextual}, the variance-agnostic cascading bandit algorithm, and CascadeWOFUL~\cite{vial2022minimax}, the state-of-the-art variance-aware cascading bandit algorithm, eventually incurring $45\%$ and $25\%$ less regret.
For detailed settings, comparisons, and discussions, see \Cref{sec:apps-exp}.

% \begin{figure}[ht]
%     \centering
%     \includegraphics[width=0.43\linewidth]{figures/ICML/real_all.png}
%     \caption{Results for MovieLens data}
%     \label{fig:syn_lin}
% \end{figure}
% \begin{figure}[ht]
%     \centering
%     \includegraphics[width=0.43\linewidth]{figures/ICML/real_genre.png}
%     \caption{Results for MovieLens data}
%     \label{fig:syn_lin}
% \end{figure}

% \vspace{-7pt}
\section{Conclusion}\label{sec:conclusion}
% \vspace{-7pt}
This paper studies contextual combinatorial bandits with probabilistically triggered arms (\ccmab-T) under a variety of smoothness conditions. 
Under the triggering probability modulated (TPM) condition, we design the C$^2$-UCB-T algorithm 
and propose a novel analysis to achieve an $\tilde{O}(d\sqrt{KT})$ regret bound, removing a potentially exponentially large factor $O(1/p_{\min})$.
Under the variance modulated conditions (VM or TPVM), we propose a new variance-adaptive algorithm  VAC$^2$-UCB and derive a regret bound $\tilde{O}(d\sqrt{T})$, which removes the batch-size $K$ dependence. As valuable by-product, we find our TPE analysis technique and variance-adaptive algorithm can be applied to the CMAB-T and \ccmab~setting, improving existing results as well. Experiments show that our algorithm can achieve at least 13\% and 25\% improvement compared with benchmark algorithms on synthetic and real-world datasets, respectively.
For the future study, it would be interesting to extend our application scenarios. One could also relax the perfectly linear assumption by introducing model mis-specifications or corruptions.

\section*{Acknowledgement}
The work of John C.S. Lui was supported in part by RGC’s GRF 14215722. The work of Mohammad Hajiesmaili is supported by NSF CAREER-2045641, CPS-2136199, CNS-2106299, and CNS-2102963. Wierman is supported by NSF grants CNS-2146814, CPS-2136197, CNS-2106403, and NGSDI-2105648.

\clearpage
\bibliography{main}
\bibliographystyle{icml2023}

\appendix

\clearpage
\onecolumn

\section*{Appendix}
% We would like to correct two typos in the first nine pages of our main paper. First, in line 267, we should have $\log(\frac{B_vK}{\Delta_{i}^{\min}})$ bins instead of $\log((B_vK)/(\Delta_{i}^{\min}))$ bins. Second, in line 6 of \cref{alg:SECUCB}, we miss one $B_v$ leading term in the writing and the correct confidence interval should be $\rho_{t}(S)=B_v\sqrt{\sum_{i \in S} \frac{C_1}{T_{t-1},i}+ \max\left\{8C_1\sqrt{\sum_{i \in S}\frac{\log(2|\cS|T)}{T_{t-1,i}^2}}, \frac{8C_1\log(2|\cS|T)}{T^{\min}_{t-1, S}}\right\}}$. The rest of the contents in the first nine pages are exactly the same as before.

The Appendix is organized as follows. 
\Cref{apdx_sec:TPM} gives the detailed proofs for theorems and lemmas in \Cref{sec:c2mab}.
\Cref{apdx_sec:TPVM} provides the detailed proofs for theorems and lemmas in \Cref{sec:va_c2mab}.
\Cref{apdx_sec:NC_C2MAB} shows how the triggering probability equivalence technique can be applied to non-contextual CMAB-T to obtain improved results.
\Cref{apdx_sec:apps} gives the detailed settings, results and comparisons included in \Cref{tab:app}.
\Cref{sec:apps-exp} provides detailed experimental setups and additional results.
\Cref{apdx_sec:Techs} summarizes the concentration bounds, facts, and technical lemmas used in this paper.
% We first compare the {\TPVMm} Condition (Condition~\ref{cond:TPVMm}) with TPM Condition (Condition~\ref{cond:TPM}) in~\cref{app:TPVMvsTPM}. 
% The comparisons between Gini-smoothness Condition with the TPVM (Condition~\ref{cond:TPVMm}) and VM (Condition~\ref{cond:VM}) Condition are introduced in \cref{app:gini}.
% The detailed proofs for \cref{thm:reg_lambda1} together with some discussions are in~\cref{apdx_sec:main_regret_analysis}.
% The detailed proofs for \cref{thm:independent} are in \cref{apdx_sec:independent}.
% The application details and the proof details related to \cref{lem:app_tab} are in~\cref{apdx_sec:proof_app}.
% The experiments for different applications are included in~\cref{apdx_sec:exp}.

\section{Proofs for \ccmab-T under the TPM Condition (Section~\ref{sec:c2mab})}
 \label{apdx_sec:TPM}
 \subsection{Proof of Theorem~\ref{thm:c2mab_reg}}\label{apdx_sec:c2mab_sub_1}
 We first give/recall some definitions and events. Recall that in \Cref{alg: C2UCB}, the gram matrix, the b-vector and the estimator are 
 \begin{align}
\bG_t&=\gamma \bI + \sum_{s< t}\sum_{i \in \tau_{s}}\bphi_{s}(i) \bphi_{s}(i)^{\top}\\
\bb_t&=\sum_{s< t}\sum_{i \in \tau_{s}}\bphi_{s}(i)X_{s,i}\\
\hat{\btheta}_t&=\bG_t^{-1}\bb_t.
 \end{align}
 Let us use $\cW_t$ to denote the nice event when the oracle can output solution $S$ with $r(S; \bmu)\ge \alpha\cdot r(S^*;\bmu)$ where $S^*=\argmax_{S \in \cS}r(S;\bmu)$ for any $\bmu$ at round $t$. We use $\cN_t$ to denote the nice event when the $\norm{\hat{\btheta}_t-\btheta^*}_{\bG_{t}} \le \rho(\delta)$ holds for any $t\in [T]$. Define the filtration to be $\cF_{t-1}=(S_1, \bphi_1, \tau_1, (X_{1,i})_{t\in \tau_1}, ..., S_{t-1}, \bphi_{t-1}, \tau_{t-1}, (X_{t-1,i})_{t\in \tau_{t-1}}, S_t, \bphi_t)$ that takes both history data $\cH_t$ and action $S_t$ to handle the randomness of the oracle, and let $\E_t[\cdot]=\E[\cdot\mid \cF_{t-1}]$. Now we bound the regret under nice event $\cW_t$ and $\cN_t$, 
\begin{align}\label{apdx_eq:c2mab_reg}
&\text{Reg}(T)\overset{(a)}{=}\E\left[\sum_{t \in [T]}\E_t[\alpha\cdot r(S^*_t;\bmu_t)-r(S_t;\bmu_t)]\right]\notag\\
&\overset{(b)}{\le}\E\left[\sum_{t \in [T]}\E_t[\alpha\cdot r(S^*_t;\bar{\bmu}_t)-r(S_t;\bmu_t)]\right]\overset{(c)}{\le} \E\left[\sum_{t \in [T]}\E_t[r(S_t;\bar{\bmu}_t)-r(S_t;\bmu_t)]\right]\notag\\
    &\overset{(d)}{\le}\E\left[\sum_{t \in [T]}\E_t\left[\sum_{i \in \tilde{S}_t}B_1p_i^{\bmu_t,S_t}(\bar{\mu}_{t,i}-\mu_{t,i})\right]\right]\notag\\
    &\overset{(e)}{=} \E\left[\sum_{t \in [T]}\E_t\left[\sum_{i \in \tau_t}B_1(\bar{\mu}_{t,i}-\mu_{t,i})\right]\right]\notag\\
&\overset{(f)}{\le} \E\left[\sum_{t \in [T]}\E_t\left[\sum_{i \in \tau_t}2B_1 \rho(\delta)\norm{\bphi_t(i)}_{\bG_{t}^{-1}}\right]\right]\overset{(g)}{=}  2B_1\rho(\delta)\E\left[\sum_{t \in [T]}\sum_{i \in \tau_t}\norm{\bphi_t(i)}_{\bG_{t}^{-1}}\right]\notag\\
&\overset{(h)}{\le}  2B_1\rho(\delta) \E\left[\sqrt{KT\sum_{t \in [T]}\sum_{i \in \tau_t}\norm{\bphi_t(i)}^2_{\bG_{t}^{-1}}}\right] \notag\\
&\overset{(i)}{\le}O\left(B_1(\sqrt{2d\log T} +\sqrt{\gamma})\sqrt{2KT\log T}\right)
\le O(B_1 d\sqrt{KT}\log T).
\end{align}
 where (a) follows from the regret definition and the tower rule, (b) is by Condition~\ref{cond:mono} and \Cref{lem:c2mab_optimism} saying that $\mu_{t,i}\le\bar{\mu}_{t,i}$, (c) is by nice event $\cW_t$ and the definition of $S_t$, (d) is by Condition~\ref{cond:TPM}, (e) follows from by the TPE trick~\Cref{apdx_lem:TPE}, (f) is by~\Cref{lem:c2mab_optimism}, (g) is by tower rule, (h) by Cauchy Schwarz inequality, and (i) is by the ellipsoidal potential lemma (\Cref{apdx_lem:ccmab_potential}). Similar to~\cite{wang2017improving} The theorem is concluded by the definition of the $(\alpha,\beta)$-approximate regret, and considering event $\neg \cW_t$ or $\neg \cN_t$, which contributes to at most $(1-\beta)T\Delta_{\max}+\delta T \Delta_{\max}$ regret.
 \subsection{Important Lemmas used for proving Theorem \ref{thm:c2mab_reg}}\label{apdx_sec:lemmas_thm1}
\lemccmabOptimism*
\begin{proof}
    For any $i\in [m], t\in [T]$, we have
    \begin{align*}
        &\abs{\inner{\hat{\btheta}_t, \bphi_t(i)}-\inner{\btheta^*, \bphi_t(i)}}\\
        &{=}\abs{\inner{\hat{\btheta}_t-\btheta^*, \bphi_t(i)} }\\
        &\overset{(a)}{\le} \norm{\hat{\btheta}_t-\btheta^*}_{\bG_t} \cdot \norm{\bphi_t(i)}_{\bG_t^{-1}}\\
    &\overset{(b)}{\le}\rho(\delta)\norm{\bphi_t(i)}_{\bG_t^{-1}},
    \end{align*}
    where $(a)$ by  Cauchy-Schwartz, (b) by \Cref{lem:c2mab_good_est}.
    Now use the definition of $\mu_{t,i}=\inner{\btheta^*, \bphi_t(i)}$ and $\bar{\mu}_{t,i}=\inner{\hat{\btheta}_t, \bphi_t(i)}+\rho(\delta)\norm{\bphi_t(i)}_{\bG_t^{-1}}$ finishes the proof.
\end{proof}

\begin{lemma}[Triggering Probability Equivalence (TPE)]\label{apdx_lem:TPE}
    $\E_t\left[\sum_{i \in \tilde{S}_t}B_1p_i^{\bmu_t,S_t}(\bar{\mu}_{t,i}-\mu_{t,i})\right]= \E_t\left[\sum_{i \in \tau_t}B_1(\bar{\mu}_{t,i}-\mu_{t,i})\right]$.
\end{lemma}
\begin{proof}
    We have 
    \begin{align}\label{apdx_eq:TPE}
% &\E_t[\alpha\cdot r(S^*_t;\bmu_t)-r(S_t;\bmu_t)]\overset{(a)}{\le}\E_t[\alpha\cdot r(S^*_t;\bar{\bmu}_t)-r(S_t;\bmu_t)]\notag\\
%    &\overset{(b)}{\le} \E_t[r(S_t;\bar{\bmu}_t)-r(S_t;\bmu_t)]\notag\\
    &\E_t\left[\sum_{i \in \tilde{S}_t}B_1p_i^{\bmu_t,S_t}(\bar{\mu}_{t,i}-\mu_{t,i})\right]\notag\\
    &\overset{(a)}{=}\E\left[\sum_{i \in \tilde{S}_t}B_1\E_{\tau_t}[\I\{i \in \tau_t\}](\bar{\mu}_{t,i}-\mu_{t,i})\mid \cF_{t-1}\right]\notag\\
    &\overset{(b)}{=}\E_t\left[\sum_{i \in \tilde{S}_t}\I\{i \in \tau_t\}B_1(\bar{\mu}_{t,i}-\mu_{t,i})\right]\notag\\
    &\overset{(c)}{=}\E_t\left[\sum_{i \in \tau_t}B_1(\bar{\mu}_{t,i}-\mu_{t,i})\right],
\end{align}
% where (a) is by Condition~\ref{cond:mono} and \Cref{lem:c2mab_optimism} saying that $\mu_{t,i}\le\bar{\mu}_{t,i}$, (b) is by nice event $\cW$, (c) is by Condition~\ref{cond:TPM}, 
(a) is because $\bar{\mu}_{t,i}, \mu_{t,i}, S_t$ are $\cF_{t-1}$-measurable so that the only randomness is from triggering set $\tau_t$ and we can substitute $p_i^{\bmu_t, S_t}$ with event $\I\{i \in \tau_t\}$ under expectation, (b) is by absorbing the expectation over $\tau_t$ to $\E_t$, and (c) is a simple change of notation. Actually, TPE can be applied whenever the quantities (other than $p_i^{D,S}$) are $\cF_{t-1}$-measurable, which would be helpful for later sections.
\end{proof}

% \begin{lemma}[Ellipsoidal Potential Lemma]\label{apdx_lem:ccmab_potential_new}
% Let $\gamma>0$, $(\bx_{t,i})_{i \in \tau_s, t \in [T]}$ be a sequence that $\bx_{t,i}\in \R^d$, $\norm{\bx_{t,i}}_{2}\le L$ and $\abs{\tau_t}\le K$. Let $\bG_t=\gamma\cdot \bI + \sum_{s=1}^{t-1}\sum_{i \in \tau_s}\bx_{t,i}\bx_{t,i}^{\top}$, then we have
% \begin{align}
% \sum_{t \in [T]}\sum_{i \in \tau_s} \min{1,\norm{\bx_i}_{\bG_t}^2}\le 2dK \log (1+ L^2KT/(d\gamma)
% \end{align}
% \end{lemma}

 \begin{lemma}[Ellipsoidal Potential Lemma]\label{apdx_lem:ccmab_potential} $\sum_{t=1}^T\sum_{i \in \tau_t}\norm{\bphi_t(i)}_{\bG_t^{-1}}^2\le 2d\log (1+KT/(\gamma d))\le 2d\log T$ when $\gamma \ge K$.
 \end{lemma}
\begin{proof}
    \begin{align}
        \det(\bG_{t+1})&\overset{(a)}{=}\det\left(\bG_{t} + \sum_{i\in \tau_{t}}\bphi_t(i)\bphi_t(i)^{\top}\right)\notag\\
        &\overset{(b)}{=}\det(\bG_{t})\cdot \det\left(\bI + \sum_{i\in \tau_{t}}G_t^{-1/2}\bphi_{t}(i) (G_t^{-1/2}\bphi_{t}(i))^{\top}\right)\notag\\
        &\overset{(c)}{\ge} \det(\bG_{t})\cdot \left(1 + \sum_{i\in \tau_{t}}\norm{\bphi_{t}(i)}_{\bG_{t}^{-1}}^2\right)\notag\\
        &\overset{(d)}{\ge} \det(\gamma \bI) \prod_{s=1}^t \left(1+\sum_{i \in \tau_s}\norm{\bphi_s(i)}_{\bG_s^{-1}}^2\right)\label{apdx_eq:prod_and_det},
    \end{align}
    where (a) follows from the definition, (b) follows from $\det(\bA\bB)=\det(\bA)\det(\bB)$ and $\bA+\bB=\bA^{1/2} (\bI + \bA^{-1/2} \bB \bA^{-1/2})\bA^{1/2}$, (c) follows from \Cref{apdx_lem:det_and_norm}, (d) follows from repeatedly applying (c).

    Since $\norm{\bphi_s(i)}_{\bG_s^{-1}}^2\le \frac{\norm{\bphi_s(i)}^2}{\lambda_{\min}(\bG_s)}\le 1/\gamma \le 1/K$, we have $\sum_{i \in \tau_s}\norm{\bphi_s(i)}_{\bG_s^{-1}}^2\le 1$. Using the fact that $2\log (1+x)\ge x$ for any $[0,1]$, we have
    \begin{align*}
        &\sum_{s \in t}\sum_{i \in \tau_s}\norm{\bphi_s(i)}_{\bG_s^{-1}}^2\\
        &{\le} 2\sum_{s=1}^t\log \left(1+\sum_{i \in \tau_s}\norm{\bphi_s(i)}_{\bG_s^{-1}}^2\right)\\
        &= 2\log \prod_{s=1}^t\left(1+\sum_{i \in \tau_s}\norm{\bphi_s(i)}_{\bG_s^{-1}}^2\right)\\
        &\overset{(a)}{\le} 2\log\left(\frac{\det(\bG_{t+1})}{\det(\gamma \bI)}\right)\\
        &\overset{(b)}{\le} 2\log \left(\frac{(\gamma + KT/d)^d}{\gamma^d}\right)=2d\log (1+KT/(\gamma d))\le 2d\log(T),
    \end{align*}
    where the (a) follows from \Cref{apdx_eq:prod_and_det}, (b) follows from \Cref{apdx_lem:det_and_trace}.
\end{proof}

\section{Proofs for \ccmab-T under the VM or TPVM Condition (Section~\ref{sec:va_c2mab})}
 \label{apdx_sec:TPVM}
\subsection{Proof of Lemma~\ref{lem:va_c2mab_good_est}}\label{apdx_sec:TPVM_key_lem}
 Our analysis is inspired by the derivation of Theorem 3 by~\cite{lattimore2015linear} to bound the key ellipsoidal radius $\norm{\btheta^*-\hat{\btheta}_t}_{\bG_t}\le \rho$ for the \ccmab-T setting, where multiple arms can be triggered in each round. Before we going into the main proof, we first introduce some notations and events as follows.

% Let us denote the filtration $\cF_{t}=(\bphi_1, \tau_1, (X_{1,i})_{i\in \tau_1}, ..., \bphi_{t}, \tau_{t}, (X_{t,i})_{i\in \tau_{t}}, \bphi_{t+1})$.

Recall that for $t\ge 1$, $X_{t,i}$ is a Bernoulli random variable with mean $\mu_{t,i}=\inner{\btheta^*, \bphi_t(i)}$, suppose $\norm{\btheta^*}_2\le 1, \norm{\bphi_t(i)}\le 1$, we can represent $X_{t,i}$ by $X_{t,i}=\mu_{t,i} + \eta_{t,i}$, where noise $\eta_{t,i}\in [-1,1]$, its mean $\E[\eta_{t,i}\mid \cF_{t-1}]=0$, and its variance $\Var[\eta_{t,i}\mid \cF_{t-1}]=\mu_{t,i}(1-\mu_{t,i})$.
Also note that in \Cref{alg:va_cucb}, the gram matrices, the b-vector and the weighted least-square estimator are the following.
\begin{align}\label{apdx_eq:def_G_b_theta_VM}
\bG_t &= \gamma \cdot \bI + \sum_{s=1}^{t-1}\sum_{i \in \tau_s} \bar{V}_{s,i}^{-1}\bphi_s(i) \bphi_s(i)^{\top},\\
\bb_t &= \sum_{s=1}^{t-1}\sum_{i \in \tau_s}\bar{V}_{s,i}^{-1} \bphi_s(i) X_{s,i},\\
\hat{\btheta}_t&=\bG_{t}^{-1} \bb_{t},
\end{align}
where we set $G_0= \gamma \bI$, and optimistic variances $\bar{V}_{s,i}$ are defined as in \Cref{eq:var_upper} of \Cref{alg:va_cucb}.

Let us define $\bZ_{t}=\sum_{s < t}\sum_{i \in \tau_s} \eta_{s,i}\bphi_t(i)/\bar{V}_{s,i}$, and the key of this proof is to bound $\bZ_t$ (this quantity is often denoted as $S_t$ in the self-normalized bound~\cite{abbasi2011improved}, but $S_t$ is occupied to denote actions at round $t$ in this work). 

We finally define failure events $F_0\subseteq F_1 \subseteq ... \subseteq F_{T}$ be a sequence of events defined by
\begin{align}\label{apdx_eq:event_F}
F_t=\{\exists s\le t \text{ such that } \norm{\bZ_s}_{\bG_{s}}+\sqrt{\gamma}\ge \rho\}.
\end{align}
These failure events are crucial in the sense that $\btheta^*$ lies in the  confidence ellipsoid $\norm{\btheta^*-\hat{\btheta}_t}_{\bG_{t}} \le \rho$ (see \Cref{apdx_lem:event2radius} for its proof).

Next, we can prove by induction that the probability of $\norm{\bZ_{t}}_{\bG_{t}}+\sqrt{\gamma}\ge \rho$ given $\neg F_{t-1}$ is very small, for $t=1, ..., T$ (see its proof in  \Cref{apdx_lem:reduction_key}). Based on this, we can have $\Pr[\neg F_{T}]=1- \Pr[F_0] - \sum_{t=1}^T \Pr\left[\norm{\bZ_{t}}_{\bG_{t}}+\sqrt{\gamma}\ge \rho \text{ and } \neg F_{t-1}\right]\ge 1-\delta$ (as $\neg F_0$ always holds), and thus by \Cref{apdx_eq:event2radius}, \Cref{lem:va_c2mab_good_est} is proved as desired.

\subsection{Proof of Theorem~\ref{thm:ccmab_reg_VM} under VM condition}
Similar to \Cref{apdx_sec:TPM},  we first give/recall some definitions and events. Recall that in \Cref{alg:va_cucb},
the gram matrices, the b-vector, and the weighted least-square estimator are defined in~\Cref{apdx_eq:def_G_b_theta_VM}
% \begin{align}\label{apdx_eq:def_G_b_theta}
% \bG_t &= \gamma \cdot \bI + \sum_{s=1}^{t-1}\sum_{i \in \tau_s} \bar{V}_{s,i}^{-1}\bphi_s(i) \bphi_s(i)^{\top},\\
% \bb_t &= \sum_{s=1}^{t-1}\sum_{i \in \tau_s}\bar{V}_{s,i}^{-1} \bphi_s(i) X_{s,i},\\
% \hat{\btheta}_t&=\bG_{t}^{-1} \bb_{t},
% \end{align}
% where we set $G_0= \gamma \bI$, and 
The optimistic variances $\bar{V}_{s,i}$ are defined as in \Cref{eq:var_upper} of \Cref{alg:va_cucb}.
 Let us use $\cW_t$ to denote the nice event when the oracle can output solution $S$ with $r(S; \bmu)\ge \alpha\cdot r(S^*;\bmu)$ where $S^*=\argmax_{S \in \cS}r(S;\bmu)$ for any $\bmu$ at round $t$. We use $\cN_t$ to denote the nice event when the $\norm{\hat{\btheta}_t-\btheta^*}_{\bG_{t}} \le \rho(\delta)$ holds for any $t\in [T]$ (which can be implied by $\neg F_T$). Define the filtration to be $\cF_{t-1}=(S_1, \bphi_1, \tau_1, (X_{1,i})_{t\in \tau_1}, ..., S_{t-1}, \bphi_{t-1}, \tau_{t-1}, (X_{t-1,i})_{t\in \tau_{t-1}}, S_t, \bphi_t)$ that takes both history data $\cH_t$ and action $S_t$ to handle the randomness of the oracle, and let $\E_t[\cdot]=\E[\cdot\mid \cF_{t-1}]$. 
 
 Let $\tilde{\bmu}_{t}$ be the vector whose $i$-th entry is the maximizer that achieves $\bar{V}_{t,i}$, i.e., $\tilde{\mu}_{t,i}=\argmax_{\mu\in [\ubar{\mu}_{t,i}, \bar{\mu}_{t,i}]}\mu(1-\mu)$. Now we bound the regret under nice event $\cW_t$ and $\cN$ (where $\cN_t$ can be implied from $\neg F_T$ by 
 derivation in~\Cref{apdx_lem:event2radius}), 
 \begin{align}
&\text{Reg}(T)\overset{(a)}{=} \E\left[\sum_{t=1}^T  \alpha r(S_t^*; \bmu_t)- r(S_t; \bmu_t)\right]\\
&\overset{(b)}{\le} \E\left[\sum_{t=1}^T \alpha r(S^*_t; \bar{\bmu}_t)- r(S_t; \bmu_t)\right]\\
&\overset{(c)}{\le} \E\left[\sum_{t=1}^T r(S_t; \bar{\bmu}_t)- r(S_t; \bmu_t)\right]\\
&\overset{(d)}{\le} \E\left[\sum_{t=1}^T \underbrace{\abs{r(S_t; \bar{\bmu}_t)- r(S_t; \tilde{\bmu}_t)}}_{\text{(\rom{1})}}+\underbrace{\abs{r(S_t; \bmu_t)- r(S_t; \tilde{\bmu}_t)}}_{\text{(\rom{2})}}\right],
% &\le B_v\E\left[\sum_{t=1}^T\left(\sqrt{\sum_{i \in [m]}\frac{(\bar{\mu}_{t,i}-\tilde{\mu}_{t,i})^2}{\bar{V}_{t,i}}}+\sqrt{\frac{(\mu_{t,i}-\tilde{\mu}_{t,i})^2}{\bar{V}_{t,i}}}\right)\right]\\
% &\le  B_v\E\left[\sqrt{T\sum_{t=1}^T\frac{(\bar{\mu}_{t,i}-\tilde{\mu}_{t,i})^2}{\bar{V}_{t,i}}}+\sqrt{T\sum_{t=1}^T\frac{(\mu_{t,i}-\tilde{\mu}_{t,i})^2}{\bar{V}_{t,i}}}\right]\\
% &\le B_v\sqrt{T}\sqrt{\E\left[\sum_{t=1}^T\frac{(\bar{\mu}_{t,i}-\tilde{\mu}_{t,i})^2}{\bar{V}_{t,i}}\right]}+\sqrt{\E\left[\sum_{t=1}^T\frac{(\mu_{t,i}-\tilde{\mu}_{t,i})^2}{\bar{V}_{t,i}}\right]}\\
\end{align}
where (a) is by definition, (b) follows from Condition~\ref{cond:mono} and \Cref{lem:va_c2mab_optimism}, (c) from event $\cW$ and the definition of $S_t$, (d) from triangle inequality.

Now we show how to bound term \text{(\rom{1})},
\begin{align}\label{apdx_eq:va_ccmab_term_1}
&\E\left[\sum_{t\in[T]}\text{(\rom{1})}\right] \overset{(a)}{\le} B_v\E\left[\sum_{t=1}^T\sqrt{\sum_{i \in \tilde{S}_t}\frac{(\bar{\mu}_{t,i}-\tilde{\mu}_{t,i})^2}{\bar{V}_{t,i}}}\right]\notag\\
& \overset{(b)}{\le}  \frac{B_v}{\sqrt{p_{\min}}}\cdot \E\left[\sum_{t=1}^T\sqrt{\sum_{i \in \tilde{S}_t}p_{i}^{\bmu_t,S_t}\frac{(\bar{\mu}_{t,i}-\tilde{\mu}_{t,i})^2}{\bar{V}_{t,i}}}\right]\notag\\
% & \le  \frac{B_v}{\sqrt{p_{\min}}}\cdot \E\left[\sqrt{T\sum_{t=1}^T\sum_{i \in \tilde{S}_t}p_{i}^{\bmu_t,S_t}\frac{(\bar{\mu}_{t,i}-\tilde{\mu}_{t,i})^2}{\bar{V}_{t,i}}}\right]\notag\\
& \overset{(c)}{\le}  \frac{B_v}{\sqrt{p_{\min}}}\cdot \E\left[\sqrt{T\sum_{t=1}^T\sum_{i \in \tilde{S}_t}p_i^{\bmu_t,S_t}\frac{(\bar{\mu}_{t,i}-\tilde{\mu}_{t,i})^2}{\bar{V}_{t,i}}}\right]\notag\\
& \overset{(d)}{\le}  \frac{B_v}{\sqrt{p_{\min}}}\cdot \sqrt{T\E\left[\sum_{t=1}^T\sum_{i \in \tilde{S}_t}p_{i}^{\bmu_t,S_t}\frac{(\bar{\mu}_{t,i}-\tilde{\mu}_{t,i})^2}{\bar{V}_{t,i}}\right]}\notag\\
&  \overset{(e)}{=} \frac{B_v}{\sqrt{p_{\min}}}\cdot \sqrt{T\E\left[\sum_{t=1}^T\sum_{i \in \tau_t}\frac{(\bar{\mu}_{t,i}-\tilde{\mu}_{t,i})^2}{\bar{V}_{t,i}}\right]}\notag\\
& \overset{(f)}{\le} \frac{B_v}{\sqrt{p_{\min}}}\cdot \sqrt{T\E\left[\sum_{t=1}^T\sum_{i \in \tau_t}\frac{(6\rho(\delta)\norm{\bphi_t(i)}_{\bG_{t}^{-1}})^2}{\bar{V}_{t,i}}\right]}\notag\\
& \overset{(g)}{\le} {O}(B_vd\sqrt{T}\log (KT)/\sqrt{p_{\min}}),
\end{align}
where (a) follows from Condition~\ref{cond:VM}, (b) follows from the definition of $p_{\min}$ s.t. $p_i^{\bmu_t,S_t}\ge p_{\min}$ for $i \in \tilde{S}_t$, (c) follows from Cauchy–Schwarz, (d) follows from Jensen's inequality, (e) follows from the TPE trick, (f) follows from \Cref{lem:va_c2mab_optimism}, (g) follows from \Cref{apdx_lem:ccmab_potential_new}.

Now for the term (\rom{2})$\le{O}(B_vd\sqrt{T}\log (KT)/\sqrt{p_{\min}}) $ follows from the similar derivation of \Cref{apdx_eq:va_ccmab_term_1} by replacing $(\bar{\mu}_{t,i}-\tilde{\mu}_{t,i})^2$ with $(\mu_{t,i}-\tilde{\mu}_{t,i})^2$. And the theorem is concluded by considering $\neg \cW_t$ and $\neg \cN_t$, similar to~\Cref{apdx_sec:TPM}.

\begin{lemma}[Weighted Ellipsoidal Potential Lemma]\label{apdx_lem:ccmab_potential_new} $\sum_{t=1}^T\sum_{i \in \tau_t}\norm{\bphi_t(i)}_{\bG_t^{-1}}^2/\bar{V}_{t,i}\le 2d\log (1+KT/(\gamma d))\le 2d\log T$ when $\neg F_T$ and $\gamma \ge 4K$.
 \end{lemma}
\begin{proof}
    \begin{align}
        \det(\bG_{t+1})&\overset{(a)}{=}\det\left(\bG_{t} + \sum_{i\in \tau_{t}}\bphi_t(i)\bphi_t(i)^{\top}/\bar{V}_{t,i}\right)\notag\\
        &\overset{(b)}{=}\det(\bG_{t})\cdot \det\left(\bI + \sum_{i\in \tau_{t}}G_t^{-1/2}\bphi_{t}(i) (G_t^{-1/2}\bphi_{t}(i))^{\top}/\bar{V}_{t,i}\right)\notag\\
        &\overset{(c)}{\ge} \det(\bG_{t})\cdot \left(1 + \sum_{i\in \tau_{t}}\norm{\bphi_{t}(i)}_{\bG_{t}^{-1}}^2/\bar{V}_{t,i}\right)\notag\\
        &\overset{(d)}{\ge} \det(\gamma \bI) \prod_{s=1}^t \left(1+\sum_{i \in \tau_s}\norm{\bphi_s(i)}_{\bG_s^{-1}}^2/\bar{V}_{t,i}\right)\label{apdx_eq:prod_and_det_2},
    \end{align}
    where (a) follows from the definition, (b) follows from $\det(\bA\bB)=\det(\bA)\det(\bB)$ and $\bA+\bB=\bA^{1/2} (\bI + \bA^{-1/2} \bB \bA^{-1/2})\bA^{1/2}$, (c) follows from \Cref{apdx_lem:det_and_norm}, (d) follows from repeatedly applying (c).

     If $\bar{V}_{s,i}=\frac{1}{4}$,  $\norm{\bphi_s(i)}_{\bG_s^{-1}}^2/\bar{V}_{s,i}\le \frac{4\norm{\bphi_s(i)}^2}{\lambda_{\min}(\bG_s)}\le 4/\gamma \le 1/K$, else if $\bar{V}_{s,i}<\frac{1}{4}$, and since $\neg \cF_T$, by \Cref{apdx_lem:beta_and_v_norm_potential}, $\norm{\bphi_s(i)}_{\bG_s^{-1}}^2/\bar{V}_{s,i}\le \frac{1}{\rho(\delta)}\frac{1}{\sqrt{\gamma}}\le \frac{1}{\gamma}\le 1/(4K)$. Therefore, we have $\sum_{i \in \tau_s}\norm{\bphi_s(i)}_{\bG_s^{-1}}^2\le 1$. Using the fact that $2\log (1+x)\ge x$ for any $[0,1]$, we have
    \begin{align*}
        &\sum_{s \in t}\sum_{i \in \tau_s}\norm{\bphi_s(i)}_{\bG_s^{-1}}^2/\bar{V}_{s,i}\\
        &{\le} 2\sum_{s=1}^t\log \left(1+\sum_{i \in \tau_s}\norm{\bphi_s(i)}_{\bG_s^{-1}}^2/\bar{V}_{s,i}\right)\\
        &= 2\log \prod_{s=1}^t\left(1+\sum_{i \in \tau_s}\norm{\bphi_s(i)}_{\bG_s^{-1}}^2/\bar{V}_{s,i}\right)\\
        &\overset{(a)}{\le} 2\log\left(\frac{\det(\bG_{t+1})}{\det(\gamma \bI)}\right)\\
        &\overset{(b)}{\le} 2\log \left(\frac{(\gamma + KT/d)^d}{\gamma^d}\right)=2d\log (1+4dK^2T^2/(\gamma d))\le 4d\log(KT),
    \end{align*}
    where the (a) follows from \Cref{apdx_eq:prod_and_det}, (b) follows from \Cref{apdx_lem:det_and_trace} by setting $L=\norm{\bphi_s(i)}^2/\bar{V}_{s,i}\le 4dKs$ (from \Cref{apdx_lem:l2_norm}).
\end{proof}

\subsection{Proof of Theorem~\ref{thm:ccmab_reg_TPVM_lambda} Under TPVM Condition}\label{apdx_sec:proof_TPVM_lambda}
In this section, we consider two cases when $\lambda \ge 2$ and $\lambda\ge 1$.
Recall that to use the TPVM condition (Condition~\ref{cond:TPVMm}), we need one additional condition over the triggering probability (Condition~\ref{cond:TPM_P}).

% \begin{condition}[\textbf{1-norm TPM Bounded Smoothness for Triggering Probability}]\label{cond:TPM_P}
% % We say that a CMAB-T instance satisfies TPM $B_1$-bounded smoothness condition, if for any action $S \in \cS$, any distribution $D,D'$ with mean vector $\bmu \in (0,1)^m$, we have $\frac{\partial r(S;\bmu)}{\partial \mu_i } \frac{1}{p_{i}^{D,S}}\le B_1 $ for $i \in \tilde{S}$ and $\frac{\partial r(S;\bmu)}{\partial \mu_i }=0$ for $i \in [m] \backslash \tilde{S}$. 
% We say that a \ccmab-T problem instance satisfies the triggering probability modulated (TPM) $B_p$-bounded smoothness condition over the triggering probability, if for any action $S \in \cS$, any mean vectors $\bmu, \bmu' \in [0,1]^m$, we have $|p_i^{\bmu', S}-p_i^{\bmu, S}|\le B_p\sum_{i \in [m]}p_{i}^{\bmu,S}|\mu_i-\mu'_i|$.
% %\siwei{seems no formal definition of this $p_{i}^{D,S}$?} 
% \end{condition}

\subsubsection{When $\lambda \ge 2$:}

We inherit the same notation and events as in \Cref{apdx_sec:c2mab_sub_1}, and start to bound term (\rom{1}) in \Cref{apdx_eq:va_ccmab_term_1} differently,
\begin{align}\label{apdx_eq:va_ccmab_term_1_new}
&\E\left[\sum_{t\in[T]}\text{(\rom{1})}\right] \overset{(a)}{\le} \E\left[\sum_{t=1}^T B_v \sqrt{\sum_{i \in \tilde{S}_t}(p_i^{\tilde{\bmu}_t,S_t})^2\frac{(\bar{\mu}_{t,i}-\tilde{\mu}_{t,i})^2}{\bar{V}_{t,i}}}\right]\\
 & \overset{(b)}{\le}\E\left[\sum_{t=1}^T B_v \sqrt{\sum_{i \in \tilde{S}_t}\left(p_i^{\bmu_t,S_t}+\min\left\{1,\sum_{j \in \tilde{S}_t}B_p p_j^{\bmu, S_t}\abs{{\mu}_{t,j}-\tilde{\mu}_{t,j}}\right\}\right)^2\cdot\frac{(\bar{\mu}_{t,i}-\tilde{\mu}_{t,i})^2}{\bar{V}_{t,i}}} \right]\\
    & \overset{(c)}{\le} \E\left[\sum_{t=1}^T B_v \sqrt{\sum_{i \in \tilde{S}_t}3p_i^{\bmu_t,S_t}\cdot\frac{(\bar{\mu}_{t,i}-\tilde{\mu}_{t,i})^2}{\bar{V}_{t,i}}} \right] \notag\\
    &+\E\left[\sum_{t=1}^T B_v \sqrt{\sum_{i \in \tilde{S}_t}\left(\sum_{j \in \tilde{S}_t}B_p p_j^{\bmu_t, S_t}\abs{{\mu}_{t,j}-\tilde{\mu}_{t,j}}\right)^2\cdot\frac{(\bar{\mu}_{t,i}-\tilde{\mu}_{t,i})^2}{\bar{V}_{t,i}}} \right]\\
    &\overset{(d)}{=} O(B_vd\sqrt{T}\log (KT)) +   B_v\E\left[\sum_{t=1}^T \sqrt{\sum_{i \in \tilde{S}_t}\frac{(\bar{\mu}_{t,i}-\tilde{\mu}_{t,i})^2}{\bar{V}_{t,i}}} \cdot \sum_{j \in \tilde{S}_t}B_p p_j^{\bmu_t, S_t}\abs{{\mu}_{t,j}-\tilde{\mu}_{t,j}}\right]\\
     % &\le B_vd\sqrt{T}\log T +   B_v\frac{1}{\sqrt{p_{\min}}}\E\left[\sum_{t=1}^T \sqrt{\sum_{i \in \tilde{S}_t}\frac{p_i^{\bmu_t,S_t}(\bar{\mu}_{t,i}-\tilde{\mu}_{t,i})^2}{\bar{V}_{t,i}}} \cdot \sqrt{K\sum_{j \in \tilde{S}_t}B^2_p p_j^{\bmu_t, S_t}\abs{{\mu}_{t,j}-\tilde{\mu}_{t,j}}^2}\right]\\
     &\overset{(e)}{\le} O(B_vd\sqrt{T}\log (KT)) +   B_v\frac{1}{\sqrt{p_{\min}}}\E\left[\sum_{t=1}^T \sqrt{\sum_{i \in \tilde{S}_t}\frac{p_i^{\bmu_t,S_t}(\bar{\mu}_{t,i}-\tilde{\mu}_{t,i})^2}{\bar{V}_{t,i}}} \cdot  \sum_{j \in \tilde{S}_t}B_p p_j^{\bmu_t, S_t}\abs{{\mu}_{t,j}-\tilde{\mu}_{t,j}}\right]\\
    &\overset{(f)}{\le} O(B_vd\sqrt{T}\log (KT)) +   B_v\frac{1}{\sqrt{p_{\min}}}\E\left[\sum_{t=1}^T \sqrt{\sum_{i \in \tilde{S}_t}\frac{p_i^{\bmu_t,S_t}(\bar{\mu}_{t,i}-\tilde{\mu}_{t,i})^2}{\bar{V}_{t,i}}} \cdot \sqrt{K\sum_{j \in \tilde{S}_t}B_p^2 p_j^{\bmu_t, S_t}\abs{{\mu}_{t,j}-\tilde{\mu}_{t,j}}^2}\right]\\
    &\overset{(g)}{\le} O(B_vd\sqrt{T}\log (KT)) +   B_vB_p\frac{\sqrt{K}}{\sqrt{p_{\min}}}\E\left[\sum_{t=1}^T \sqrt{\sum_{i \in \tilde{S}_t}\frac{p_i^{\bmu_t,S_t}(\bar{\mu}_{t,i}-\tilde{\mu}_{t,i})^2}{\bar{V}_{t,i}}} \cdot \sqrt{\sum_{j \in \tilde{S}_t} 
 p_j^{\bmu_t, S_t}\abs{{\mu}_{t,j}-\tilde{\mu}_{t,j}}^2/\bar{V}_{t,j}}\right]\\
    &\overset{(h)}{\le}  O(B_vd\sqrt{T}\log (KT)) +   B_vB_p\frac{\sqrt{K}}{\sqrt{p_{\min}}}\E\left[ \sum_{t=1}^T \sum_{i \in \tilde{S}_t}\frac{p_i^{\bmu_t,S_t}(\bar{\mu}_{t,i}-\ubar{\mu}_{t,i})^2}{\bar{V}_{t,i}} \right]\\
    &\overset{(i)}{\le}  O(B_vd\sqrt{T}\log (KT) +B_vB_p\frac{\sqrt{K}}{\sqrt{p_{\min}}} (d\log (KT))^2)=O(B_vd\sqrt{T}\log (KT)),
\end{align}
where (a) follows from Condition~\ref{cond:TPVMm}, (b) is by applying Condition~\ref{cond:TPM_P} for triggering probability $p_i^{\bar{\bmu}_{t}, \tilde{S}_t}$ and $p_i^{\bar{\bmu}_t, \tilde{S}_t}, p_i^{{\bmu}_t, \tilde{S}_t}\le 1$, (c) follows from $\sqrt{a+b}\le \sqrt{a} + \sqrt{b}$, (d) follows from same derivation from \Cref{apdx_eq:va_ccmab_term_1}, (e) follows from $p_i^{\tilde{\bmu}_t,S_t}\ge p^i_{\min}$, (f) follows from Cauchy-Schwarz, (g) follows from $\bar{V}_{t,j}\le 1/4$, (h) follows from $\tilde{\mu}_{t,i}, \mu_{t,i} \in [\bar{\mu}_{t,i}, \ubar{\mu}_{t,i}]$ by event $\cN_t$, (i) follows from the similar analysis of (d)-(g) in \Cref{apdx_eq:va_ccmab_term_1} inside the square-root without considering the additional $B_v\sqrt{T}/\sqrt{p_{\min}}$. 

For the term (\rom{2}), one can easily verify it follows from the similar deviation of the term (\rom{1}) with the difference in constant terms. And \Cref{thm:ccmab_reg_TPVM_lambda} is concluded by considering small probability $\neg \cW_t$ and $\cN_t$ events.

\subsubsection{{When $\lambda \ge 1$:}}

We inherit the same notation and events as in \Cref{apdx_sec:c2mab_sub_1}, and start to bound term (\rom{1}) in \Cref{apdx_eq:va_ccmab_term_1_new} as follows,
\begin{align}\label{apdx_eq:va_ccmab_term_1_new_2}
&\E\left[\sum_{t\in[T]}\text{(\rom{1})}\right] \overset{(a)}{\le} \E\left[\sum_{t=1}^T B_v \sqrt{\sum_{i \in \tilde{S}_t}(p_i^{\tilde{\bmu}_t,S_t})\frac{(\bar{\mu}_{t,i}-\tilde{\mu}_{t,i})^2}{\bar{V}_{t,i}}}\right]\\
 & \overset{(b)}{\le}\E\left[\sum_{t=1}^T B_v \sqrt{\sum_{i \in \tilde{S}_t}\left(p_i^{\bmu_t,S_t}+\min\left\{1,\sum_{j \in \tilde{S}_t}B_p p_j^{\bmu, S_t}\abs{{\mu}_{t,j}-\tilde{\mu}_{t,j}}\right\}\right)\cdot\frac{(\bar{\mu}_{t,i}-\tilde{\mu}_{t,i})^2}{\bar{V}_{t,i}}} \right]\\
    & \overset{(c)}{\le} \E\left[\sum_{t=1}^T B_v \sqrt{\sum_{i \in \tilde{S}_t}p_i^{\bmu_t,S_t}\cdot\frac{(\bar{\mu}_{t,i}-\tilde{\mu}_{t,i})^2}{\bar{V}_{t,i}}} \right] \notag\\
    &+\E\left[\sum_{t=1}^T B_v \sqrt{\sum_{i \in \tilde{S}_t}\left(\sum_{j \in \tilde{S}_t}B_p p_j^{\bmu_t, S_t}\abs{{\mu}_{t,j}-\tilde{\mu}_{t,j}}\right)\cdot\frac{(\bar{\mu}_{t,i}-\tilde{\mu}_{t,i})^2}{\bar{V}_{t,i}}} \right]\\
    &\overset{(d)}{=} O(B_vd\sqrt{T}\log (KT)) +   B_v\E\left[\sum_{t=1}^T \sqrt{\sum_{i \in \tilde{S}_t}\frac{(\bar{\mu}_{t,i}-\tilde{\mu}_{t,i})^2}{\bar{V}_{t,i}}} \cdot \sqrt{\sum_{j \in \tilde{S}_t}B_p p_j^{\bmu_t, S_t}\abs{{\mu}_{t,j}-\tilde{\mu}_{t,j}}}\right]\\
     % &\le B_vd\sqrt{T}\log T +   B_v\frac{1}{\sqrt{p_{\min}}}\E\left[\sum_{t=1}^T \sqrt{\sum_{i \in \tilde{S}_t}\frac{p_i^{\bmu_t,S_t}(\bar{\mu}_{t,i}-\tilde{\mu}_{t,i})^2}{\bar{V}_{t,i}}} \cdot \sqrt{K\sum_{j \in \tilde{S}_t}B^2_p p_j^{\bmu_t, S_t}\abs{{\mu}_{t,j}-\tilde{\mu}_{t,j}}^2}\right]\\
     &\overset{(e)}{\le} O(B_vd\sqrt{T}\log (KT)) +   B_v\frac{1}{\sqrt{p_{\min}}}\E\left[\sum_{t=1}^T \sqrt{\sum_{i \in \tilde{S}_t}\frac{p_i^{\bmu_t,S_t}(\bar{\mu}_{t,i}-\tilde{\mu}_{t,i})^2}{\bar{V}_{t,i}}} \cdot  \sqrt{\sum_{j \in \tilde{S}_t}B_p p_j^{\bmu_t, S_t}\abs{{\mu}_{t,j}-\tilde{\mu}_{t,j}}}\right]\\
    &\overset{(f)}{\le} O(B_vd\sqrt{T}\log (KT)) +   B_v\frac{1}{\sqrt{p_{\min}}}\E\left[\sum_{t=1}^T \sqrt{\sum_{i \in \tilde{S}_t}\frac{p_i^{\bmu_t,S_t}(\bar{\mu}_{t,i}-\tilde{\mu}_{t,i})^2}{\bar{V}_{t,i}}} \cdot \left({K\sum_{j \in \tilde{S}_t}B_p^2 p_j^{\bmu_t, S_t}\abs{{\mu}_{t,j}-\tilde{\mu}_{t,j}}^2}\right)^{1/4}\right]\\
    &\overset{(g)}{\le} O(B_vd\sqrt{T}\log (KT)) +   B_v\sqrt{B_p}\frac{K^{1/4}}{\sqrt{p_{\min}}}\E\left[\sum_{t=1}^T \sqrt{\sum_{i \in \tilde{S}_t}\frac{p_i^{\bmu_t,S_t}(\bar{\mu}_{t,i}-\tilde{\mu}_{t,i})^2}{\bar{V}_{t,i}}} \cdot \left(\sum_{j \in \tilde{S}_t} 
 p_j^{\bmu_t, S_t}\abs{{\mu}_{t,j}-\tilde{\mu}_{t,j}}^2/\bar{V}_{t,j}\right)^{1/4}\right]\\
    &\overset{(h)}{\le} O(B_vd\sqrt{T}\log (KT)) +   B_v\sqrt{B_p}\frac{{K}^{1/4}}{\sqrt{p_{\min}}}\E\left[ \sum_{t=1}^T \left(\sum_{i \in \tilde{S}_t}\frac{p_i^{\bmu_t,S_t}(\bar{\mu}_{t,i}-\ubar{\mu}_{t,i})^2}{\bar{V}_{t,i}}\right)^{3/4} \right]\\
    &\overset{(i)}{\le} O(B_vd\sqrt{T}\log (KT)) +   B_v\sqrt{B_p}\frac{{(KT)}^{1/4}}{\sqrt{p_{\min}}}\E\left[  \left(\sum_{t=1}^T\sum_{i \in \tilde{S}_t}\frac{p_i^{\bmu_t,S_t}(\bar{\mu}_{t,i}-\ubar{\mu}_{t,i})^2}{\bar{V}_{t,i}}\right)^{3/4} \right]\\
    &\overset{(j)}{\le}  O\left(B_vd\sqrt{T}\log (KT) +B_v\sqrt{B_p}\frac{(KT)^{1/4}}{\sqrt{p_{\min}}} \left(d\log (KT)\right)^{3/2}\right)=O(B_vd\sqrt{T}\log (KT)),
\end{align}
where (a) follows from Condition~\ref{cond:TPVMm}, (b) is by applying Condition~\ref{cond:TPM_P} for triggering probability $p_i^{\bar{\bmu}_{t}, \tilde{S}_t}$ and $p_i^{\bar{\bmu}_t, \tilde{S}_t}, p_i^{{\bmu}_t, \tilde{S}_t}\le 1$, (c) follows from $\sqrt{a+b}\le \sqrt{a} + \sqrt{b}$, (d) follows from same derivation from \Cref{apdx_eq:va_ccmab_term_1}, (e) follows from $p_i^{\tilde{\bmu}_t,S_t}\ge p^i_{\min}$, (f) follows from Cauchy-Schwarz, (g) follows from $\bar{V}_{t,j}\le 1/4$, (h) follows from $\tilde{\mu}_{t,i}, \mu_{t,i} \in [\bar{\mu}_{t,i}, \ubar{\mu}_{t,i}]$ by event $\cN_t$, (i) follows from Holder's inequality with $p=4, q=4/3$, (j) follows from the similar analysis of (d)-(g) in \Cref{apdx_eq:va_ccmab_term_1} inside the square-root without considering the additional $B_v\sqrt{T}/\sqrt{p_{\min}}$. 

For the term (\rom{2}), one can easily verify it follows from the similar deviation of the term (\rom{1}) with the difference in constant terms. And \Cref{thm:ccmab_reg_TPVM_lambda} is concluded by considering small probability $\neg \cW_t$ and $\cN_t$ events.

 \section{Proofs for TPE Trick to Improve Non-Contextual CMAB-T}\label{apdx_sec:NC_C2MAB}

We first introduce some definitions that are used in~\cite{wang2017improving} and \cite{liu2022batch}. Recall that non-contextual CMAB-T is a degenerate case when $\bphi_t(i)=\boldsymbol{e}_i$ and $\btheta^*=\bmu$, where $\bmu\triangleq\E_{\bX_{t}\sim \cD}[\bX_{t}\mid \cH_t]$ is the mean of the true outcome distribution $D$.

\begin{definition}[(Approximation) Gap]\label{def:gap}
Fix a distribution $D \in \cD$ and its mean vector $\bmu$, for each action $S \in \cS$, we define the (approximation) gap as $\Delta_S=\max\{0, \alpha r(S^*;\bmu)-r(S;\bmu)\}$. 
% the arm set that could be triggered by $S$ as $\tilde{S}=\{i \in [m]: p_i^{D,S}>0\}$. 
For each arm $i$, we define $\Delta_i^{\min}=\inf_{S \in \cS: p_i^{D,S} > 0,\text{ } \Delta_S > 0}\Delta_S$, $\Delta_i^{\max}=\sup_{S \in \cS: p_i^{D,S} > 0, \Delta_S > 0}\Delta_S$.
As a convention, if there is no action $S \in \cS$ such that $p_i^{D,S}>0$ and $\Delta_S>0$, 
% \wei{Should this be $\Delta_S > 0$?}, 
	then $\Delta_i^{\min}=+\infty, \Delta_i^{\max}=0$. We define $\Delta_{\min}=\min_{i \in [m]} \Delta_i^{\min}$ and $\Delta_{\max}=\max_{i \in [m]} \Delta_i^{\max}$. 
\end{definition}

\begin{definition}[Event-Filtered Regret]\label{apdx_def:event_filter_reg} For any series of events $(\cE_t)_{t \in [T]}$ indexed by round number $t$, we define the $Reg^{A}_{\alpha,\bmu}(T,(\cE_t)_{t \in [T]})$ as the regret filtered by events $(\cE_t)_{t \in [T]}$, or the regret is only counted in $t$ if $\cE$ happens in $t$. Formally, 
\begin{align}
    Reg^{A}_{\alpha,\bmu}(T,(\cE_t)_{t \in [T]}) = \E\left[\sum_{t\in[T]}\I(\cE_t)(\alpha\cdot r(S^*;\bmu)-r(S_t;\bmu))\right].
\end{align}
For simplicity, we will omit $A,\alpha,\bmu,t\in[T]$ and rewrite $Reg^{A}_{\alpha,\bmu}(T,(\cE_t)_{t \in [T]})$ as $Reg(T, \cE_t)$ when contexts are clear.

\end{definition}

 \subsection{Reproducing  Theorem 1 of \cite{wang2017improving} under 1-norm 
TPM Condition}\label{apdx_sec:reproduce_TPM}
 \begin{theorem}\label{thm:reg_reproduce}
For a CMAB-T problem instance $([m], \cS, \cD, D_{\text{trig}},R)$ that satisfies monotonicity (Condition \ref{cond:mono}), 
% 1-norm TPM bounded smoothness (Condition \ref{cond:TPM}) with coefficient $B_1$ \siwei{Why we need condition \ref{cond:TPM} here?},
and TPM bounded smoothness (Condition \ref{cond:TPM}) with coefficient $B_1$, 
% let constants $c_1=12\sqrt{6}, c_2=34$,
if $\lambda \ge 1$, { CUCB \cite{wang2017improving} with an $(\alpha,\beta)$-approximation oracle achieves an $(\alpha,\beta)$-approximate distribution-dependent regret} bounded by
\begin{equation}\label{eq:reg_reproduce}
Reg(T)\le   \sum_{i \in [m]}\frac{48B_1^2 K \log T}{\Delta_i^{\min}} + 2B_1m  +\frac{\pi^2}{3}\cdot \Delta_{\max}.
\end{equation}
And the distribution-independent regret, 
\begin{align}
Reg(T)\le  14B_1\sqrt{m K T\log T} + 2B_1m  + \frac{\pi^2}{3}\cdot \Delta_{\max}.
\end{align}
\end{theorem}

The main idea is to use TPE trick to replace $\tilde{S}_t$ (arms that could be probabilistically triggered by action $S_t$) with $\tau_{t}$ (arms that are actually triggered by action $S_t$) under conditional expectation, so that we can use the simpler Appendix B.2 of \citet{wang2017improving} to avoid the much more involved Appendix B.3 of \citet{wang2017improving}. Such replacement bypasses the triggering group analysis (and its counter $N_{t,i,j}$) in Appendix B.3, which uses $N_{t,i,j}$ to associate $T_{t,i}$ with the counters for $\tilde{S}_t$. For our simplified analysis, we can directly associate the $T_{t,i}$ with the arm triggering for the arms $\tau_t$ that are actually triggered/observed and eventually reproduce the regret bounds of \cite{wang2017improving}. 

We follow exactly the same CUCB algorithm (Algorithm 1 \cite{wang2017improving}), conditions (Condition 1, 2 \cite{wang2017improving}). We also inherit the event definitions of $\cN_t^s$ (Definition 4~\cite{wang2017improving}) that for every arm $i\in[m]$, $\abs{\hat{\mu}_{t-1,i}-\mu_i} < \rho_{t,i}=\sqrt{\frac{3\log t}{2T_{t-1,i}}}$, and the event $F_t$ being $\{r(S_t;{\bar{\bmu}}_t)<\alpha\cdot \text{opt}(\bar{\bmu}_t)\}$.
Let us further denote $\Delta_{S_t}=\alpha r(S^*;\bmu)-r(S_t;\bmu)$, $\tau_t$ be the arms actually triggered by $S_t$ at time $t$. Let filtration $\cF_{t-1}$ be $(\phi_1, S_1, \tau_1, (X_{1,i})_{i \in \tau_1}, ..., \phi_{t-1}, S_{t-1}, \tau_{t-1}, (X_{t-1,i})_{i \in \tau_{t-1}}, \phi_t, S_t)$, and let $\E_t[\cdot]=\E[\cdot\mid \cF_{t-1}]$ We also have that $\cF_{t-1}$, $T_{t-1,i}, \hat{\mu}_{t,i}$ are measurable. Also note that we use $p_i^{D,S}$ to denote the triggering probability $p_i^{\bmu,S}$ for any $i\in[m], S\in \cS$ in order to match the notation of \citet{wang2017improving}.

\begin{proof}
Under event $\cN^s_t$ and $\neg F_t$, and given filtration $\cF_{t-1}$, we have 
\begin{align}
    \Delta_{S_t}&\overset{(a)}{\le} B_1 \sum_{i \in [m]}p_i^{D, S_t}(\bar{\mu}_{t,i}-\mu_i)\\
    &\overset{(b)}{\le} -\Delta_{S_t} + 2B_1 \sum_{i \in [m]}p_i^{D, S_t}(\bar{\mu}_{t,i}-\mu_i)\\
    &=-\frac{\sum_{i \in [m]}p_i^{D, S_t}\Delta_{S_t}}{\sum_{i \in [m]}p_i^{D, S_t}}+2B_1 \sum_{i \in [m]}p_i^{D, S_t}(\bar{\mu}_{t,i}-\mu_i)\\
    &\overset{(c)}{\le} 2B_1 \sum_{i \in [m]}p_i^{D,S_t}\left(-\frac{\Delta_i^{\min}}{2B_1K}+ (\bar{\mu}_{t,i}-\mu_i)\right)\\
    &\overset{(d)}{\le}  2B_1 \sum_{i \in [m]}p_i^{D,S_t}\left(-\frac{\Delta_i^{\min}}{2B_1K}+ \min\left\{1, \sqrt{\frac{6\log T}{T_{t-1,i}}}\right\}\right),\label{apdx_eq:TPM_simple_1}
    % &\le \sum_{i \in [m]}p_i^{D,S_t}\kappa_T(M_i, T_{t-1,i}).
\end{align}
where (a) follows from exactly the Equation (10) of Appendix B.3 in \citet{wang2017improving}, (b) is by the reverse amortization trick that multiplies two to both sides of (a) and rearranges the terms, (c) is by $p_i^{D,S_t}\le 1$ and $\Delta_{i}^{\min}\le \Delta_{S_t}$, (d) by event $\cN^s_t$ so that $(\bar{\mu}_{t,i}-\mu_i)\le \min\{1, 2\rho_{t,i}\}=\left\{1, \sqrt{\frac{6\log T}{T_{t-1,i}}}\right\}$.

Let \begin{equation}
\kappa_{i,T}(\ell) = \begin{cases}
2B_1, &\text{if $\ell=0$,}\\
2B_1\sqrt{\frac{6\log T}{\ell}}, &\text{if $1\le\ell\le L_{i,T}$,}\\
0, &\text{if $\ell > L_{i,T}$,}
\end{cases}
\end{equation}
where $L_{i,T}=\frac{24B_1^2K^2 \log T}{(\Delta_i^{\min})^2}$.

It follows that 
\begin{align}
    \Delta_{S_t}&=\E_t[\Delta_{S_t}]\overset{(a)}{\le} \E_t\left[2B_1 \sum_{i \in [m]}p_i^{D,S_t}\left(-\frac{\Delta_i^{\min}}{2B_1K}+ \min\left\{1, \sqrt{\frac{6\log T}{T_{t-1,i}}}\right\}\right)\right]\\
    &\overset{(b)}{=} \E_t\left[2B_1\sum_{i \in [m]}\I\{i \in \tau_t\}\left(-\frac{\Delta_i^{\min}}{2B_1K}+ \min\left\{1, \sqrt{\frac{6\log T}{T_{t-1,i}}}\right\}\right)\right]\\
    &= \E_t\left[2B_1\sum_{i \in \tau_t}\left(-\frac{\Delta_i^{\min}}{2B_1K}+ \min\left\{1, \sqrt{\frac{6\log T}{T_{t-1,i}}}\right\}\right)\right]\\
    &\overset{(c)}{\le} \E_t \left[\sum_{i \in \tau_t}\kappa_{i,T}(T_{t-1,i})\right],\label{apdx_eq:TPM_simple_2}
\end{align}
where (a) follows from \Cref{apdx_eq:TPM_simple_1}, (b) follows from the TPE trick to replace $p_i^{D,S_t}=\E_t[\I\{i \in \tau_t\}]$, (c) follows from that if $T_{t-1,i}\le L_{i,T}$, we have $\min\{\sqrt{\frac{6\log T}{T_{t-1,i}}}, 1\}\le \frac{1}{2B_1}\kappa_{i,T}(T_{t-1, i})$, and if $T_{t-1, i}\ge L_{i,T} +1$, then $\min\left\{1, \sqrt{\frac{6\log T}{T_{t-1,i}}}\right\}\le \frac{\Delta_{i}^{\min}}{2B_1K}$, so $-\frac{\Delta_i^{\min}}{2B_1K}+\min\left\{1, \sqrt{\frac{6\log T}{T_{t-1,i}}}\right\}\le 0=\kappa_{i,T}(T_{t-1,i})$.

Now we apply the definition of the event-filtered regret,
\begin{align}
    Reg(\cN_t^s, \neg F_t)&= \E\left[\sum_{t=1}^T \Delta_{S_t}\right]\\
    &\overset{(a)}{\le} \E\left[\sum_{t=1}^T\E_t\left[\sum_{i \in \tau_t}\kappa_{i,T}(T_{t-1,i})\right]\right]\\
    &\overset{(b)}{=}\E\left[\sum_{t=1}^T\sum_{i \in \tau_t}\kappa_{i,T}(T_{t-1,i})\right]\\
     &\overset{(c)}{=}\E\left[\sum_{i \in [m]}\sum_{s=0}^{T_{T-1,i}}\kappa_{i,T}(s)\right]\\
     &\le \sum_{i \in [m]}\sum_{s=0}^{L_{i,T}} \kappa_{i,T}(s)\\
     &= 2B_1m + \sum_{i \in [m]}\sum_{s=1}^{L_{i,T}} 2B_1 \sqrt{\frac{6\log T}{s}}\\
     &\overset{(d)}{\le} 2B_1m + \sum_{i \in [m]}\int_{s=0}^{L_{i,T}} 2B_1 \sqrt{\frac{6\log T}{s}}\cdot ds\\
     &\le 2B_1m  + \sum_{i \in [m]}\frac{48B_1^2 K \log T}{\Delta_i^{\min}},
\end{align}
where (a) follows from \Cref{apdx_eq:TPM_simple_2}, (b) follows from the tower rule, (c) follows from that $T_{t-1,i}$ is increased by $1$ if and only if $i \in \tau_t$, (d) is by the sum $\&$ integral inequality  $\int_{L-1}^{U}f(x)dx\ge\sum_{i=L}^Uf(i)\ge \int_{L}^{U+1}f(x)dx$ for non-increasing function $f$.

Following \citet{wang2017improving} to handle small probability events $\neg \cN_t^s$ and $F_t$, we have 
\begin{align}
Reg(T)\le   \sum_{i \in [m]}\frac{48B_1^2 K \log T}{\Delta_i^{\min}} + 2B_1m  +\frac{\pi^2}{3}\cdot \Delta_{\max},
\end{align}
and the distribution-independent regret is
\begin{align}
Reg(T)\le  14B_1\sqrt{m K T\log T} + 2B_1m  + \frac{\pi^2}{3}\cdot \Delta_{\max}.
\end{align}
\end{proof}

\subsection{Improving Theorem 1 of \cite{liu2022batch} under 
TPVM Condition}
We first show the regret bound of using our TPE technique in \Cref{thm:reg_lambda1} and its prior result in \Cref{apdx_prop:liu_1}.
\begin{restatable}{theorem}{lambda1}\label{thm:reg_lambda1}
For a CMAB-T problem instance $([m], \cS, \cD, D_{\text{trig}},R)$ that satisfies monotonicity (Condition \ref{cond:mono}), 
% 1-norm TPM bounded smoothness (Condition \ref{cond:TPM}) with coefficient $B_1$ \siwei{Why we need condition \ref{cond:TPM} here?},
and TPVM bounded smoothness (Condition \ref{cond:TPVMm}) with coefficient $(B_v, B_1, \lambda)$, 
% let constants $c_1=12\sqrt{6}, c_2=34$,
if $\lambda \ge 1$, { BCUCB-T \cite{liu2022batch} with an $(\alpha,\beta)$-approximation oracle achieves an $(\alpha,\beta)$-approximate distribution-dependent regret} bounded by
\begin{equation}\label{eq:reg_lambda1}
    O\left(\sum_{i \in [m]}\frac{B_v^2 \log K \log T}{\tDelta_{i,\lambda}^{\min}} + \sum_{i \in [m]}B_1\log\left(\frac{B_1K}{\Delta_i^{\min}}\right)\log T\right),
\end{equation}
where  $ \tilde{\Delta}_{i,\lambda}^{\min}=\min_{S: p_i^{D,S}>0, \Delta_S>0}\Delta_{S_t}/(p_i^{D,S_t})^{\lambda-1}$.
And the distribution-independent regret, 
\begin{align}
Reg(T) \le O\left( B_v\sqrt{m(\log K) T\log T} + B_1m \log(KT)\log T\right).
\end{align}
\end{restatable}

\begin{proposition}[Theorem 1, \citet{liu2022batch}]\label{apdx_prop:liu_1}
For a CMAB-T problem instance $([m], \cS, \cD, D_{\text{trig}},R)$ that satisfies monotonicity (Condition \ref{cond:mono}), 
% 1-norm TPM bounded smoothness (Condition \ref{cond:TPM}) with coefficient $B_1$ \siwei{Why we need condition \ref{cond:TPM} here?},
and TPVM bounded smoothness (Condition \ref{cond:TPVMm}) with coefficient $(B_v, B_1, \lambda)$, 
% let constants $c_1=12\sqrt{6}, c_2=34$,
if $\lambda \ge 1$,  { BCUCB-T with an $(\alpha,\beta)$-approximation oracle achieves an $(\alpha,\beta)$-approximate regret} bounded by
\begin{equation}\label{eq:reg_lambda2}
    % O\left(\left[ \sum_{i=1}^m \left(\log\frac{8c_1^2B_v^2 K}{(\Delta_{i}^{\min})^2}\right)\frac{288c_1^2 B_v^2}{\Delta_{i}^{\min}} (3+\log K)  +\sum_{i=1}^m\left(144c_2B_1\log\frac{4c_2B_1 K}{\Delta_{i}^{\min}}\right) \left(1+\log \frac{\Delta_i^{\max} K}{\Delta_{i}^{\min}}\right) \right]\log T\right)
    O\left(\sum_{i \in [m]} \log\left(\frac{B_vK}{\Delta_i^{\min}}\right) \frac{B_v^2\log K \log T}{\Delta_i^{\min}} + \sum_{i \in [m]}B_1\log^2\left(\frac{B_1K}{\Delta_i^{\min}}\right)\log T\right).
\end{equation}
And the distribution-independent regret, 
\begin{align}
Reg(T) \le O\left( B_v\sqrt{m(\log K) T} \log (KT) + B_1m \log^2(KT)\log T\right).
\end{align}
\end{proposition}

Looking at our regret bound (\Cref{thm:reg_lambda1}), there are two improvements compared with \Cref{apdx_prop:liu_1}: (1) the min gap is improved to $\tDelta_{i,\lambda}^{\min}\ge \Delta_i^{\min}$, (2) we remove a $O(\log (\frac{B_v K}{\Delta_i^{\min}}))$ for the leading term.
For (2), it translates to a $O(\sqrt{\log T})$ improvement for the distribution-independent bound.

\begin{proof}
Similar to \Cref{apdx_sec:reproduce_TPM}, the main idea is to use TPE trick to replace $\tilde{S}_t$ (arms that could be probabilistically triggered by action $S_t$) with $\tau_{t}$ (arms that are actually triggered by action $S_t$) under conditional expectation to avoid the usage of much more involved triggering group analysis \cite{wang2017improving}. Such replacement bypasses the triggering group analysis (and its counter $N_{t,i,j}$) \cite{liu2022batch}, which uses $N_{t,i,j}$ to associate $T_{t,i}$ with the counters for $\tilde{S}_t$. By doing so, we do not need a union bound over the group index $j$, which saves a $\log (B_vK/\Delta_i^{\min}$ (or $\log (B_1K/\Delta_i^{\min}$) factor.

We follow exactly the same BCUCB-T algorithm (Algorithm 1 \cite{liu2022batch}), conditions (Condition 1, 2, 3 \cite{liu2022batch}). We also inherit the event definitions of $\cN_t^s$ (Definition 6~\cite{liu2022batch}) that (1) for every base arm $i \in [m]$, $|\hat{\mu}_{t-1, i}-\mu_i|\le \rho_{t,i}$, where $\rho_{t,i}=\sqrt{\frac{6\hat{V}_{t-1,i} \log t}{T_{t-1, i}}} + \frac{9\log t}{T_{t-1, i}}$; (2) for every base arm $i \in [m]$, $\hat{V}_{t-1,i} \le 2\mu_i(1-\mu_i) + \frac{3.5\log t}{T_{t-1,i}}$. We use the event $F_t$ being $\{r(S_t;\bar{\bmu}_t)<\alpha\cdot \text{opt}(\bar{\bmu}_t)\}$.
Let us further denote $\Delta_{S_t}=\alpha r(S^*;\bmu)-r(S_t;\bmu)$, $\tau_t$ be the arms actually triggered by $S_t$ at time $t$. Let filtration $\cF_{t-1}$ be $(\phi_1, S_1, \tau_1, (X_{1,i})_{i \in \tau_1}, ..., \phi_{t-1}, S_{t-1}, \tau_{t-1}, (X_{t-1,i})_{i \in \tau_{t-1}}, \phi_t, S_t)$, and let $\E_t[\cdot]=\E[\cdot\mid \cF_{t-1}]$ We also have that $\cF_{t-1}$, $T_{t-1,i}, \hat{\mu}_{t,i}$ are measurable. Also note that we use $p_i^{D,S}$ to denote the triggering probability $p_i^{\bmu,S}$ for any $i\in[m], S\in \cS$ in order to match the notation of \citet{wang2017improving, liu2022batch}.

We follow the same regret decomposition as in Lemma 9 of \citet{liu2022batch}, to decompose the event-filtered regret $Reg(T, \cN^s_t, \neg F_t)$ into two event-filtered regret $Reg(T, E_{t,1})$ and $Reg(T,E_{t,2})$ under events $\cN^s_t, \neg F_t$.
\begin{align}\label{apdx_eq:decompose}
    Reg(T) \le Reg(T, E_{t,1})+Reg(T, E_{t,2}) ,
\end{align}
where event $E_{t,1}=\{\Delta_{S_t} \le 2e_{t,1}(S_t)\}$, event $E_{t,2}=\{\Delta_{S_t} \le 2e_{t,2}(S_t)\}$,  $e_{t,1}(S_t)=4\sqrt{3}B_v\sqrt{\sum_{i\in \tilde{S}_t}(\frac{\log t}{T_{t-1,i}}\wedge \frac{1}{28})(p_{i}^{D,S_t})^\lambda}$,$ e_{t,2}(S_t)=28B_1\sum_{i \in \tilde{S}_t}(\frac{\log t}{T_{t-1,i}}\wedge \frac{1}{28})(p_{i}^{D,S_t})$.

\subsubsection{Bounding the $Reg(T, E_{t,1})$ term}
 We bound the leading $Reg(T, E_{t,1})$ term under two cases when $\lambda\in [1,2)$ and $\lambda \in [2,\infty)$. 

\textbf{(a) When $\lambda \in [1,2)$,} 

 Let $c_1=4\sqrt{3}$, $\tDelta_{S_t}=\Delta_{S_t}/(p_i^{D,S})^{\lambda-1}$. Given filtration $\cF_{t-1}$ and event $E_{t,1}$, we have 

\begin{align}
     \Delta_{S_t} &\overset{(a)}{\le} \sum_{i \in [m]} \frac{ 4c_1^2 B_v^2 (p_{i}^{D,S_t})^{\lambda}\frac{\log t}{T_{t-1,i}}}{\Delta_{S_t}} \\
    &\overset{(b)}{=}-\Delta_{S_t} + 2\sum_{i \in [m]} \frac{ 4c_1^2 B_v^2 (p_{i}^{D,S_t})^{\lambda}\frac{\log t}{T_{t-1,i}}}{\Delta_{S_t}}\\
    &=-\frac{\sum_{i\in \tilde{S}_t}p_i^{D,S_t}\Delta_{S_t}/(p_i^{D,S_t})^{\lambda-1}}{\sum_{i \in[m]}(p_i^{D,S_t})^{2-\lambda}} + 2\sum_{i \in [m]} \frac{ 4c_1^2 B_v^2 (p_{i}^{D,S_t})^{\lambda}\frac{\log t}{T_{t-1,i}}}{\Delta_{S_t}}\\
    &\overset{(c)}{\le} \sum_{i \in [m]}p_i^{D,S_t} \left(\frac{ 8c_1^2 B_v^2 \frac{\log t}{T_{t-1,i}}}{\Delta_{S_t}/(p_i^{D,S_t})^{\lambda-1}}-\frac{\Delta_{S_t}/(p_i^{D,S_t})^{\lambda-1}}{K}\right)\\
    &\overset{(d)}{=} \sum_{i \in [m]}p_i^{D,S_t} \left(\frac{ 8c_1^2 B_v^2 \frac{\log t}{T_{t-1,i}}}{{\tDelta}_{S_t}}-\frac{\tilde{\Delta}_{S_t}}{K}\right),\label{eq:ra_def_e1_1to2}
\end{align}
where (a) follows from event $E_{t,1}$, (b) is by the reverse amortization trick that multiplies two to both sides of (a) and rearranges the terms, (c), (d) are by definition of $K,\tDelta_{S_t}$.

It follows that

\begin{align}
    \Delta_{S_t} =\E_t[\Delta_{S_t}]  &\overset{(a)}{\le}  \E_t\left[\sum_{i \in [m]}p_i^{D,S_t}\left(\frac{ 8c_1^2 B_v^2 \frac{\log t}{T_{t-1,i}}}{\tilde{\Delta}_{S_t}}-\frac{\tilde{\Delta}_{S_t}}{K}\right)\right]\\
    &\overset{(b)}{=} \E_t\left[ \sum_{i \in \tau_t}\left(\frac{ 8c_1^2 B_v^2 \frac{\log t}{T_{t-1,i}}}{\tDelta_{S_t}}-\frac{\tDelta_{S_t}}{K}\right)\right]\\
    &\overset{(c)}{\le} \E_t\left[\sum_{i \in \tau_t}{ \kappa_{i,T}(T_{t-1,i})}\right]\label{apdx_eq:kappa_tpvm}
\end{align}
where (a) follows from \Cref{eq:ra_def_e1_1to2}, (b) follows from TPE trick to replace $p_i^{D,S_t}=\E_t[\I\{i \in \tau_t\}]$, (c) is because we define a regret allocation function
\begin{equation}
\kappa_{i,T}(\ell) = \begin{cases}
\frac{c_1^2 B_v^2} {\tilde{\Delta}_i^{\min}}, &\text{if $\ell=0$,}\\
2\sqrt{\frac{4c_1^2 B_v^2 \log T}{\ell }}, &\text{if $1\le\ell\le L_{i,T,1}$,}\\
\frac{8c_1^2 B_v^2 \log T}{\tilde{\Delta}_i^{\min} }\frac{1}{\ell}, &\text{if $L_{i,T,1} < \ell \le L_{i,T,2}$,}\\
0, &\text{if $\ell > L_{i,T,2}$,}
\end{cases}
\end{equation}
where $L_{i,T,1}=\frac{4c_1^2 B_v^2\log T}{(\tilde{\Delta}_{i}^{\min})^2}$, $L_{i,T,2}=\frac{8c_1^2 B_v^2K\log T}{(\tDelta_i^{\min})^2}$, $ \tilde{\Delta}_{i}^{\min}=\min_{S: p_i^{D,S}>0, \Delta_S>0}\Delta_{S_t}/(p_i^{D,S_t})^{\lambda-1}$, and (c) holds due to \Cref{apdx_lem:kappa_tpvm}.

\begin{align}
    Reg(T, E_{t,1})&= \E\left[\sum_{t=1}^T\Delta_{S_t}\right]\\
    &\overset{(a)}{\le} \E\left[\sum_{t\in [T]}\E_t\left[\sum_{i \in \tau_t}{ \kappa_{i,T}(T_{t-1,i})}\right]\right]\\
    &\overset{(b)}{=} \E\left[\sum_{t\in [T]}\sum_{i \in \tau_t}{ \kappa_{i,T}(T_{t-1,i})}\right]\\
    &\overset{(c)}{=}\E\left[\sum_{i \in [m]}\sum_{s=0}^{T_{T-1,i}}\kappa_{i,T}(s)\right]\\
    &\le \sum_{i \in [m]}  \frac{c_1^2 B_v^2} {\tilde{\Delta}_i^{\min}} + \sum_{i \in [m]} \sum_{s=1}^{L_{i,T,1}}2\sqrt{\frac{4c_1^2 B_v^2 \log T}{s}} + \sum_{i \in [m]}\sum_{s=L_{i,T,1}+1}^{L_{i,T,2}}\frac{8c_1^2 B_v^2 \log T}{\tilde{\Delta}_i^{\min} }\frac{1}{s}\\
    &\le \sum_{i\in [m]}\frac{c_1^2 B_v^2 }{\tDelta_i^{\min}} + \sum_{i \in [m]} \int_{s=0}^{L_{i,T,1}}2\sqrt{\frac{4c_1^2 B_v^2 \log T}{s}}\cdot ds + \sum_{i \in [m]}\int_{s=L_{i,T,1}}^{L_{i,T,2}}\frac{8c_1^2 B_v^2 \log T}{\tilde{\Delta}_i^{\min} }\frac{1}{s}\cdot ds\\
    &\le\sum_{i\in [m]}\frac{c_1^2 B_v^2 }{\tDelta_i^{\min}}+  \sum_{i \in [m]} \frac{8c_1^2 B_v^2 \log T}{\tDelta_{i}^{\min} } (3+ \log K),
\end{align}
where (a) follows from \Cref{apdx_eq:kappa_tpvm}, (b) follows from the tower rule, (c) follows from that $T_{t-1,i}$ is increased by $1$ if and only if $i \in \tau_t$.

\textbf{(b) When $\lambda \in [2,\infty)$,}

Let $\tDelta_{S_t,\lambda}=\Delta_{S_t}/(p_i^{D,S_t})^{\lambda-1}$, $\tilde{\Delta}_{S_t}=\Delta_{S_t}/p_i^{D,S_t}$. Note that $\tDelta_{S,\lambda} = \tDelta_{S}$ when $\lambda = 2$, $\tDelta_{S,\lambda} \ge \tDelta_{S}$ when $\lambda \ge 2$, and $\tDelta_{S,\lambda} \le \tDelta_{S}$, when $\lambda \le 2$, for any $i, S$. Given filtration $\cF_{t-1}$ and under event $E_{t,1}$, we have 

\begin{align}
     \Delta_{S_t} &\le \sum_{i \in [m]} \frac{ 4c_1^2 B_v^2 (p_{i}^{D,S_t})^{\lambda}\frac{\log t}{T_{t-1,i}}}{\Delta_{S_t}} \label{eq:ra_def_e1}\\
    &=-\Delta_{S_t} + 2\sum_{i \in [m]} \frac{ 4c_1^2 B_v^2 (p_{i}^{D,S_t})^{\lambda}\frac{\log t}{T_{t-1,i}}}{\Delta_{S_t}}\\
    &=-\frac{\sum_{i\in \tilde{S}_t}p_i^{D,S_t}\Delta_{S_t}/p_i^{D,S_t}}{K} + 2\sum_{i \in [m]} \frac{ 4c_1^2 B_v^2 (p_{i}^{D,S_t})^{\lambda}\frac{\log t}{T_{t-1,i}}}{\Delta_{S_t}}\\
    &\le \sum_{i \in [m]}p_i^{D,S_t} \left(\frac{ 8c_1^2 B_v^2 \frac{\log t}{T_{t-1,i}}}{\Delta_{S_t}/(p_i^{D,S_t})^{\lambda-1}}-\frac{\Delta_{S_t}/p_i^{D,S_t}}{K}\right)\\
    &= \sum_{i \in [m]}p_i^{D,S_t} \left(\frac{ 8c_1^2 B_v^2 \frac{\log t}{T_{t-1,i}}}{{\tDelta}_{S_t,\lambda}}-\frac{\tilde{\Delta}_{S_t}}{K}\right).
\end{align}

\begin{align}
    \Delta_{S_t} =\E_t[\Delta_{S_t}]  &\le  \E_t\left[\sum_{i \in [m]}p_i^{D,S_t}\left(\frac{ 8c_1^2 B_v^2 \frac{\log t}{T_{t-1,i}}}{\tilde{\Delta}_{S_t,\lambda}}-\frac{\tilde{\Delta}_{S_t}}{K}\right)\right]\\
    &= \E_t\left[ \sum_{i \in \tau_t}\left(\frac{ 8c_1^2 B_v^2 \frac{\log t}{T_{t-1,i}}}{\tDelta_{S_t,\lambda}}-\frac{\tDelta_{S_t}}{K}\right)\right]\\
    &\le \E_t\left[\sum_{i \in \tau_t}{ \kappa_{i,T}(T_{t-1,i})}\right]\label{apdx_eq:kappa_tpvm_general}
\end{align}
where the last inequality is by \Cref{apdx_lem:kappa_tpvm_general} and we define a regret allocation function 
\begin{equation}
\kappa_{i,T}(\ell) = \begin{cases}
\frac{c_1^2 B_v^2} {\tilde{\Delta}_{i,\lambda}^{\min}}, &\text{if $\ell=0$,}\\
2\sqrt{\frac{4c_1^2 B_v^2 \log T}{\ell }}, &\text{if $1\le\ell\le L_{i,T,1}$,}\\
\frac{8c_1^2 B_v^2 \log T}{\tilde{\Delta}_{i,\lambda}^{\min} }\frac{1}{\ell}, &\text{if $L_{i,T,1} < \ell \le L_{i,T,2}$,}\\
0, &\text{if $\ell > L_{i,T,2}$,}
\end{cases}
\end{equation}
where $L_{i,T,1}=\frac{4c_1^2 B_v^2\log T}{\tilde{\Delta}_{i}^{\min}\cdot \tilde{\Delta}_{i,\lambda}^{\min}}$, $L_{i,T,2}=\frac{8c_1^2 B_v^2K\log T}{\tilde{\Delta}_{i}^{\min}\cdot \tilde{\Delta}_{i,\lambda}^{\min}}$, $\tilde{\Delta}_{i}^{\min}=\min_{S: p_i^{D,S}>0, \Delta_S>0}\Delta_{S_t}/p_i^{D,S_t}$, $ \tilde{\Delta}_{i,\lambda}^{\min}=\min_{S: p_i^{D,S}>0, \Delta_S>0}\Delta_{S_t}/(p_i^{D,S_t})^{\lambda-1}$.

% \begin{align}
% \text{Reg}(T) &\le\sum_{i\in [m]}\frac{c_1^2 B_v^2 }{\tDelta_{i,\lambda}^{\min}}+  \sum_{i \in [m]} \frac{8c_1^2 B_v^2 \log T}{\tDelta_{i,\lambda}^{\min} } (1+\log K) \\
% &+ \sum_{i \in [m]} \frac{16c_1^2 B_v^2 \log T}{\sqrt{\tDelta_{i,\lambda}^{\min}\cdot\tDelta_{i}^{\min}} }.    
% \end{align}

\begin{align}
    Reg(T, E_{t,1})&= \E\left[\sum_{t=1}^T\Delta_{S_t}\right]\\
    &\overset{(a)}{\le} \E\left[\sum_{t\in [T]}\E_t\left[\sum_{i \in \tau_t}{ \kappa_{i,T}(T_{t-1,i})}\right]\right]\\
    &\overset{(b)}{=} \E\left[\sum_{t\in [T]}\sum_{i \in \tau_t}{ \kappa_{i,T}(T_{t-1,i})}\right]\\
    &\overset{(c)}{=}\E\left[\sum_{i \in [m]}\sum_{s=0}^{T_{T-1,i}}\kappa_{i,T}(s)\right]\\
    &\le \sum_{i \in [m]}  \frac{c_1^2 B_v^2} {\tilde{\Delta}_i^{\min}} + \sum_{i \in [m]} \sum_{s=1}^{L_{i,T,1}}2\sqrt{\frac{4c_1^2 B_v^2 \log T}{s}} + \sum_{i \in [m]}\sum_{s=L_{i,T,1}+1}^{L_{i,T,2}}\frac{8c_1^2 B_v^2 \log T}{\tilde{\Delta}_i^{\min} }\frac{1}{s}\\
    &\le \sum_{i\in [m]}\frac{c_1^2 B_v^2 }{\tDelta_i^{\min}} + \sum_{i \in [m]} \int_{s=0}^{L_{i,T,1}}2\sqrt{\frac{4c_1^2 B_v^2 \log T}{s}}\cdot ds + \sum_{i \in [m]}\int_{s=L_{i,T,1}}^{L_{i,T,2}}\frac{8c_1^2 B_v^2 \log T}{\tilde{\Delta}_{i,\lambda}^{\min} }\frac{1}{s}\cdot ds\\
    &\le\sum_{i\in [m]}\frac{c_1^2 B_v^2 }{\tDelta_{i,\lambda}^{\min}}+  \sum_{i \in [m]} \frac{8c_1^2 B_v^2 \log T}{\tDelta_{i,\lambda}^{\min} } (1+\log K) + \sum_{i \in [m]} \frac{16c_1^2 B_v^2 \log T}{\sqrt{\tDelta_{i,\lambda}^{\min}\cdot\tDelta_{i}^{\min}} },
\end{align}
where (a) follows from \Cref{apdx_eq:kappa_tpvm_general}, (b) follows from the tower rule, (c) follows from that $T_{t-1,i}$ is increased by $1$ if and only if $i \in \tau_t$.

\subsubsection{Bounding the $Reg(T,E_{t,2})$ term}
Let $c_2=28$. Given filtration $\cF_{t-1}$ and event $E_{t,2}$, we have 
\begin{align}
     \Delta_{S_t} &\overset{(a)}{\le} \sum_{i \in \tilde{S}_t} 2c_2 B_1 p_{i}^{D,S_t} \min\left\{1/28, \frac{\log T}{T_{t-1,i}}\right\}\notag\\
    &\overset{(b)}{\le} -\Delta_{S_t} + 2\sum_{i \in \tilde{S}_t} 2c_2 B_1 p_{i}^{D,S_t} \min\left\{1/28, \frac{\log T}{T_{t-1,i}}\right\} \notag\\
    &=-\frac{\sum_{i \in [m]}p_i^{D, S_t}\Delta_{S_t}}{\sum_{i \in [m]}p_i^{D, S_t}}+ 2\sum_{i \in[m]} 2c_2 B_1 p_{i}^{D,S_t} \min\left\{1/28, \frac{\log T}{T_{t-1,i}}\right\} \\
     &\overset{(c)}{\le}   \sum_{i \in [m]}p_i^{D,S_t}\left(-\frac{\Delta_{S_t}}{K}+ 4c_2 B_1\min\left\{1/28, \frac{\log T}{T_{t-1,i}}\right\}\right),\label{eq:ra_inf_prob_e2}
\end{align}
where (a) follows from event $E_{t,2}$, (b) is by the reverse amortization trick that multiplies two to both sides of (a) and rearranges the terms, (c) follows from $p_i^{D,S_t}\le 1$.

It follows that
\begin{align}
    \Delta_{S_t} =\E_t[\Delta_{S_t}]  &\overset{(a)}{\le}  \E_t\left[\sum_{i \in [m]}p_i^{D,S_t}\left(-\frac{\Delta_{S_t}}{K}+ 4c_2 B_1\min\left\{1/28, \frac{\log T}{T_{t-1,i}}\right\}\right)\right]\notag\\
    &\overset{(b)}{=} \E_t\left[ \sum_{i \in \tau_t}\left(-\frac{\Delta_{S_t}}{K}+ 4c_2 B_1\min\left\{1/28, \frac{\log T}{T_{t-1,i}}\right\}\right)\right]\notag\\
    &\overset{(c)}{\le} \E_t\left[\sum_{i \in \tau_t}{ \kappa_{i,T}(T_{t-1,i})}\right]\label{apdx_eq:tpvm_lambda_term_2}
\end{align}
regret allocation function 
\begin{equation}
\kappa_{i,T}(\ell) = \begin{cases}
\Delta_{i}^{\max}, &\text{if $0\le\ell\le L_{i,T,1}$}\\
\frac{4c_2 B_1 \log T}{\ell}, &\text{if $L_{i,1} < \ell \le L_{i,2}$}\\
0, &\text{if $\ell > L_{i,T,2} + 1$,}
\end{cases}
\end{equation}
where $L_{i,T,1}=\frac{4c_2 B_1\log T}{\Delta_{i}^{\max}}$, $L_{i,T,2}=\frac{4c_2 B_1 K\log T}{\Delta_i^{\min}}$.
And (a) follows from \Cref{eq:ra_inf_prob_e2}, (b) from the TPE, (c) follows from \Cref{apdx_lem:tpvm_lambda_term_2}.

\begin{align}
    Reg(T, E_{t,2})&= \E\left[\sum_{t=1}^T\Delta_{S_t}\right]\\
    &\overset{(a)}{\le} \E\left[\sum_{t\in [T]}\E_t\left[\sum_{i \in \tau_t}{ \kappa_{i,T}(T_{t-1,i})}\right]\right]\\
    &\overset{(b)}{=} \E\left[\sum_{t\in [T]}\sum_{i \in \tau_t}{ \kappa_{i,T}(T_{t-1,i})}\right]\\
    &\overset{(c)}{=}\E\left[\sum_{i \in [m]}\sum_{s=0}^{T_{T-1,i}}\kappa_{i,T}(s)\right]\\
     &\le m \Delta_{\max} + \sum_{i\in [m]}\sum_{\ell=1}^{L_{i,T,1}}\Delta_i^{\max} +  \sum_{i\in [m]} \sum_{\ell=L_{i,T,1}+1}^{L_{i,T,2}}\frac{4c_2 B_1 \log T}{\ell}\\
    &\le m \Delta_{\max} +\sum_{i\in [m]} 4c_2B_1\log T + \sum_{i\in [m]} 4c_2 B_1 \log (\frac{K\Delta_i^{\max}}{\Delta_{i}^{\min}}
    ) \log T\\
    &=m \Delta_{\max} +\sum_{i\in [m]} 4c_2 B_1 \left(1+\log (\frac{K\Delta_i^{\max}}{\Delta_{i}^{\min}})\right)\log T\\
    &\le m \Delta_{\max} + \sum_{i\in [m]} 4 c_2 B_1  \left(1+\log (\frac{K\Delta_i^{\max}}{\Delta_{i}^{\min}})\right)\log T,
\end{align}
where (a) follows from \Cref{apdx_eq:tpvm_lambda_term_2}, (b) follows from the tower rule, (c) follows from that $T_{t-1,i}$ is increased by $1$ if and only if $i \in \tau_t$.
\end{proof}

\section{Applications}\label{apdx_sec:apps}
For convenience, we show our table again in \Cref{tab:app_restate}.
\begin{table*}[ht]
	\caption{Summary of the coefficients, regret  bounds and improvements for various applications.}	\label{tab:app_restate}	
         \centering 
		\resizebox{1.00\columnwidth}{!}{
	\centering
	\begin{threeparttable}
	\begin{tabular}{|ccccc|}
		\hline
		\textbf{Application}&\textbf{Condition}& \textbf{$(B_v, B_1, \lambda)$} & \textbf{Regret}& \textbf{Improvement}\\
	\hline
  Online Influence Maximization \cite{wen2017online} & TPM &$(-,|V|,-)$ $^{\dagger}$  & $O(d|V|\sqrt{|E|T}\log T)$ & $\tilde{O}(\sqrt{|E|})$\\
	    \hline
		Disjunctive Combinatorial Cascading Bandits~\cite{li2016contextual} &  TPVM & $(1,1,1)$ & $O(d\sqrt{T}\log T)$ & $\tilde{O}(\sqrt{K}/p_{\min})$$^\ddagger$ \\
			\hline
% 			\rowcolor{LightGray}
				Conjunctive Combinatorial Cascading Bandits~\cite{li2016contextual} & TPVM & $(1,1,1)$ & $O( d\sqrt{T}\log T)$ & $\tilde{O}(\sqrt{K} /r_{\max})$\\
			\hline
   Linear Cascading Bandits~\cite{vial2022minimax}$^*$  & TPVM & $(1,1,1)$ & $O(d\sqrt{T}\log T)$ & $\tilde{O}(\sqrt{K/d})$$^\ddagger$ \\
		   \hline
				% \rowcolor{LightGray}
% 			Influence Maximization on TPG \cite{wang2017improving} & TPVM & $(\sqrt{2}|V|,|V|,1)$ & $O( \sum_{i \in [m]}\log \frac{|E|}{\Delta_{\min}}\frac{|V|^2\log |E| \log T}{\Delta_i^{\min}})$  & $O(|E|/(\log |E|\log \frac{|E|}{\Delta_{\min}}))$\\
				% \hline

      Multi-layered Network Exploration~\citep{liu2021multi} & TPVM & $(\sqrt{1.25|V|},1,2)$ $^\dagger$ & $O(d\sqrt{|V|T}\log T)$ & $\tilde{O}(\sqrt{n}/p_{\min})$\\
			\hline
	   % 	\rowcolor{LightGray}
	Probabilistic Maximum Coverage \cite{chen2013combinatorial}$^{**}$& VM & $(3\sqrt{2|V|},1,-)$ & $O(d\sqrt{|V|T}\log T)$ & $\tilde{O}(\sqrt{k})$\\
	    \hline
     	   
	\end{tabular}
	  \begin{tablenotes}[para, online,flushleft]
	\footnotesize%\smallskip
	\item[]\hspace*{-\fontdimen2\font}$^{\dagger}$ $|V|, |E|, n, k, L$ denotes the number of target nodes, the number of edges that can be triggered by the set of seed nodes, the number of layers, the number of seed nodes and the length of the longest directed path, respectively;
  \item[]\hspace*{-\fontdimen2\font}$^{\ddagger}$ K is the length of the ordered list, $r_{\max}=\alpha\cdot\max_{t\in [T], S\in \cS} r(S;\bmu_t)$;
   \item[]\hspace*{-\fontdimen2\font}$^*$ A special case of disjunctive combinatorial cascading bandits. 
 \item[]\hspace*{-\fontdimen2\font}$^{**}$ This row is for \ccmab~application and the rest of rows are for \ccmab-T applications. 
% 	\item[]\hspace*{-\fontdimen2\font}$^\ddagger$ $|V|, |E|, n, |L|$ denotes the number of layers, the number of longest;
% 	\item[]\hspace*{-\fontdimen2\font}$^\ddagger$ $L$ is the number of ;
% 	\item[]\hspace*{-\fontdimen2\font}$^{\S}$ This work gives sufficient smoothness condition with factor $\gamma_g$ and translates to $B_v=3\sqrt{2}\gamma_g$ in our setting.
	\end{tablenotes}
			\end{threeparttable}
	}
\end{table*}
\subsection{Online Influence Maximization Bandit~\cite{wang2017improving} and Its Contextual Generalization~\cite{wen2017online}}
Following the setting of \citep[Section 2.1]{wang2017improving}, we consider a weighted directed graph $G(V,E,p)$, where $V$ is the set of vertices, $E$ is the set of directed edges, and each edge $(u,v)\in E$ is associated with a probability $p(u,v)\in [0,1]$. When the agent selects a set of seed nodes $S \subseteq V$, the influence propagates as follows: At time $0$, the seed nodes $S$ are activated; At time $t > 1$, a node $u$ activated at time $t-1$ will have one chance to activate its inactive out-neighbor $v$ with independent probability $p(u,v)$. The influence spread of $S$ is denoted as $\sigma(S)$ and is defined as the expected number of activated nodes after the propagation process ends. The problem of Influence Maximization is to find seed nodes $S$ with $|S|\le k$ so that the influence spread $\sigma(S)$ is maximized. 

For the problem of online influence maximization (OIM), we consider $T$ rounds repeated influence maximization tasks and the edge probabilities $p(u,v)$ are assumed to be unknown initially. For each round $t\in [T]$, the agent selects $k$ seed nodes as $S_t$, the influence propagation of $S_t$ is observed and the reward is the number of nodes activated in round $t$. The agent's goal is to accumulate as much reward as possible in $T$ rounds. The OIM fits into CMAB-T framework: the edges $E$ are the set of base arms $[m]$, the (unknown) outcome distribution $D$ is the joint of $m$ independent Bernoulli random variables for the edge set $E$, the action $S$ are any seed node sets with size $k$ at most $k$. For the arm triggering, the triggered set $\tau_t$ is the set of edges $(u,v)$ whose source node $u$ is reachable from $S_t$. Let $X_t$ be the outcomes of the edges $E$ according to probability $p(u,v)$ and the live-edge graph $G_t^{\text{live}}(V,E^{\text{live}})$ be a induced graph with edges that are alive, i.e., $e \in E^{\text{live}}$ iff $X_{t,e}=1$ for $e \in E$. The triggering probability distribution $D_{\text{trig}}(S_t,X_t)$ degenerates to a deterministic triggered set, i.e., $\tau_t$ is deterministically decided given $S_t$ and $X_t$. The reward $R(S_t, X_t, \tau_t)$ equals to the number activated nodes at the end of $t$, i.e., the nodes that are reachable from $S_t$ in the live-edge graph $G_t^{\text{live}}$. The offline oracle is a $(1-1/e-\varepsilon, 1/|V|)$-approximation algorithm given by the greedy algorithm from~\cite{kempe2003maximizing}.

Now consider OIM's contextual generalization for large-scale OIM, we follow~\citet{wen2017online}, where each edge $e=(u,v)$ is associated with a known feature vector $\bx_e \in \R^d$ and an unknown parameter $\btheta^* \in \R^d$, and the edge probability is $p(u,v)=\inner{\bx_e, \btheta^*}$. By Lemma 2 of \cite{wang2017improving}, $B_1=\tilde{C}\le |V|$, where $\tilde{C}$ is the largest number of nodes any node can reach and batch size $K\le |E|$, so by \Cref{thm:c2mab_reg}, C$^2$-UCB-T obtain a worst-case $\tilde{O}(d|V|\sqrt{|E|T})$ regret bound. Compared with IMLinUCB algorithm \cite{wen2017online} that achieves $\tilde{O}(d(|V|-k)|E|\sqrt{T})$, our regret achieves a improvement up to a factor of $\tilde{O}(\sqrt{|E|})$.

Now for the claim of the triggering probability satisfies $B_p=B_1$, it follows from the Theorem 4 of  \citet{li2020online} by identifying $f(S,w,v)=p_i^{w, S}$.

\subsection {Contextual Combinatorial Cascading Bandits~\cite{li2016contextual}}\label{sec:cascading}
Contextual Combinatorial cascading bandits have two categories: conjunctive cascading bandits and disjunctive cascading bandits~\cite{li2016contextual}. We also compare with a special case of linear cascading bandits that also uses variance-adaptive algorithms and achieve very competitive results.

\textbf{Disjunctive form.}
For the disjunctive form, we want to select an ordered list $S$ of $K$ items from total $L$ items, so as to maximize the probability that at least one of the outcomes of the selected items are $1$.
Each item is associated with a Bernoulli random variable with mean $\mu_{t,i} \in [0,1]$ at round $t$, indicating whether the user will be satisfied with the item if he scans the item. To leverage the contextual information, \citet{li2016contextual} assumes $\bmu_{t,i}=\inner{\bx_{t,i}, \btheta^*},$ where $\bx_{t,i} \in \R^d$ is the known context at round $t$ for arm $i$, $\btheta \in \R^d$ is the unknown parameter to be learned.
This setting models the movie recommendation system where the user sequentially scans a list of recommended items and the system is rewarded when the user is satisfied with any recommended item.
After the user is satisfied with any item or scans all $K$ items but is not satisfied with any of them, the user leaves the system.
Due to this stopping rule, the agent can only observe the outcome of items until (including) the first item whose outcome is $1$. If there are no satisfactory items, the outcomes must be all $0$. In other words, the triggered set is the prefix set of items until the stopping condition holds.

Without loss of generality, let the action be $\{1,...,K\}$, then the reward function is $r(S;\bmu)=1-\prod_{j=1}^K(1-\mu_j)$ and the triggering probability is $p_{i}^{\bmu,S}=\prod_{j=1}^{i-1}(1-\mu_j)$.
Let $\bar{\bmu}=(\bar{\mu}_1, ..., \bar{\mu}_K)$ and $\bmu=(\mu_1, ..., \mu_K)$, where $\bar{\bmu}=\bmu+\boldzeta + \boldeta$ with $\bar{\bmu}, \bmu \in (0,1)^K, \boldzeta, \boldeta \in [-1,1]^K$. By Lemma 19 in \citet{liu2021bandit}, disjunctive CB satisfies Condition~\ref{cond:TPVMm} with $(B_v, B_1, \lambda)=(1,1,1)$. Also, we can verify that disjunctive CB also satisfies  $B_p=B_1=1$ as follows: 
\begin{align}
    &\abs{p_i^{\bar{\bmu},S}-p_i^{\bmu,S}}\notag\\
    &=\abs{\prod_{j=1}^i(1-\mu_j)-\prod_{j=1}^i(1-\bar{\mu}_j)}\\
    &=\sum_{j=1}^i \abs{\bar{\mu}_j-\mu_j}(1-\mu_1)...(1-\mu_{j-1}) (1-\bar{\mu}_{j+1})...(1-\bar{\mu}_{i})\\
    &\le \sum_{j=1}^i \abs{\bar{\mu}_j-\mu_j}(1-\mu_1)...(1-\mu_{j-1})\\
    &=\sum_{j=1}^i \abs{\bar{\mu}_j-\mu_j}p_j^{\bmu, S}.
\end{align}
Now by \Cref{thm:ccmab_reg_TPVM_lambda}, VAC$^2$-UCB obtains a regret bound of $O(d\sqrt{T}\log T)$. Compared with Corollary 4.5 in \citet{li2016contextual} that yields a $O(d\sqrt{KT}\log T/p_{\min})$ regret, our results improves by a factor of $O(\sqrt{K}/p_{\min})$.

\textbf{Conjunctive form.}
For the conjunctive form, the learning agent wants to select $K$ paths from total $L$ paths (i.e., base arms) so as to maximize the probability that the outcomes of the selected paths are all $1$.
Each item is associated with a Bernoulli random variable with mean $\mu_{t,i}$ at round $t$, indicating whether the path will be live if the package will transmit via this path.
% The outcome of the selected items are observed only before (or at) the first position whose outcome is 0.
Such a setting models the network routing problem~\cite{kveton2015combinatorial}, where the items are routing paths and the package is delivered when all paths are alive. 
The learning agent will observe the outcome of the first few paths till the first one that is down, since the transmission will stop if any of the path is down. In other words, the triggered set is the prefix set of paths until the stopping condition holds. 

Without loss of generality, let the action be $\{1,...,K\}$, then the reward function is $r(S;\bmu)=1-\prod_{j=1}^K(\mu_j)$ and the triggering probability is $p_{i}^{\bmu,S}=\prod_{j=1}^{i-1}(\mu_j)$.
Let $\bar{\bmu}=(\bar{\mu}_1, ..., \bar{\mu}_K)$ and $\bmu=(\mu_1, ..., \mu_K)$, where $\bar{\bmu}=\bmu+\boldzeta + \boldeta$ with $\bar{\bmu}, \bmu \in (0,1)^K, \boldzeta, \boldeta \in [-1,1]^K$. By Lemma 20 in \citet{liu2021bandit}, conjunctive CB satisfies Condition~\ref{cond:TPVMm} with $(B_v, B_1,\lambda)=(1,1,1)$. Also, we can verify that conjunctive CB also satisfies $B_p=B_1=1$ as follows: 
\begin{align}
    &\abs{p_i^{\bar{\bmu},S}-p_i^{\bmu,S}}\notag\\
    &=\abs{\prod_{j=1}^i\mu_j-\prod_{j=1}^i\bar{\mu}_j}\\
    &=\sum_{j=1}^i \abs{\bar{\mu}_j-\mu_j}(\mu_1)...(\mu_{j-1}) (\bar{\mu}_{j+1})...(\bar{\mu}_{i})\\
    &\le \sum_{j=1}^i \abs{\bar{\mu}_j-\mu_j}(\mu_1)...(\mu_{j-1})\\
    &=\sum_{j=1}^i \abs{\bar{\mu}_j-\mu_j}p_j^{\bmu, S}.
\end{align}
Now by \Cref{thm:ccmab_reg_TPVM_lambda}, VAC$^2$-UCB obtains a regret bound of $O(d\sqrt{T}\log T)$. Compared with Corollary 4.6 in \citet{li2016contextual} that yields a $O(d\sqrt{KT}\log T/r_{\max})$ regret, our results improves by a factor of $O(\sqrt{K}/r_{\max})$.

\textbf{Linear Cascading Bandit.} Linear cascading bandit~\cite{vial2022minimax} is a special case of combinatorial cascading bandit~\cite{li2016contextual}. The former assumes that action space $\mathcal{S}$ is the collection of all permutations whose size equals to $K$ (i.e., a uniform matroid). In this case, the items in the feasible solutions are \textbf{exchangeable} (a critical property for matroids), i.e., $S - \{e_1\} + \{e_2\} \in \mathcal{S}$, for any $S \in \mathcal{S}, e_1, e_2 \in [m]$. Based on this property, their analysis can get the correct results. For the latter, however, $\mathcal{S}$ (i.e., $\Theta$ in [16]) consists of arbitrary feasible actions (perhaps with different sizes), e.g., $S \in \mathcal{S}$ could refer to any path that connects the source and the destination in network routing applications.

Other than the above difference, linear cascading bandits follow the same setting as disjunctive contextual combinatorial bandits. Following the similar argument of disjunctive contextual combinatorial bandits, the regret bound of VAC$^2$-UCB is $O(d\sqrt{T}\log T)$. Compared with CascadeWOFUL that achieves $\tilde{O}(\sqrt{d(d+K)T})$ by Theorem 4 in \citet{vial2022minimax}, our regret improves a factor of $\tilde{O}(\sqrt{1+K/d})$. For the empirical comparison, see \Cref{sec:app} for details.

\subsection{Multi-layered Network Exploration Problem (MuLaNE)~\cite{liu2021multi}}
We consider the MuLaNE problem with random node weights.
After we apply the bipartite coverage graph, the corresponding graph is a tri-partite graph $(n,V,R)$ (i.e., a 3-layered graph where the first layer and the second layer forms a bipartite graph, and the second and the third layer forms another bipartite graph), where the left nodes represent $n$ random walkers; Middle nodes are $|V|$ possible targets $V$ to be explored; Right nodes $R$ are $V$ nodes, each of which has only one edge connecting the middle edge. The MuLaNE task is to allocate $B$ budgets into $n$ layers to explore target nodes $V$ and the base arms are $\cA=\{(i,u,b): i \in [n], u \in V, b \in [B]\}$.

With budget allocation $k_1, ..., k_L$, the (effective) base arms consist of two parts: 

(1) $\{(i,j): i\in[n], j \in V\}$, each of which is associated with visiting probability $x_{i,j}\in[0,1]$ indicating whether node $j$ will be visited by explorer $i$ given $k_i$ budgets. All these base arms corresponds to budget $k_i, i\in [n]$ are triggered.

(2) $y_j\in[0,1]$ for $j \in V$ represents the random node weight. The triggering probability $p_{j}^{\bmu, S}=1-\prod_{i \in [n]}{(1-x_{i,j})}$.

For its contextual generalization, we assume $x_{i,j}=\inner{\bphi_x(i,j), \btheta^*}$, $y_{j}=\inner{\bphi_y(j), \btheta^*}$, where $\bphi_x(i,j), \bphi_y(j)$ are the known features for visiting probability and the node weights for large-scale MuLaNE applications, respectively. Let effective base arms $\bmu=(\bx,\by) \in (0,1)^{(n|V|+|V|)}, \bar{\bmu}=(\bar{\bx}, \bar{\by})\in (0,1)^{(n|V|+|V|)}$, where $\bar{\bx}=\boldzeta_{x}+\boldeta_{x}+\bx, \bar{\by}=\boldzeta_{y}+\boldeta_{y}+\by$, for $\boldzeta, \boldeta \in [-1,1]^{(n|V|+|V|)}$.
For the target node $j \in V$, the per-target reward function $r_j(S;\bx,\by)=y_j(1-\prod_{i \in [n]}(1-x_{i,j}))$.
Denote $\bar{p}_j^{\bmu,S}=1-\prod_{i \in [n]}{(1-\bar{x}_{i,j})}$. Based on Lemma 21 in \citet{liu2022batch}, contextual MuLaNE satisfies Condition~\ref{cond:TPVMm} with $(B_v,B_1,\lambda)=(\sqrt{1.25|V|},1,2)$. To validate that this application satisfies Condition~\ref{cond:TPM_P} with $B_p=B_1=1$, we have
\begin{align}
    &\abs{p_j^{\bar{\bmu},S}-p_j^{\bmu,S}}\notag\\
    &=\abs{\prod_{i\in [n]}(1-x_{i,j})-\prod_{i \in [n]}(1-\bar{x}_{i,j})}\\
    &=\sum_{i\in [n]} \abs{\bar{x}_{i,j}-x_{i,j}}(1-x_{1,j})...(1-x_{i-1,j}) (1-\bar{x}_{i+1,j})...(1-\bar{x}_{i,j})\\
    &=\sum_{i\in [n]} \abs{\bar{x}_{i,j}-x_{i,j}}.
\end{align}
By \Cref{thm:ccmab_reg_TPVM_lambda}, we obtain $O(d\sqrt{|V|T}\log T)$, which improves the result $O(d\sqrt{n|V|T}\log T/p_{\min})$ that follows the result of C$^3$UCB algorithm \cite{li2016contextual} by a factor of $O(\sqrt{n/p_{\min}})$.

\subsection{Probabilistic Maximum Coverage Bandit~\cite{chen2016combinatorial,merlis2019batch}}
In this section, we consider the probabilistic maximum coverage (PMC) problem. PMC is modeled by a weighted bipartite graph $G=(L,V,E)$, where $L$ are the source nodes, $V$ is the target nodes and each edge $(u,v) \in E$ is associated with a probability $p(u,v)$. The task of PMC is to select a set $S \subseteq L$ of size $k$ so as to maximize the expected number of nodes activated in $V$, where a node $v \in V$ can be activated by a node $u \in S$ with an independent probability $p(u,v)$. PMC can naturally model the advertisement placement application, where $L$ are candidate web-pages, $V$ are the set of users, and $p(u,v)$ is the probability that a user $v$ will click on web-page $u$. 

PMC fits into the non-triggering CMAB framework: each edge $(u,v) \in E$ corresponds to a base arm, the action is the set of edges that are incident to the set $S \subseteq L$, the unknown mean vectors $\bmu \in (0,1)^E$ with $\mu_{u,v}=p(u,v)$ and we assume they are independent across all base arms. In this context, the reward function $r(S;\bmu)=\sum_{v\in V}(1-\prod_{u \in S}(1-\mu_{u,v}))$.

In this paper, we consider a contextual generalization by assuming that $p(u,v)=\inner{\bphi(u,v), \btheta^*}$, where $\bphi(u,v)\in \R^d$ is the known context and $\btheta^*\in \R^d$ is the unknown parameter. By Lemma 24 in \citet{liu2022batch}, PMC satisfies Condition~\ref{cond:VM} with $(B_v,B_1)=(3\sqrt{2|V|},1)$. Following \Cref{thm:ccmab_reg_VM}, VAC$^2$-UCB obtains $O(d\sqrt{|V|T}\log T)$, which improves the C$^3$UCB algorithm's bound $O(d\sqrt{k|V|T}\log T)$ \cite{li2016contextual} by a factor of $O(\sqrt{k})$.

\section{Experiments}\label{sec:apps-exp}
\begin{figure}[ht]
    \centering
    \includegraphics[width=1\linewidth]{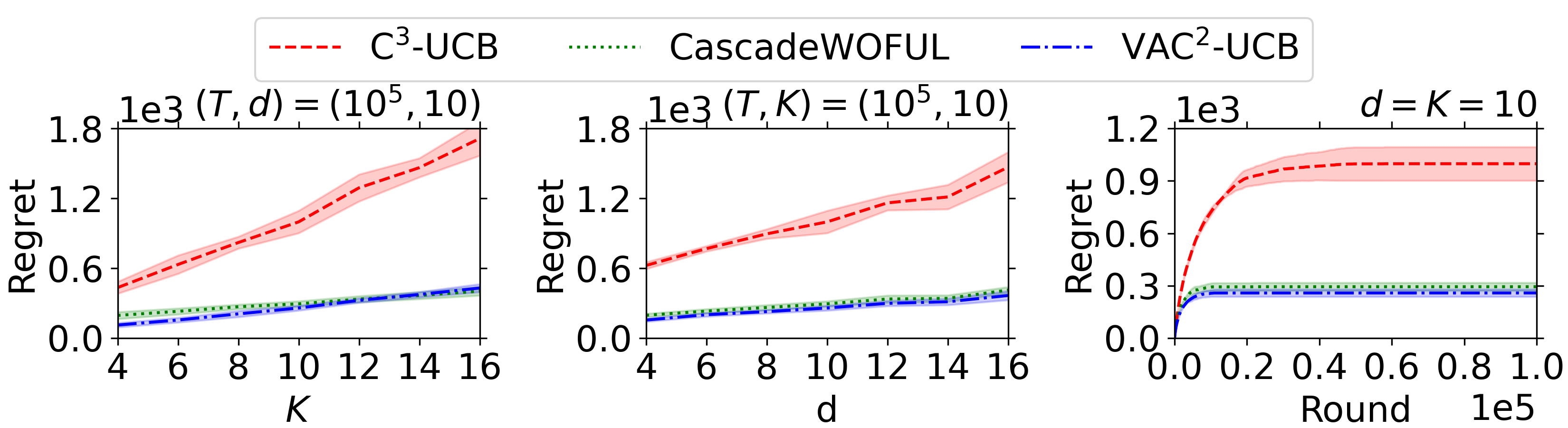}
    \caption{Results for synthetic data}
    \label{fig:syn_lin}
\end{figure}

% (essentially the same as C$^3$UCB~\cite{li2016contextual} by setting the tunable parameter $\sigma=1$)
\textbf{Synthetic data}. We consider the same disjunctive linear cascading bandit setting as in \cite{vial2022minimax}, where the goal is to choose $K\in\{2i\}_{i=2}^8$ out of $m=100$ items to maximize the reward. Notice that the linear cascading bandit problem is a simplified version of the contextual cascading bandit problem where the feature vectors of base arms are fixed in all rounds (see \Cref{sec:cascading} for details).
For each $K$, we sample the click probability $\mu_i$ of item $i$ uniformly in $[\frac{2}{3K},\frac{1}{K}]$ for $i\le K$ and in $[0, \frac{1}{3K}]$ for $i>K$. We vary $d\in\{2i\}_{i=2}^{8}$ to generate the same $\bm{\mu}$ and compute unit-norm vectors $\theta^*$ and $\phi(i)$ satisfying $\mu_i = \langle\theta^*, \phi(i)\rangle$. We compare VAC$^2$-UCB to C$^3$-UCB~\cite{li2016contextual} and CascadeWOFUL~\cite{vial2022minimax}: C$^3$-UCB is the variance-agnostic cascading bandit algorithm (essentially the same as CascadeLinUCB~\cite{zong2016cascading} in the linear cascading setting by using the tunable parameter $\sigma=1$) and CascadeWOFUL is the state-of-the-art variance-aware cascading bandit algorithm. 
As shown in Figure~\ref{fig:syn_lin}, the regret of our VAC$^2$-UCB algorithm has superior dependence on $K$ and $d$ over that of C$^3$-UCB. When $d=K=10$, VAC$^2$-UCB achieves sublinear regret; it incurs $75\%$ and $13\%$ less regret than C$^3$-UCB and  CascadeWOFUL after $100,000$ rounds. Notice that CascadeWOFUL is also variance-aware but specifically designed for cascading bandits, while our algorithm can be applied to general C$^2$MAB-T.

\textbf{Real data}. We conduct experiments on the MovieLens-1M dataset which contains user ratings for $m\approx4000$ movies. Following the same experimental setup in~\cite{vial2022minimax}, we set $d = 20, K = 4$, and the goal is to choose $K$ out of $m$ movies to maximize the reward of the cascading recommendation. We use their learned feature mapping $\phi$ from movies to the probability that a uniformly random user rated the movie more than three stars. We point the reader to Section 6 of \cite{vial2022minimax} for more details.
In each round, we sample a random user $J_t$ and define the potential click result $X_{t,i} = \I\{\text{user $J_t$ rated movie $i$ more than 3 stars}\}$. In other words, we observe the actual feedback of user $J_t$ instead of using the Bernoulli click model.
\Cref{fig:real_all} shows that VAC$^2$-UCB outperforms C$^3$-UCB and CascadeWOFUL, incurring $45\%$ and $25\%$ less regret after $100,000$ rounds. To model platforms like Netflix that recommend movies in specific categories, we also run experiments while restricting the candidate items to movies of a particular genre. \Cref{fig:real_genre} shows that VAC$^2$-UCB is superior for all genres compared to C$^3$-UCB and CascadeWOFUL.

 \section{Concentration Bounds, Facts, and Technical Lemmas}\label{apdx_sec:Techs}
 In this section, we first give key concentration bounds and then provide lemmas that are useful for the analysis. 
 \subsection{Concentration Bounds}
 We mainly use the following concentration bounds, which is essentially a modification of the Freedman's version of the Bernstein's inequality~\cite{Bernstein1946bandit,freedman1975tail}.
 \begin{proposition}
[Theorem 9, \citet{lattimore2015linear}]\label{apdx_lem:concen_unknown}
     Let $\delta \in (0,1)$ and $X_1, ..., X_n$ be a sequence of random variables adapted to filtration $\{\cF_t\}$ with $\E[X_t\mid \cF_{t-1}]=0$. Let $Z \subseteq [n]$ be such that $\I\{t \in Z\}$ is $\cF_{t-1}$-measurable and let $R_t$ be $\cF_{t-1}$ measurable such that $|X_t|\le R_t$ almost surely. Let $V=\sum_{t \in Z}\Var[X_t\mid\cF_{t-1}]+\sum_{t \notin Z}R_t^2/2$, $R=\max_{t \in Z}R_t$, and $S=\sum_{t=1}^nX_t$. Then $\Pr[S \ge f(R,V)]\le \delta$, where $f(r,v)=\frac{2(r+1)}{3}\log \frac{2}{\delta_{r,v}}+\sqrt{2(v+1)\log \frac{2}{\delta_{r,v}}}$, and $\delta_{r,v}=\frac{\delta}{3(r+1)^2(v+1)}$.
 \end{proposition}
 
\subsection{Facts}

\begin{fact}\label{apdx_fact:matrix}
For any positive-definite matrices $\bA, \bB >\boldsymbol{0}^{d}$ and any vectors $\bx, \by \in \R^d$. It holds that
\begin{enumerate}
\item If $\bA \le \bB$, then $\bA^{-1} \ge \bB^{-1}$.
\item If $\bA \le \bB$, then $\norm{\bx}_{\bA}\le \norm{\bx}_{\bB}$.
\item Suppose $\bA$ has maximum eigenvalue $\lambda_{\max}$, then $\norm{\bA\bx}_2\le \lambda_{\max}\cdot \norm{\bx}_2$ and $\lambda_{\max}\le \trace(\bA)$.
\end{enumerate}
\end{fact}

\subsection{Technical Lemmas}
Recall that event $F_t$ is defined in~\Cref{apdx_eq:event_F}, gram matrix $\bG_t$ is defined in~\Cref{apdx_eq:def_G_b_theta_VM}, optimistic variance $\bar{V}_{t,i}$ is defined in~\Cref{eq:var_upper}, $R_{\bv}$ is defined in~\Cref{apdx_eq:R}.

\begin{lemma}\label{apdx_lem:reduction_key}
$\Pr\left[\norm{\bZ_{t}}_{\bG_{t}}+\sqrt{\gamma}\ge \rho \text{ and } \neg F_{t-1}\right]\le \delta/T$, for $t=1, ..., T$.
\end{lemma}

\begin{proof}[Proof of \cref{apdx_lem:reduction_key}]
Let $\bv \in \R^d$ and define 
\begin{equation}\label{apdx_eq:var_upper}
V_{s,i, \bv}= \begin{cases}
\Var[\eta_{s,i}\mid \cF_{s-1}] \inner{\bphi_s(i), \bv}^2/\bar{V}_{s,i}^{2}, &\text{if $\bar{V}_{s,i}<\frac{1}{4}$}\\
 \inner{\bphi_s(i), \bv}^2/\bar{V}_{s,i}, &\text{otherwise}.
\end{cases}
\end{equation} 
\begin{equation}\label{apdx_eq:R}
    R_{\bv}=\max_{s< t, i \in \tau_s}\{ \inner{\bphi_s(i), \bv}/\bar{V}_{s,i}: \bar{V}_{s,i}<\frac{1}{4}\}
\end{equation}
By applying the \Cref{apdx_lem:concen_unknown}, with probability at least $1-\delta/T$ it holds that
\begin{align}\label{apdx_eq:va_Z}
    \inner{Z_{t}, \bv}=\sum_{s<t}\sum_{i \in \tau_s}\eta_{s,i}\inner{\bphi_s(i), \bv}/\bar{V}_{s,i}\le \frac{2(R_{\bv}+1)}{3}\log \frac{1}{\delta_{\bv}}+\sqrt{2(1+\sum_{s<t}\sum_{i \in \tau_s}V_{s,i,\bv})\log \frac{1}{\delta_{\bv}}}
\end{align}
where  $\delta_{\bv}=\frac{3\delta}{T(1+R_{\bv})^2(1+\sum_{s<t}\sum_{i \in \tau_s}V_{s,i,
\bv})^2}$.

Since $\bv$ could be a random variable in later proofs, we use the covering argument trick (Chap.20, ~\citet{lattimore2020bandit}) to handle $\bv$. Specifically, we define the covering set $\Lambda=\{j\cdot \varepsilon: j=-\frac{C}{\varepsilon}, -\frac{C}{\varepsilon}+1, ..., \frac{C}{\varepsilon}-1, \frac{C}{\varepsilon}\}^d$, with size $N=|\Lambda|=(2C/\varepsilon)^d$ and parameters $C,\varepsilon$ will be determined shortly after. By applying union bound on \Cref{apdx_eq:va_Z}, we have with probability at least $1-\delta$ that 
\begin{align}
\inner{Z_{t}, \bv}\le \frac{2(R_{\bv}+1)}{3}\log \frac{N}{\delta_{\bv}}+\sqrt{2(1+\sum_{s<t}\sum_{i \in \tau_s}V_{s,i,\bv})\log \frac{N}{\delta_{\bv}}} \text{ for all } \bv \in \Lambda.
\end{align}
Now we can set $\bv=\bG_t^{-1}\bZ_t$, and it follows from \Cref{apdx_lem:l1_norm} that $\norm{\bv}_{\infty} \le \norm{\bZ_t}_{1} \le  2dK^2t^2=C$. Based on our construction of the covering set $\Lambda$, there exists $\bv'\in \Lambda$ with $\bv'\le \bv$, and $\norm{\bv'-\bv}_{\infty}\le \varepsilon$, such that 
\begin{align}\label{apdx_eq:Zt_Gt}
\norm{\bZ_t}^2_{\bG_t^{-1}}&=\inner{\bZ_t, \bv}\le \norm{\bZ_t}_1 \varepsilon + \inner{\bZ_t, \bv'}\\
&\le \norm{\bZ_t}_1 \varepsilon +  \frac{2(R_{\bv}+1)}{3}\log \frac{N}{\delta_{\bv}}+\sqrt{2(1+\sum_{s<t}\sum_{i \in \tau_s}V_{s,i,\bv})\log \frac{N}{\delta_{\bv}}}\label{apdx_eq:Zt_Gt_1}\\
&\le \norm{\bZ_t}_1 \varepsilon +  \frac{2(R_{\bv}+1)}{3}\log \frac{N}{\delta_{\bv}}+\sqrt{2(1+\norm{\bZ_t}_{\bG_t^{-1}}^2)\log \frac{N}{\delta_{\bv}}}\label{apdx_eq:Zt_Gt_2}
\end{align}
where \Cref{apdx_eq:Zt_Gt_1} uses the fact that $R_{\bv'}\le R_{\bv}, V_{s,i,\bv'}\le V_{s,i, \bv}, \frac{1}{\delta_{\bv'}}\le \frac{1}{\delta_{\bv}}$ for any $\bv'\le \bv$, \Cref{apdx_eq:Zt_Gt_2} follows from the following derivation,
\begin{align}
    \sum_{s<t}\sum_{i \in \tau_s}V_{s,i,\bv}&\le \sum_{s<t}\sum_{i \in \tau_s}\inner{\bphi_s(i), \bv}^2/\bar{V}_{s,i}\label{apdx_eq:sum_var_bound_1}\\
    &=\sum_{s<t}\sum_{i \in \tau_s}(\bG_t^{-1}\bZ_t)^{\top}\bphi_s(i)\bphi_s(i)^{\top} \bG_{t}^{-1}\bZ_t/\bar{V}_{s,i}\label{apdx_eq:sum_var_bound_2}\\
    &=(\bG_t^{-1}\bZ_t)^{\top}\left(\sum_{s<t}\sum_{i \in \tau_s}\bphi_s(i)\bphi_s(i)^{\top}/\bar{V}_{s,i}\right) \bG_{t}^{-1}\bZ_t\\
    &\le (\bG_t^{-1}\bZ_t)^{\top} \bG_t (\bG_t^{-1}\bZ_t)\label{apdx_eq:sum_var_bound_3}\\
    &=\norm{\bZ_t}_{\bG_t^{-1}}^2\label{apdx_eq:sum_var_bound_4},
    \end{align}
where \Cref{apdx_eq:sum_var_bound_1} follows from $\neg F_{s-1}$ implies $\norm{\btheta^*-\hat{\btheta}_s}_{\bG_{s}} \le \rho$ for $s<t$ by \Cref{apdx_lem:event2radius} and thus $\bar{V}_{t,i} \ge \Var[\eta_{s,i}\mid \cF_{s-1}]$, \Cref{apdx_eq:sum_var_bound_2} follows from definition of $\bv$, \Cref{apdx_eq:sum_var_bound_3} follows from $\sum_{s<t}\sum_{i \in \tau_s}\bphi_s(i)\bphi_s(i)^{\top}/\bar{V}_{s,i} < \bG_t$.

Now we set $\varepsilon=1/C=1/(2K^2t^2d)$, we have
\begin{align}
    \norm{\bZ_t}^2_{\bG_t^{-1}} &\le \text{RHS of \Cref{apdx_eq:Zt_Gt_2}}\\
    &\le C \varepsilon +  \frac{2(2\norm{\bZ_t}_{\bG_t^{-1}}/\rho+1)}{3}\log \frac{N}{\delta_{\bv}}+\sqrt{2(1+\norm{\bZ_t}_{\bG_t^{-1}}^2)\log \frac{N}{\delta_{\bv}}}\label{apdx_eq:before_math_0}\\
    &\le 1+ 2\log \frac{N}{\delta_{\bv}}+\sqrt{2(1+\norm{\bZ_t}_{\bG_t^{-1}}^2)\log \frac{N}{\delta_{\bv}}}\label{apdx_eq:before_math}
\end{align}
where \Cref{apdx_eq:before_math_0} is to bound $R_{\bv}$ by \Cref{apdx_lem:R}, \Cref{apdx_eq:before_math} is by the definition of $\rho$ as an upper bound of $\norm{\bZ_t}_{\bG_t^{-1}}$.

By rearranging and simplifying \Cref{apdx_eq:before_math}, we have
\begin{align}
    \norm{\bZ_t}_{\bG_t^{-1}} + \sqrt{\gamma} &\le 1+ \sqrt{\gamma} + 4\sqrt{\log \frac{N}{\delta_{\bv}}}\\
    &\le1 + \sqrt{\gamma} + 4\sqrt{\log \left(\frac{6TN}{\delta}(1+\norm{\bZ_t}_{\bG_t^{-1}}^2)\right)},
\end{align}
where the last inequality is because of $\delta_{\bv}\ge \frac{3\delta}{T(1+\norm{\bZ_t}_{\bG_t^{-1}}^2)}$ from the definition of $\delta_{\bv}$, \Cref{apdx_lem:R}, and \Cref{apdx_eq:sum_var_bound_4}. 
Finally, we solve the above equation and set $\rho=1 + \sqrt{\gamma} + 4\sqrt{\log \left(\frac{6TN}{\delta}\log(\frac{3TN}{\delta})\right)}$ , which completes the reduction on $t$ to show the probability $ \Pr[\norm{\bZ_t}_{\bG_t^{-1}} + \sqrt{\gamma} \ge \rho]\ge 1-\delta/T$ under event $\neg F_{t-1}$.
\end{proof}

\begin{lemma}\label{apdx_lem:event2radius}
If $\neg F_{t}$ holds, then it holds that,
\begin{align}\label{apdx_eq:event2radius}
\norm{\btheta^*-\hat{\btheta}_t}_{\bG_{t}} \le \rho.
\end{align}
\end{lemma}
\begin{proof}
We have 
\begin{align}
\norm{\btheta^*-\hat{\btheta}_t}_{\bG_{t}}&=\norm{\bG_t^{-1}(\sum_{s<t}\sum_{i \in \tau_s}\bphi_s(i)X_{s,i}/\bar{V}_{s,i})-\bG_t^{-1}\bG_t \btheta^*}_{\bG_t}\label{apdx_eq:event2radius_proof_1}\\
&=\norm{\bG_t^{-1}\bZ_t + \bG_t^{-1}(\sum_{s<t}\sum_{i \in \tau_s}\bphi_s(i)\bphi_s(i)^{\top} \btheta^*/\bar{V}_{s,i})-\bG_t^{-1}\bG_t \btheta^*}_{\bG_t}\label{apdx_eq:event2radius_proof_2}\\
&=\norm{\bG_t^{-1}\bZ_t - \gamma \bG_t^{-1}\btheta^*}_{\bG_t}\label{apdx_eq:event2radius_proof_3}\\
&\le \norm{\bZ_t}_{\bG_t^{-1}} + \gamma \norm{\btheta^*}_{\bG_t^{-1}}\label{apdx_eq:event2radius_proof_4}\\
&\le \norm{\bZ_t}_{\bG_t^{-1}} + \sqrt{\gamma} \label{apdx_eq:event2radius_proof_5}\\
&\le \rho - \sqrt{\gamma} + \sqrt{\gamma} = \rho,
\end{align}
where Equation (\ref{apdx_eq:event2radius_proof_1})-(\ref{apdx_eq:event2radius_proof_3}) follow from definition and math calculation, \Cref{apdx_eq:event2radius_proof_4} from $\bG_t \ge \bG_0=\gamma \bI$ and $\norm{\btheta}_2\le 1$, \Cref{apdx_eq:event2radius_proof_5} from that if $\neg F_{t}$ holds, then $\norm{\bZ_t}_{\bG_{t}}+\sqrt{\gamma}\le \rho$.
\end{proof}

\begin{lemma}\label{apdx_lem:beta_and_v_norm_potential}
For any $s < t$, $\frac{\norm{\bphi_s(i)}_{\bG_t^{-1}}}{\bar{V}_{s,i}}\le \frac{\norm{\bphi_s(i)}_{\bG_s^{-1}}}{\bar{V}_{s,i}}$, and if $\neg F_{t-1}$ holds and $\bar{V}_{s,i} < \frac{1}{4}$, $\frac{\norm{\bphi_s(i)}_{\bG_s^{-1}}}{\bar{V}_{s,i}} \le \frac{2}{\rho}\le 1$  for any $i \in [m]$.
\end{lemma}
\begin{proof}
The first inequality is by $\bG_t \ge \bG_s$ and Fact~\ref{apdx_fact:matrix}. For the second inequality, when $\neg F_{t-1}$ holds, $\norm{\btheta^*-\hat{\btheta}_s}_{\bG_{s}} \le \rho$ as in \Cref{apdx_eq:event2radius}, and since $\bar{V}_{s,i} < \frac{1}{4}$, it follows from the definition of $\bar{V}_{s,i}$~\Cref{eq:var_upper} that at least one of the following is true:
    \begin{align}
        \bar{V}_{s,i}&\ge \frac{1}{2} (\inner{\bphi_s(i), \hat{\btheta}_s+2 \rho \norm{\bphi_s(i)}_{\bG_s^{-1}}})\ge \rho \norm{\bphi_s(i)}_{\bG_s^{-1}}/2,\\
        \bar{V}_{s,i}&\ge \frac{1}{2} (1-\inner{\bphi_s(i), \hat{\btheta}_s+2 \rho \norm{\bphi_s(i)}_{\bG_s^{-1}}})\ge \rho \norm{\bphi_s(i)}_{\bG_s^{-1}}/2,
    \end{align}
    which concludes the second inequality.
\end{proof}

\begin{lemma}\label{apdx_lem:R}
    Let $\bv=\bG_t^{-1}\bb_t$, if $\neg F_{t-1}$, then $R_{\bv}\le \frac{2{\norm{\bZ_t}_{\bG_t^{-1}}}}{\rho}.$
\end{lemma}
\begin{proof}
     For all $s <t$ and $i \in [m]$, we have $\inner{\bphi_s(i), \bv}/\bar{V}_{s,i} \le \frac{2\inner{\bphi_s(i), \bv}}{\norm{\bphi_s(i)}_{\bG_t^{-1}}\cdot \rho}=\frac{2\inner{\bphi_s(i), \bG_{t}^{-1}\bZ_t}}{\norm{\bphi_s(i)}_{\bG_t^{-1}}\cdot \rho}\le \frac{2\norm{\bphi_s(i)}_{\bG_t^{-1}}\cdot \norm{\bG_t^{-1}\bZ_t}_{\bG_t}}{\norm{\bphi_s(i)}_{\bG_t^{-1}}\cdot \rho}=\frac{2\norm{\bZ_t}_{\bG_t^{-1}}}{\rho}$, where the first inequality follows from \Cref{apdx_lem:beta_and_v_norm_potential}, the last inequality follows from the Cauchy-Schwarz inequality.
\end{proof}

\begin{lemma}\label{apdx_lem:l2_norm}
If $\neg F_t$, then $\norm{\bphi_t(i)}_{2}^2/\bar{V}_{t,i} \le 4dKt$.
\end{lemma}
\begin{proof}
If $\bar{V}_{t,i}=\frac{1}{4}$, the inequality trivially holds since $\norm{\bphi_t(i)}\le 1$. Consider $\bar{V}_{t,i}<\frac{1}{4}$, and $\lambda_{\max}$ be the maximum eigenvalue of $\bG_t$. Then, it holds that $\norm{\bphi_t(i)}_{2}^2/\bar{V}_{t,i} \le \frac{\norm{\bphi_t(i)}_{2}^2}{\rho \norm{\bphi_t(i)}_{\bG_t^{-1}}}\le\frac{\norm{\bphi_t(i)}_{2}}{ \norm{\bphi_t(i)}_{\bG_t^{-1}}}=\frac{\norm{\bG_t^{1/2}\bG_t^{-1/2} \bphi_t(i)}_{2}}{ \norm{\bphi_t(i)}_{\bG_t^{-1}}} \le \sqrt{\lambda_{\max}}$, where the first inequality follows from \Cref{apdx_lem:beta_and_v_norm_potential}, the second inequality is by $\rho\ge 1, \norm{\bphi_t(i)}\le 1$, and the last is by \Cref{apdx_fact:matrix}.3.

Now Assume $\norm{\bphi_s(i)}_{2}^2/\bar{V}_{s,i} \le 4s$ for $s< t$, which always holds for $t=1$. By reduction, we consider round $t$, it holds that $\norm{\bphi_t(i)}_{2}^2/\bar{V}_{t,i}\le\sqrt{\lambda_{\max}}\le \sqrt{\trace(\bG_t)}= \sqrt{\gamma d+\sum_{s=1}^{t-1}\sum_{i \in \tau_s}\norm{\bphi_s(i)}^2_{2}/\bar{V}_{s,i}}\le \sqrt{Kd+\sum_{s=1}^{t-1}4dK^2s}\le \sqrt{d(K + 2K^2t(t-1))}\le 4dKt$, where the first inequality follows from the analysis in the last paragraph, the third inequality follows from reduction over $s<t$, and the last inequality is by math calculation.
\end{proof}

\begin{lemma}\label{apdx_lem:l1_norm}
If $\neg F_t$, then $\norm{\bphi_t(i)}_{1}/\bar{V}_{t,i} \le 4dKt$.
\end{lemma}
\begin{proof}
Similar to the proof of \Cref{apdx_lem:l2_norm}, $\norm{\bphi_t(i)}_{1}/\bar{V}_{t,i}\le\sqrt{d}\norm{\bphi_t(i)}_{2}/\bar{V}_{t,i} \le \frac{\sqrt{d}\norm{\bphi_t(i)}_{2}}{\rho \norm{\bphi_t(i)}_{\bG_t^{-1}}}\le\frac{\norm{\bphi_t(i)}_{2}}{ \norm{\bphi_t(i)}_{\bG_t^{-1}}}=\frac{\norm{\bG_t^{1/2}\bG_t^{-1/2} \bphi_t(i)}_{2}}{ \norm{\bphi_t(i)}_{\bG_t^{-1}}} \le \sqrt{\lambda_{\max}}\le 4dKt$, where the first inequality uses Cauchy-Schwarz, the second inequality uses $\rho \ge \sqrt{d}$, and the rest follows from the proof of \Cref{apdx_lem:l2_norm}.
\end{proof}

\begin{lemma}\label{apdx_lem:Zt_l1_norm}
If $\neg F_{t-1}$, then $\norm{\bZ_t}_1\le 2dK^2t^2 $.
\end{lemma}
\begin{proof}
$\norm{\bZ_t}_1=\norm{\sum_{s<t}\sum_{i \in \tau_s}\eta_{s,i}\bphi_s(i)/\bar{V}_{s,i}}_{1}\le \sum_{s<t}\sum_{i \in \tau_s} \norm{\bphi_s(i)/\bar{V}_{s,i}}_1\le \sum_{s<t}\sum_{i \in \tau_s}4dKt\le 2dK^2t^2$, where the first inequality follows from $\eta_{s,i}\in [-1,1]$, the second inequality follows from \Cref{apdx_lem:l1_norm}.
\end{proof}

\begin{lemma}[Lemma A.3, \cite{li2016contextual}]\label{apdx_lem:det_and_norm}
    Let $\bx_i \in \R^d$, $1\le i \le n$. Then we have 
    \begin{align*}
        \det\left(\bI + \sum_{i=1}^n{\bx_i \bx_i^{\top}}\right)\ge 1+\sum_{i=1}^n\norm{\bx}_{2}^2.
    \end{align*}
\end{lemma}

\begin{lemma}[Lemma 11, \cite{abbasi2011improved}]\label{apdx_lem:det_and_trace}
Let $\bx_i \in \R^d$ with $\norm{\bx_i}_2\le L$, $1\le i \le n$ and let $\bG_t=\gamma \bI + \sum_{i=1}^{t-1} \bx_i \bx_i^{\top}$, then
\begin{align*}
 \det(\bG_{t+1}) \le  (\gamma + tL^2/d)^d.
\end{align*}

\end{lemma}

\begin{lemma}\label{apdx_lem:kappa_tpvm}
    \Cref{apdx_eq:kappa_tpvm} holds.
\end{lemma}
\begin{proof}

\textbf{When $T_{t-1, i} > L_{i, T, 2}=\frac{8c_1^2 B_v^2K\log T}{(\tDelta_i^{\min})^2}$,}

we have $(\ref{apdx_eq:kappa_tpvm}, i)\le  \frac{8c_1^2 B_v^2 \log T}{T_{t-1, i, }\cdot \tDelta_{S_t}} -\frac{\tDelta_{S_t}}{K} <\frac{(\tDelta_i^{\min})^2}{K\tDelta_{S_t}}-\frac{\tDelta_{S_t}}{K}\le  0 = \kappa_{i, T} (T_{t-1, i})$.

\textbf{When $ L_{i, T, 1} < T_{t-1, i} \le L_{i, T, 2}$,}

We have $(\ref{apdx_eq:kappa_tpvm}, i)\le  \frac{8c_1^2 B_v^2 \log T}{T_{t-1, i, }\cdot \tDelta_{S_t}} -\frac{\tDelta_{S_t}}{K}< \frac{8c_1^2 B_v^2 \log T}{T_{t-1, i, }\cdot \tDelta_{S_t}} \le \frac{8c_1^2 B_v^2 \log T}{T_{t-1, i}\cdot \tDelta^i_{\min}}=   \kappa_{i, T} (T_{t-1, i,j_i^{S_t}})$.

\textbf{When $T_{t-1, i} \le L_{i, T, 1}$},

We further consider two different cases $T_{t-1, i} \le \frac{4c_1^2 B_v^2 \log T}{(\tDelta_{S_t})^2}$ or $\frac{4c_1^2 B_v^2 \log T}{(\tDelta_{S_t})^2} < T_{t-1, i} \le L_{i,T,1}= \frac{4c_1^2 B_v^2 \log T}{(\tDelta_{i}^{\min})^2}$.

For the former case, if there exists $i\in \tau_t$ so that 
$T_{t-1, i} \le \frac{4c_1^2 B_v^2\log T}{(\tDelta_{S_t})^2}$, then we know  $\sum_{q \in \tilde{S}_t} \kappa_{q, T} (T_{t-1, q}) \ge \kappa_{i, T} (T_{t-1, i})=2\sqrt{\frac{4c_1^2 B_v^2 \log T}{T_{t-1, i}}}  \ge 2\tDelta_{S_t} > \Delta_{S_t}$, which makes \cref{apdx_eq:kappa_tpvm} holds no matter what. This means we do not need to consider this case for good.

% For the former case, if  $N_{t-1, i,j_i^{S_t}} \le \frac{24c_1^2 B_v^2 2^{(-j_i^{S_t}+1)({\lambda-1})}\log T}{(\Delta_{S_t})^2}$, then we know  $\sum_{q \in \tilde{S}_t} \kappa_{i,j_q^{S_t}, T} (N_{t-1, q,j_q^{S_t}}) \ge \kappa_{i,j_i^{S_t}, T} (N_{t-1, i,j_i^{S_t}})=2\sqrt{\frac{24c_1^2 B_v^2 2^{(-j_i^{S_t}+1)({\lambda-1})}\log T}{N_{t-1, i,j_i^{S_t}}}}  \ge 2\Delta_{S_t} > \Delta_{S_t}$, which makes \cref{apdx_eq:kappa_e1} holds no matter what.

For the later case, when $\frac{4c_1^2 B_v^2 \log T}{(\tDelta_{S_t})^2} < T_{t-1, i}$, we know that $(\ref{apdx_eq:kappa_tpvm}, i)\le\frac{8c_1^2 B_v^2\log T}{\tDelta_{S_t}} \frac{1}{T_{t-1, i}} =2\sqrt{\frac{4c_1^2 B_v^2 \log T}{(\tDelta_{S_t})^2} \frac{1}{T_{t-1, i}}}\sqrt{\frac{4c_1^2 B_v^2 \log T}{T_{t-1, i}}} \le 2\sqrt{\frac{4c_1^2 B_v^2 \log T}{T_{t-1, i}}}=\kappa_{i, T} (T_{t-1, i})$. 

\textbf{When  $\ell=0$,}

We have $(\ref{apdx_eq:kappa_tpvm}, i)\le \frac{8c_1^2 B_v^2 }{\tDelta_{S_t}}\cdot \frac{1}{28} -\frac{\tDelta_{S_t}}{K} \le \frac{c_1^2 B_v^2 }{\tDelta_{S_t}}\le \frac{c_1^2 B_v^2 }{\tDelta_i^{\min}}=\kappa_{i, T} (T_{t-1, i})$.

Combining all above cases, we have  $\Delta_{S_t} \le \E[\sum_{i \in \tau_t} \kappa_{i, T} (T_{t-1, i})]$.
\end{proof}

\begin{lemma}\label{apdx_lem:kappa_tpvm_general}
    \Cref{apdx_eq:kappa_tpvm_general} holds.
\end{lemma}
\begin{proof}

\textbf{When $T_{t-1, i} > L_{i, T, 2}=\frac{8c_1^2 B_v^2K\log T}{\tDelta_i^{\min}\cdot \tDelta_{i,\lambda}^{\min}}$,}

we have $(\ref{apdx_eq:kappa_tpvm_general}, i)\le  \frac{8c_1^2 B_v^2 \log T}{T_{t-1, i, }\cdot \tDelta_{S_t,\lambda}} -\frac{\tDelta_{S_t}}{K} <\frac{\tDelta_i^{\min}\cdot \tDelta_{i,\lambda}^{\min}}{K\tDelta_{S_t,\lambda}}-\frac{\tDelta_{S_t}}{K}\le  0 = \kappa_{i, T} (T_{t-1, i})$.

\textbf{When $ L_{i, T, 1} < T_{t-1, i} \le L_{i, T, 2}$,}

We have $(\ref{apdx_eq:kappa_tpvm_general}, i)\le  \frac{8c_1^2 B_v^2 \log T}{T_{t-1, i}\cdot \tDelta_{S_t,\lambda}} -\frac{\tDelta_{S_t}}{K}< \frac{8c_1^2 B_v^2 \log T}{T_{t-1, i, }\cdot \tDelta_{S_t,\lambda}} \le \frac{8c_1^2 B_v^2 \log T}{T_{t-1, i}\cdot \tDelta^{i,\lambda}_{\min}}=   \kappa_{i, T} (T_{t-1, i})$.

\textbf{When $T_{t-1, i} \le L_{i, T, 1}$},

We further consider two different cases $T_{t-1, i} \le \frac{4c_1^2 B_v^2 \log T}{\tDelta_{S_t,\lambda}\cdot \tDelta_{S_t}}$ or $\frac{4c_1^2 B_v^2 \log T}{\tDelta_{S_t,\lambda}\cdot \tDelta_{S_t}} < T_{t-1, i} \le L_{i,T,1}= \frac{4c_1^2 B_v^2 \log T}{\tDelta_{i,\lambda}^{\min}\cdot\tDelta_{i}^{\min}}$.

For the former case, if there exists $i\in \tau_t$ so that 
$T_{t-1, i} \le \frac{4c_1^2 B_v^2\log T}{\tDelta_{S_t,\lambda}\cdot \tDelta_{S_t}}$, then we know  $\sum_{q \in \tilde{S}_t} \kappa_{q, T} (T_{t-1, q}) \ge \kappa_{i, T} (T_{t-1, i})=2\sqrt{\frac{4c_1^2 B_v^2 \log T}{T_{t-1, i}}}  \ge 2\sqrt{\tDelta_{S_t,\lambda}\cdot \tDelta_{S_t}} \ge \Delta_{S_t}$, which makes \cref{apdx_eq:kappa_tpvm_general} holds no matter what. This means we do not need to consider this case for good.

% For the former case, if  $N_{t-1, i,j_i^{S_t}} \le \frac{24c_1^2 B_v^2 2^{(-j_i^{S_t}+1)({\lambda-1})}\log T}{(\Delta_{S_t})^2}$, then we know  $\sum_{q \in \tilde{S}_t} \kappa_{i,j_q^{S_t}, T} (N_{t-1, q,j_q^{S_t}}) \ge \kappa_{i,j_i^{S_t}, T} (N_{t-1, i,j_i^{S_t}})=2\sqrt{\frac{24c_1^2 B_v^2 2^{(-j_i^{S_t}+1)({\lambda-1})}\log T}{N_{t-1, i,j_i^{S_t}}}}  \ge 2\Delta_{S_t} > \Delta_{S_t}$, which makes \cref{apdx_eq:kappa_e1} holds no matter what.

For the later case, when $\frac{4c_1^2 B_v^2 \log T}{\tDelta_{S_t,\lambda}\cdot \tDelta_{S_t}} < T_{t-1, i}$, we know that $(\ref{apdx_eq:kappa_tpvm_general}, i)\le\frac{8c_1^2 B_v^2\log T}{\tDelta_{S_t,\lambda}} \frac{1}{T_{t-1, i}} =2\sqrt{\frac{4c_1^2 B_v^2 \log T}{(\tDelta_{S_t,\lambda})^2} \frac{1}{T_{t-1, i}}}\sqrt{\frac{4c_1^2 B_v^2 \log T}{T_{t-1, i}}} \le 2\sqrt{\frac{\tDelta_{S_t}}{\tDelta_{S_t,\lambda}}}\sqrt{\frac{4c_1^2 B_v^2 \log T}{T_{t-1, i}}}\le 2\sqrt{\frac{4c_1^2 B_v^2 \log T}{T_{t-1, i}}}=\kappa_{i, T} (T_{t-1, i})$. 

\textbf{When  $\ell=0$,}

We have $(\ref{apdx_eq:kappa_tpvm_general}, i)\le \frac{8c_1^2 B_v^2 }{\tDelta_{S_t,\lambda}}\cdot \frac{1}{28} -\frac{\tDelta_{S_t}}{K} \le \frac{c_1^2 B_v^2 }{\tDelta_{S_t,\lambda}}\le \frac{c_1^2 B_v^2 }{\tDelta_{i,\lambda}^{\min}}=\kappa_{i, T} (T_{t-1, i})$.

Combining all above cases, we have  $\Delta_{S_t} \le \E[\sum_{i \in \tau_t} \kappa_{i, T} (T_{t-1, i})]$.
\end{proof}

\begin{lemma}\label{apdx_lem:tpvm_lambda_term_2}
    \Cref{apdx_eq:tpvm_lambda_term_2} holds.
\end{lemma}
\begin{proof}

\textbf{When $T_{t-1, i} > L_{i,T,2}=\frac{4c_2 B_1 K\log T}{\Delta_i^{\min}}$,}

we have $(\ref{apdx_eq:tpvm_lambda_term_2}, i)\le 4c_2 B_1  \frac{\log T}{T_{t-1, i}} -\frac{\Delta_{S_t}}{K} 
<\frac{\Delta_i^{\min}}{K}-\frac{\Delta_{S_t}}{K}\le  0 = \kappa_{i, T} (T_{t-1, i})$.

\textbf{When $T_{t-1, i} \le L_{i, T, 2}$,}

We have $(\ref{apdx_eq:tpvm_lambda_term_2}, i)\le 4c_2 B_1  \frac{\log T}{T_{t-1, i}} -\frac{\Delta_{S_t}}{K}  <\frac{4c_2 B_1\log T}{T_{t-1, i}} = \kappa_{i,j_i^{S_t}, T} (N_{t-1, i,j_i^{S_t}})$.

\textbf{When $T_{t-1, i} \le L_{i, T, 1}$,}

If there exists $i \in \tilde{S}_t$ so that $T_{t-1, i}\le  L_{i, T, 1}$, then we know $\sum_{q \in \tilde{S}_t} \kappa_{i, T} (T_{t-1, q}) \ge \kappa_{i, T} (T_{t-1, i})=\Delta_{i}^{\max} \ge \Delta_{S_t}$, which makes \cref{apdx_eq:tpvm_lambda_term_2} holds no matter what. This means we do not need to consider this case for good.

Combining all above cases, we have  $\Delta_{S_t} \le \E_t[\sum_{i \in \tau_t} \kappa_{i, T} (T_{t-1, i})]$.
\end{proof}
\end{document}